\documentclass[lettersize,journal]{IEEEtran}
\usepackage{amsmath,amsfonts}
\usepackage{algorithmic}
\usepackage{algorithm}
\usepackage{array}
\usepackage[caption=false,font=scriptsize,labelfont=sf,textfont=sf]{subfig}
\usepackage{textcomp}
\usepackage{stfloats}
\usepackage{url}
\usepackage{verbatim}
\usepackage{graphicx}
\usepackage{tikz}
\usepackage{cite}
\usepackage{hyperref}
\usepackage{bm}
\usepackage{multirow}
\usepackage{booktabs}
\usepackage{gensymb}
\usepackage{caption}
\usepackage[dvipsnames,table,xcdraw]{xcolor}
\usepackage{ragged2e}
\usepackage{amssymb}
\definecolor{lightgray}{gray}{0.93}

\hypersetup{
    colorlinks=true,
    linkcolor=blue,
    citecolor=green,
    urlcolor=magenta,
    hidelinks,
}

\begin{document}
\title{Learning-based 3D Reconstruction in Autonomous Driving: A Comprehensive Survey}

\author{
Liewen Liao,
Weihao Yan,
Wang Xu,
Ming Yang~\IEEEmembership{Member, ~IEEE},
Songan Zhang*,
H. Eric Tseng

\thanks{Liewen Liao, Songan Zhang are with the Global Institute of Future Technology, Shanghai Jiao Tong University, Shanghai, 200240, China.}
\thanks{Weihao Yan, Wang Xu, Ming Yang are with the School of Automation and Intelligent Sensing, Shanghai Jiao Tong University, Shanghai, 200240, China}
\thanks{H. Eric Tseng is with the Department of Electrical Engineering, University of Texas at Arlington, Arlington, TX 76019, USA}
\thanks{This work was supported in part by the National Nature Science Foundation of China under Grant 52402504 and Grant 62473250. \emph{Corresponding author:} Songan Zhang, email: songanz@sjtu.edu.cn.}
}


\markboth{Journal of \LaTeX\ Class Files,~Vol.~14, No.~8, August~2021}%
{Liao \MakeLowercase{\textit{et al.}}: Learning-based 3D Reconstruction in Autonomous Driving: A Comprehensive Survey}

\maketitle

\begin{abstract}
Learning-based 3D reconstruction has emerged as a transformative technique in autonomous driving, enabling precise modeling of environments through advanced neural representations. It has inspired pioneering solutions for vital tasks in autonomous driving, such as dense mapping and closed-loop simulation, as well as comprehensive scene feature for driving scene understanding and reasoning. Given the rapid growth in related research, this survey provides a comprehensive review of both technical evolutions and practical applications in autonomous driving. We begin with an introduction to the preliminaries of learning-based 3D reconstruction to provide a solid technical background foundation, then progress to a rigorous, multi-dimensional examination of cutting-edge methodologies, systematically organized according to the distinctive technical requirements and fundamental challenges of autonomous driving. Through analyzing and summarizing development trends and cutting-edge research, we identify existing technical challenges, along with insufficient disclosure of on-board validation and safety verification details in the current literature, and ultimately suggest potential directions to guide future studies.
\end{abstract}


\section{Introduction}\label{sec1}

\IEEEPARstart{A}{utonomous} driving has garnered tremendous research attention due to its potential to improve transportation safety and efficiency. Achieving reliable self-driving capability fundamentally hinges on precise perception and a holistic understanding of the 3D environment, which demands large-scale, diverse data collected with a professional sensor suite. However, acquiring such extensive multimodal datasets is expensive and poses safety risks, especially for rare and emergency scenarios. Learning-based 3D reconstruction offers a groundbreaking solution to this challenge. By creating photorealistic and geometrically accurate digital twins of the physical world, 3D reconstruction can synthetically generate abundant training data at a low cost and minimal risk. These virtual yet realistic replicas of driving scenes enable data augmentation and scenario simulation, alleviating the data bottleneck in autonomous driving development.

Over the past few years, the field has witnessed a paradigm shift from traditional techniques to learning-based approaches. Traditional methods have well-known limitations of photogrammetric methods (e.g., structure-from-motion and multi-view stereo) struggle under varying lighting and suffer from distortion, while active sensors (LiDAR, depth cameras) provide accuracy at high equipment and processing cost. The advent of Neural Radiance Fields (NeRF)~\cite{nerf} in 2020 marked a turning point, demonstrating that neural scene representations can achieve unprecedented fidelity and scalability in 3D reconstruction. NeRF introduced a learning-based modeling approach using neural networks to represent scenes, which significantly outperformed hand-crafted models and prior reconstruction pipelines in rendering novel views with high realism. Building on this foundation, a wave of neural 3D reconstruction techniques~\cite{Mller2022InstantNGP, Barron2021MipNeRF, Barron2021MipNeRF360, feng2022neuralpoints, xu2022point, yang2022neumesh} have emerged. In particular, 3D Gaussian Splatting (3DGS)~\cite{3dgs} has recently gained prominence as an explicit alternative, representing scenes as sets of 3D Gaussian primitives that can be optimized and rendered directly. This explicit representation yields substantial speed advantages as 3DGS can leverage efficient rasterization with GPU acceleration to achieve real-time rendering, addressing one of NeRF’s key limitations. Together with other advanced neural techniques, NeRF and 3DGS exemplify the rise of learning-based 3D reconstruction methods that overcome many limitations of traditional approaches while offering improved fidelity and efficiency.

Learning-based 3D reconstruction has quickly become a foundational technology in the autonomous driving stack, with far-reaching applications across core tasks. Initially, 3D reconstruction was explored primarily for data augmentation and multimodal sensor simulation~\cite{turki2023suds, yuan2025uni}, but recent research has expanded its role to perception enhancement, improved scene semantics, and end-to-end world modeling for autonomous driving~\cite{gao2025rad}. High-fidelity 3D environment models enrich perception, providing geometric context to improve object detection and occupancy prediction~\cite{yuan2024presight, yang2024unipad}. Besides, learned 3D models contribute to general scene understanding, as they inherently capture the spatial layout and can be imbued with semantic information~\cite{wu2024emie, zheng2024gaussianad}. Perhaps most prominently, realistic 3D reconstructions enable immersive simulation by generating virtual cities and scenarios, where one can safely test and train autonomous driving systems in a controlled setting~\cite{yang2023unisim, yan2024oasim, wu2023mars, ljungbergh2024neuroncap, zhou2024hugsim, wei2024editable}. These reconstructions are increasingly integrated throughout the autonomous driving pipeline, from perception to prediction and planning, underscoring their importance. The rapid proliferation of publications in this area attests to its emerging significance.

\begin{figure*}[ht]
\centering
\captionsetup{type=figure}
\begin{tikzpicture}
    \node[anchor=south west, inner sep=0](image) at (0,0)
    {\includegraphics[width=\textwidth]{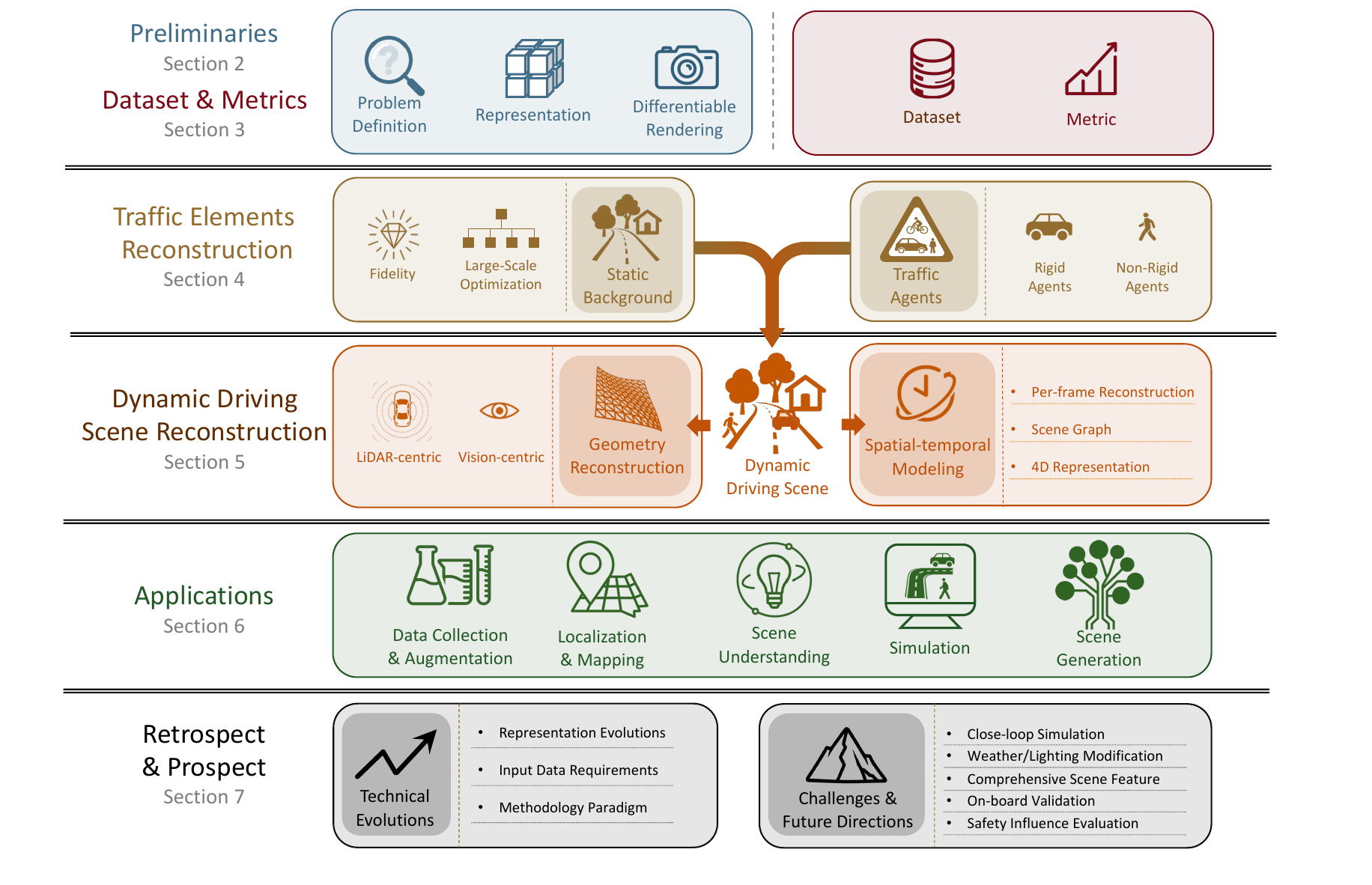}};
    \begin{scope}[x={(image.south east)}, y={(image.north west)}]
        \node at (0.15, 0.95) [minimum width=0.3\textwidth, minimum height=0.01\textwidth] {\hyperref[sec:prelim]{\phantom{\rule{1.9cm}{0.8cm}}}};
        \node at (0.285, 0.905) [minimum width=0.3\textwidth, minimum height=0.01\textwidth] {\hyperref[sec:definition]{\phantom{\rule{1.1cm}{1.3cm}}}};
        \node at (0.388, 0.91) [minimum width=0.3\textwidth, minimum height=0.01\textwidth] {\hyperref[sec:representation]{\phantom{\rule{1.35cm}{1.1cm}}}};
        \node at (0.5, 0.901) [minimum width=0.3\textwidth, minimum height=0.01\textwidth] {\hyperref[sec:rendering]{\phantom{\rule{1.25cm}{1.2cm}}}};
        \node at (0.15, 0.87) [minimum width=0.3\textwidth, minimum height=2cm] {\hyperref[sec:data]{\phantom{\rule{2.3cm}{0.8cm}}}};
        \node at (0.68, 0.907) [minimum width=0.3\textwidth, minimum height=0.01\textwidth] {\hyperref[sec:dataset]{\phantom{\rule{0.9cm}{1.0cm}}}};
        \node at (0.795, 0.907) [minimum width=0.3\textwidth, minimum height=0.01\textwidth] {\hyperref[sec:metric]{\phantom{\rule{0.9cm}{1.0cm}}}};
        \node at (0.15, 0.72) [minimum width=0.3\textwidth, minimum height=2cm] {\hyperref[sec:element]{\phantom{\rule{2.15cm}{1.15cm}}}};
        \node at (0.287, 0.721) [minimum width=0.3\textwidth, minimum height=0.01\textwidth] {\hyperref[sec:fidel]{\phantom{\rule{0.68cm}{0.8cm}}}};
        \node at (0.366, 0.715) [minimum width=0.3\textwidth, minimum height=0.01\textwidth] {\hyperref[sec:largescale]{\phantom{\rule{1.0cm}{1.1cm}}}};
        \node at (0.4572, 0.7162) [minimum width=0.3\textwidth, minimum height=0.01\textwidth] {\hyperref[sec:static]{\phantom{\rule{1.2cm}{1.4cm}}}};
        \node at (0.67, 0.7163) [minimum width=0.3\textwidth, minimum height=0.01\textwidth] {\hyperref[sec:agent]{\phantom{\rule{1.35cm}{1.44cm}}}};
        \node at (0.765, 0.7162) [minimum width=0.3\textwidth, minimum height=2cm] {\hyperref[sec:rigid]{\phantom{\rule{0.7cm}{0.9cm}}}};
        \node at (0.8358, 0.7166) [minimum width=0.3\textwidth, minimum height=2cm] {\hyperref[sec:nonrigid]{\phantom{\rule{0.7cm}{0.9cm}}}};
        \node at (0.15, 0.515) [minimum width=0.3\textwidth, minimum height=2cm] {\hyperref[sec:scene_recon]{\phantom{\rule{3.1cm}{1.0cm}}}};
        \node at (0.292, 0.522) [minimum width=0.3\textwidth, minimum height=2cm] {\hyperref[sec:lidarcentric]{\phantom{\rule{1.1cm}{0.95cm}}}};
        \node at (0.366, 0.522) [minimum width=0.3\textwidth, minimum height=2cm] {\hyperref[sec:visioncentric]{\phantom{\rule{1.07cm}{1.05cm}}}};
        \node at (0.456, 0.519) [minimum width=0.3\textwidth, minimum height=2cm] {\hyperref[sec:geometry]{\phantom{\rule{1.5cm}{1.6cm}}}};
        \node at (0.677, 0.52) [minimum width=0.3\textwidth, minimum height=2cm] {\hyperref[sec:spatialtemporal]{\phantom{\rule{1.55cm}{1.62cm}}}};
        \node at (0.805, 0.557) [minimum width=0.3\textwidth, minimum height=2cm] {\hyperref[sec:perframe]{\phantom{\rule{2.3cm}{0.21cm}}}};
        \node at (0.775, 0.513) [minimum width=0.3\textwidth, minimum height=2cm] {\hyperref[sec:scenegraph]{\phantom{\rule{1.2cm}{0.21cm}}}};
        \node at (0.79, 0.473) [minimum width=0.3\textwidth, minimum height=2cm] {\hyperref[sec:4drepre]{\phantom{\rule{1.6cm}{0.21cm}}}};
        \node at (0.15, 0.31) [minimum width=0.3\textwidth, minimum height=2cm] {\hyperref[sec:application]{\phantom{\rule{1.7cm}{0.55cm}}}};
        \node at (0.328, 0.315) [minimum width=0.3\textwidth, minimum height=2cm] {\hyperref[sec::data_collection]{\phantom{\rule{1.5cm}{1.5cm}}}};
        \node at (0.44, 0.315) [minimum width=0.3\textwidth, minimum height=2cm] {\hyperref[sec:slam]{\phantom{\rule{1.35cm}{1.6cm}}}};
        \node at (0.565, 0.32) [minimum width=0.3\textwidth, minimum height=2cm] {\hyperref[sec:understand]{\phantom{\rule{1.5cm}{1.5cm}}}};
        \node at (0.68, 0.32) [minimum width=0.3\textwidth, minimum height=2cm] {\hyperref[sec:simulation]{\phantom{\rule{1.5cm}{1.5cm}}}};
        \node at (0.802, 0.317) [minimum width=0.3\textwidth, minimum height=2cm] {\hyperref[sec:scenegen]{\phantom{\rule{1.3cm}{1.6cm}}}};
        \node at (0.15, 0.135) [minimum width=0.3\textwidth, minimum height=2cm] {\hyperref[sec:retro]{\phantom{\rule{1.6cm}{0.95cm}}}};
        \node at (0.289, 0.126) [minimum width=0.3\textwidth, minimum height=2cm] {\hyperref[sec:techevo]{\phantom{\rule{1.3cm}{1.4cm}}}};
        \node at (0.412, 0.087) [minimum width=0.3\textwidth, minimum height=2cm] {\hyperref[sec:repreevo]{\phantom{\rule{2.3cm}{0.22cm}}}};
        \node at (0.415, 0.172) [minimum width=0.3\textwidth, minimum height=2cm] {\hyperref[sec:inputreq]{\phantom{\rule{2.2cm}{0.22cm}}}};
        \node at (0.413, 0.13) [minimum width=0.3\textwidth, minimum height=2cm] {\hyperref[sec:methpara]{\phantom{\rule{2.2cm}{0.22cm}}}};
        \node at (0.617, 0.126) [minimum width=0.3\textwidth, minimum height=2cm] {\hyperref[sec:challenge]{\phantom{\rule{1.8cm}{1.4cm}}}};
        \node at (0.751, 0.172) [minimum width=0.3\textwidth, minimum height=2cm] {\hyperref[sec:closesim]{\phantom{\rule{2.1cm}{0.13cm}}}};
        \node at (0.77, 0.147) [minimum width=0.3\textwidth, minimum height=2cm] {\hyperref[sec:weather]{\phantom{\rule{2.7cm}{0.13cm}}}};
        \node at (0.756, 0.121) [minimum width=0.3\textwidth, minimum height=2cm] {\hyperref[sec:integration]{\phantom{\rule{2.7cm}{0.13cm}}}};
        \node at (0.745, 0.095) [minimum width=0.3\textwidth, minimum height=2cm] {\hyperref[sec:onboard]{\phantom{\rule{2.0cm}{0.13cm}}}};
        \node at (0.765, 0.071) [minimum width=0.3\textwidth, minimum height=2cm] {\hyperref[sec:safety]{\phantom{\rule{2.5cm}{0.13cm}}}};
    \end{scope}
\end{tikzpicture}
\vspace{-3.0em}
\captionof{figure}{Survey outlines. We start from preliminaries and essentials of 3D reconstruction, then elaborate on the technical evolutions of traffic elements and dynamic driving scene reconstruction with tailored taxonomy, and diverse applications within autonomous driving. Finally, we summarize and delineate challenges and future directions. Click the section title or icon to jump to the corresponding section.}\label{fig::sections}\vspace{-1.5em}
\end{figure*}

Given the transformative developments outlined above, a comprehensive survey of learning-based 3D reconstruction in autonomous driving is timely and valuable. On one hand, the state of the art has advanced rapidly with the introduction of NeRF, 3DGS, and numerous follow-up works, resulting in a flourishing and complex landscape of methods. The volume of related research has grown exponentially in recent years, making it difficult for beginners to obtain a holistic perspective. On the other hand, existing review efforts have been limited in scope. A few recent surveys focus exclusively on specific techniques, such as neural radiance fields~\cite{he2024nerfsurvey} or 3D Gaussian splatting~\cite{bao20253dgssurvey, zhu20243dgssurvey}, or they examine 3D reconstruction outside the autonomous driving context. These surveys provide depth on particular approaches but tend to overlook the broader picture and the cross-technique synergies that are crucial for practical applications within autonomous driving. In contrast, our survey aims to offer a holistic view of the field, bridging multiple methodologies and linking them to practical autonomous driving needs, thereby equipping researchers and practitioners with a comprehensive and well-founded guide to the current state-of-the-art. The contributions of this survey can be summarized as follows:

\begin{itemize}
    \item \textbf{Technical foundation}: We introduce problem definitions, sensor modalities, datasets, and key representation/rendering techniques.
    \item \textbf{Systematic review}: We organize state-of-the-art methods with hierarchical taxonomy based on the practical challenge within autonomous driving.
    \item \textbf{Application-centric view}: We discuss how reconstruction supports core tasks including data augmentation, mapping \& localization, simulation and scene generation.
    \item \textbf{Future outlook}: We delineate the technical trends and further identify persistent challenges including simulation realism, on-board validation and safety concerns, and suggest future directions.
\end{itemize}

The organization of this survey is shown in Fig.~\ref{fig::sections}: In Section \ref{sec:prelim}, we introduce preliminaries for 3D reconstruction, including problem definition, representation, and rendering methods for learning-based 3D reconstruction. Section \ref{sec:data} presents the datasets and metrics used to evaluate 3D reconstruction methods. In Section \ref{sec:element}, we elaborate on the distinct characteristics and challenges within different elements of the driving scene. Section \ref{sec:scene_recon} systematically examines reconstructions of the dynamic driving scene. In Section \ref{sec:application}, we review applications of 3D reconstruction-based technology. Finally, we highlight persisting challenges, safety concerns, and potential directions based on the analysis of development trends in Section \ref{sec:retro}.

\section{Preliminary}\label{sec:prelim}
3D reconstruction aims to recover the complete and accurate geometry and appearance details of a target from 2D observations, which encompasses traditional photogrammetric methods, such as Structure from Motion (SfM), as well as learning-based methods like Neural Radiance Fields~\cite{nerf} or 3D Gaussian Splatting~\cite{3dgs}. Learning-based methods do not strictly follow the traditional pipeline of feature extraction, matching, and bundle adjustment; instead, they perform optimization in an end-to-end manner directly through parameterized representations and gradient-based optimization algorithms with differentiable rendering. Comparing with traditional methods, learning-based methods offer distinct advantages in fidelity and scalability, yet they also facing challenges such as aliasing artifacts and computationally intensive. In this section, we will introduce the technical preliminaries of learning-based 3D reconstruction by covering problem definition, representation, and differentiable rendering methods.\vspace{-0.5em}

\subsection{Problem Definition}\label{sec:definition}

Given a set of observations $\{o_i, i\in[1, \dots, N]\}$ from perspectives $\{p_i, i\in[1, \dots, N]\}$, learning-based 3D reconstruction methods optimize a parameterized scene representation $F_\theta(\cdot)$ to accurately capture the geometry and appearance, which is commonly formulated as correctly reconstructing observation $\{\hat{o}_i, i\in[1, \dots, N]\}$ from input views and synthesizing observation $\hat{o}_i, i\in[N+1, \dots]$ from test perspectives $\{p_i, i\in[N+1, \dots]$. The optimization can be formulated as:

\vspace{-0.6em}
\begin{equation}
    \arg\min_\theta \sum_{i}\mathcal{L}(F_\theta(p_i),\hat{o}_i), i\in[1, \dots, N, N+1, \dots]
\end{equation}
\vspace{-1.1em}

\noindent where $\mathcal{L}(\cdot)$ indicates the reconstruction loss from perspective $\hat{o}_i$, $F_\theta$ indicates the parameterized representation. With the support of differentiable rendering techniques, this optimization problem can be solved in a gradient-based manner.\vspace{-1.0em}

\begin{table*}[t]
\centering
\caption{Pros \& Cons of 3D representations}
\label{tab:prosncons}
\rowcolors{2}{lightgray}{white}
\resizebox{\textwidth}{!}{%
\begin{tabular}{lll}
\toprule
Representation   & Pros                                                                                                                                                                         & Cons                                                                                        \\ \midrule
Implicit Surface & Precise analytical representation for geometry                                                                                                                               & Prohibitive complexity for complex geometry                                                 \\
SDF              & Efficient geometry representation                                                                                                                                            & No visual details                                                                           \\
NeRF             &\begin{tabular}[l]{@{}l@{}}High fidelity rendering; \\ End-to-end optimization with differentiable rendering\end{tabular}                                                    & \begin{tabular}[l]{@{}l@{}}High inference latency;\\ Severe aliasing artifacts\end{tabular} \\
Point Cloud      & \begin{tabular}[l]{@{}l@{}}Precise geometry representation;\\ High flexibility for aggregation and partitioning\end{tabular}                                                 & \begin{tabular}[l]{@{}l@{}}Unstructured data; \\ Low rendering fidelity\end{tabular}        \\ 
Voxel            & \begin{tabular}[l]{@{}l@{}}Structured data; \\ Capturing both surface and internal information; \\ Flexible attribute encoding\end{tabular}                                  & Lack of specific optimized hardware for voxel rendering                                     \\
Mesh             & Efficient and high-fidelity rendering                                                                                                                                        & Computational-intensive construction and modification                                       \\
3DGS             &\begin{tabular}[l]{@{}l@{}}Efficient and high-fidelity rendering; \\ End-to-end optimization with differentiable rendering;\\  Simple and parametric definition\end{tabular} &  Low geometric precision                                                                     \\ \bottomrule
\end{tabular}%
}\vspace{-1.5em}
\end{table*}

\subsection{Representation}\label{sec:representation}

\begin{figure}[t]
    \centering
    \vspace{-2.0em}
\subfloat[Implicit Surface~\cite{enwiki:1197924651}\hfill]{
\includegraphics[width=0.28\columnwidth]{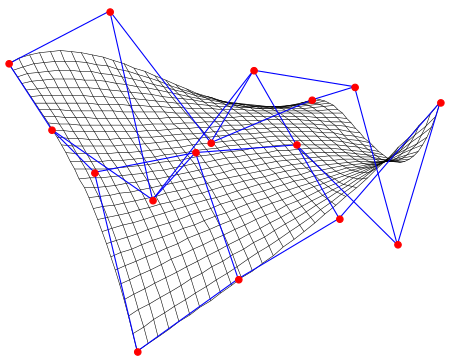}
}\hfill
\subfloat[SDF~\cite{li2023diffusionsdf}\hfill]{
\includegraphics[width=0.28\columnwidth]{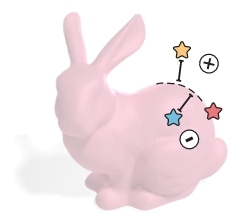}
}\hfill
\subfloat[NeRF~\cite{nerf}\hfill]{
\includegraphics[width=0.30\columnwidth]{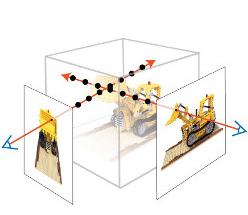}
}\quad
\subfloat[Point Cloud]{
\includegraphics[width=0.23\columnwidth]{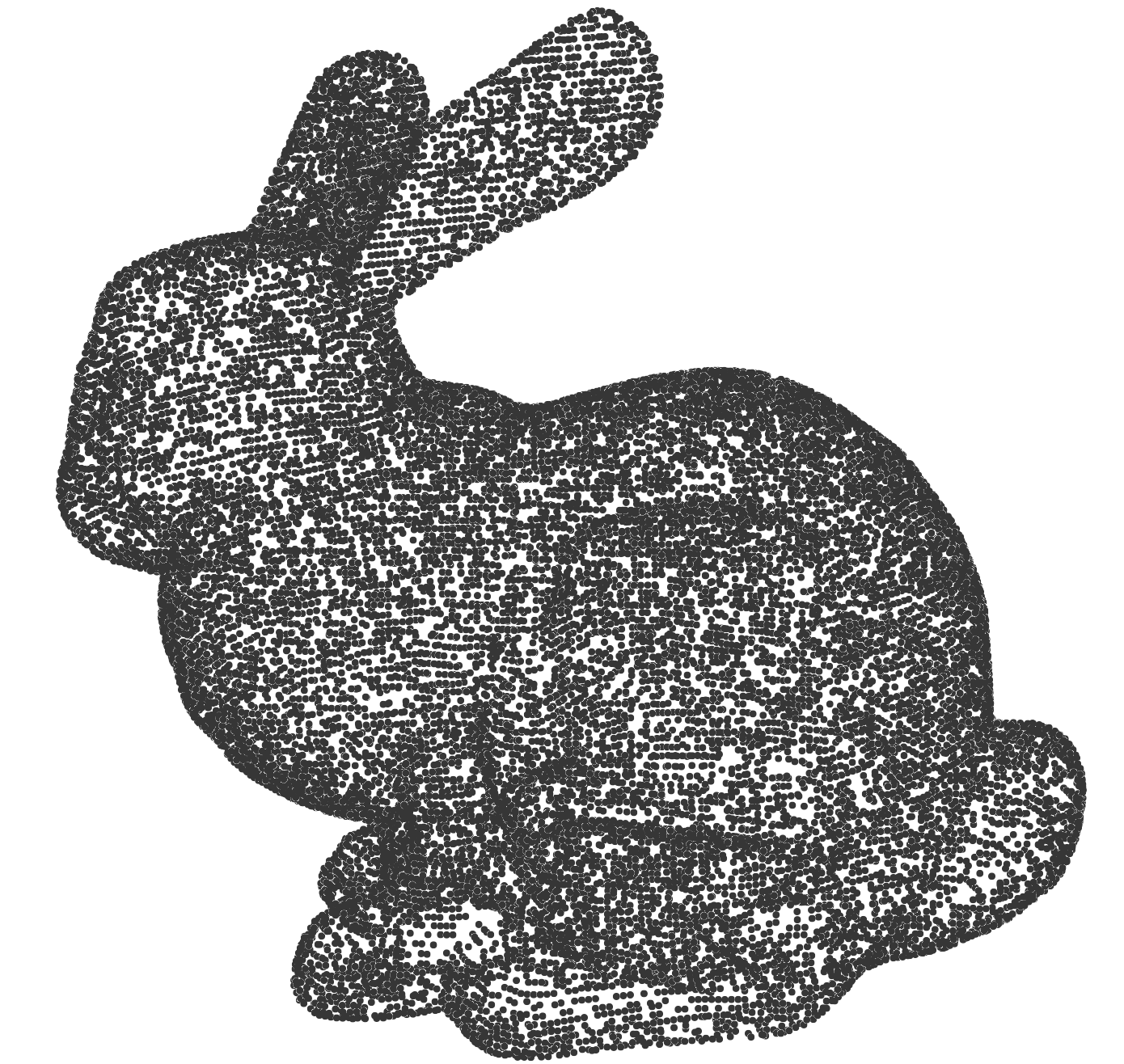}
}
\subfloat[Voxel]{
\includegraphics[width=0.21\columnwidth]{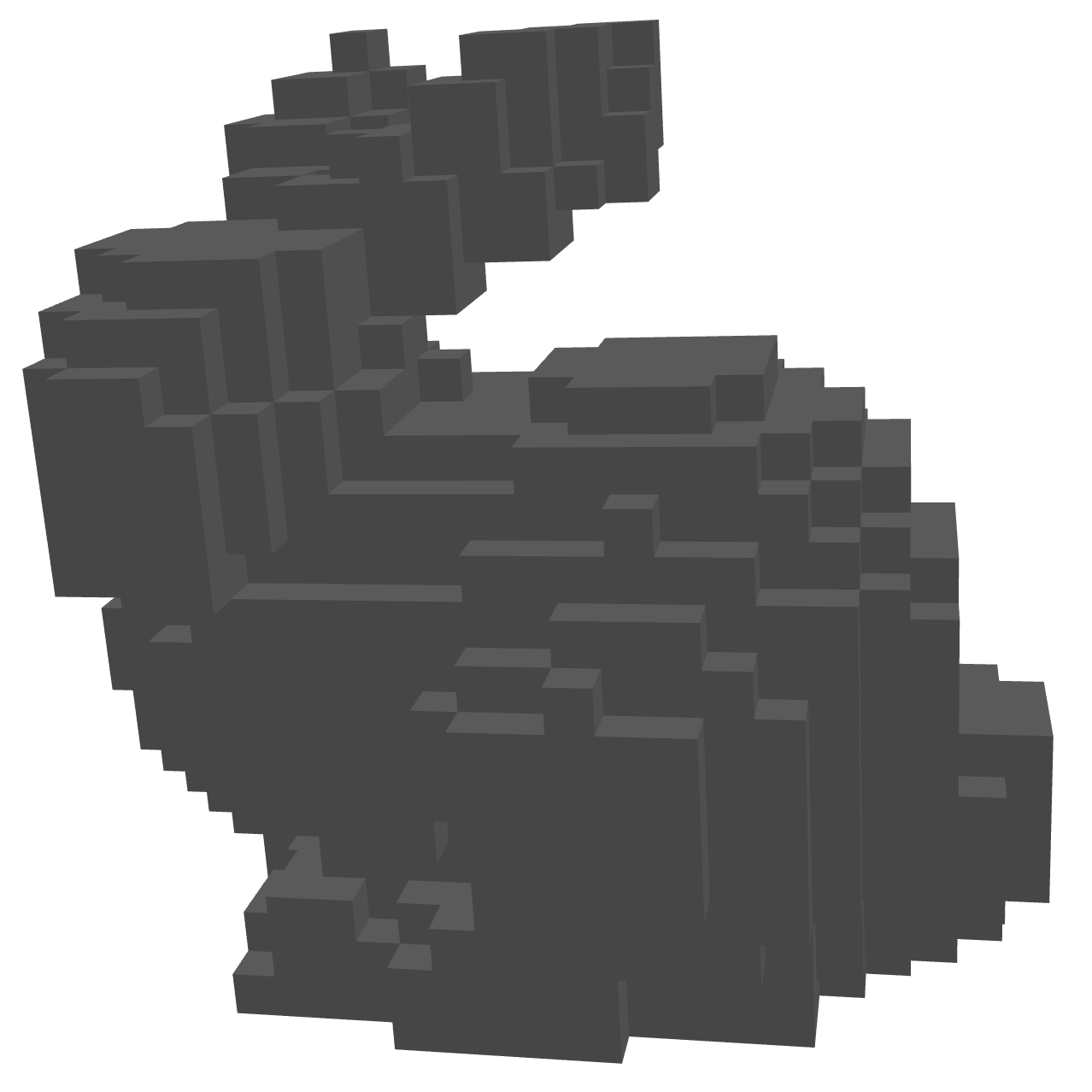}
}
\subfloat[Mesh]{
\includegraphics[width=0.21\columnwidth]{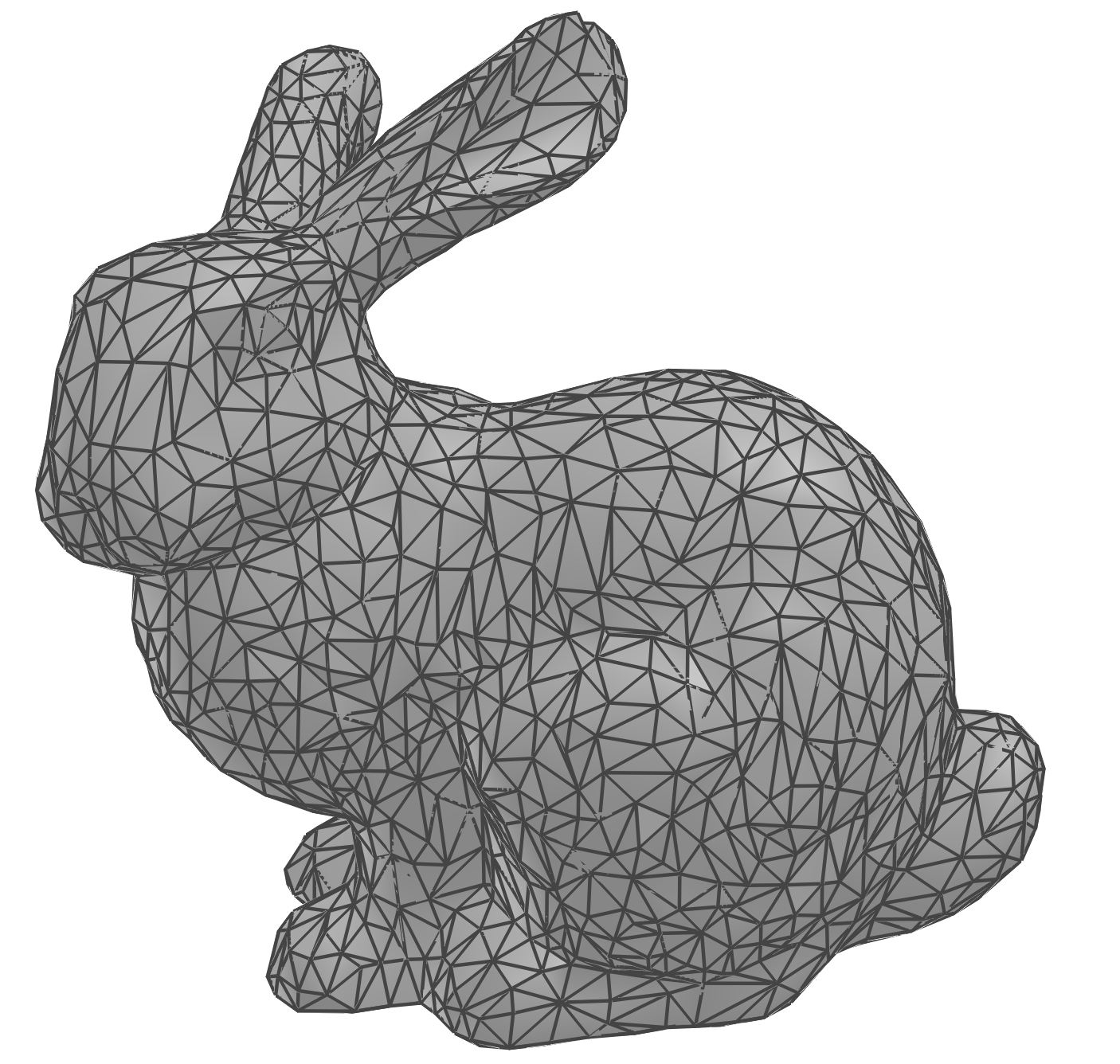}
}
\hspace{0.3em}\subfloat[3DGS~\cite{3dgs}]{
\includegraphics[width=0.2\columnwidth]{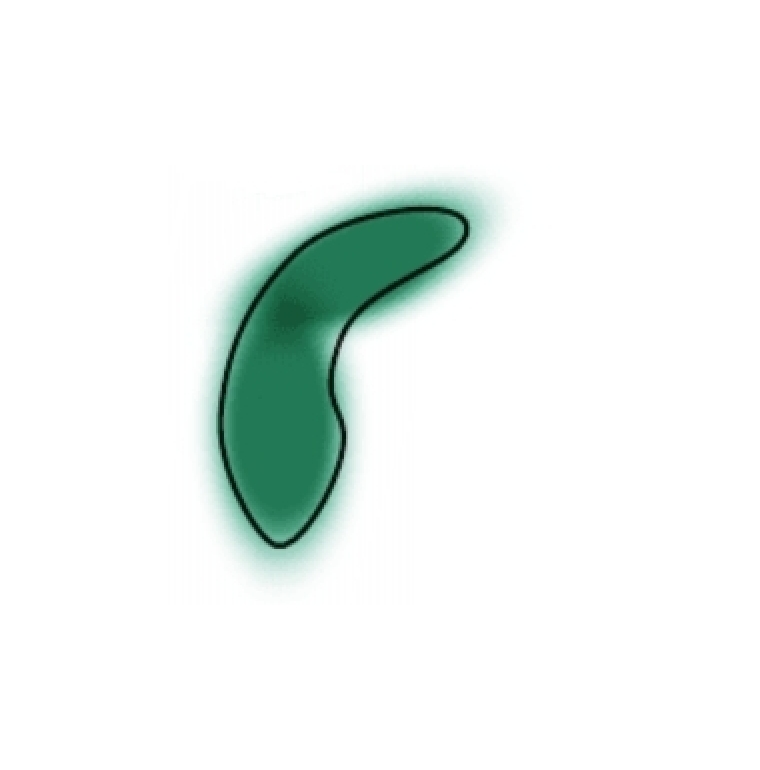}
}
    \caption{illustrations of different representations. \textit{Top:} Implicit Representations; \textit{Bottom:} Explicit Representations.}
    \label{fig:schematic}
    \vspace{-1.0em}
\end{figure}

Representation has a profound impact on reconstruction fidelity and computational efficiency, and can be classified into two main categories: implicit and explicit representations. By examining each representation in detail, we aim to highlight the respective advantages and disadvantages as summarized in Table~\ref{tab:prosncons}. Figure~\ref{fig:schematic} exhibits schematic illustrations of different representations.

\subsubsection{Implicit Representation}\hfill

\textbf{Implicit Surfaces} describe the surface of objects with equations, such as B\'ezier surfaces~\cite{bsurface, bsurface2, Rosli_Zulkifly_2023bsurface }, Non-Uniform Rational B-Splines (NURBS)~\cite{nurbs1, nurbs2}. Implicit surfaces provide precise analytical representation of object geometry, making them highly suitable for scenarios that require extremely high precision. However, implicit surfaces suffer from prohibitive complexity~\cite{bsurface} as the geometry details grow, which hinders the application in practical scenarios.

\textbf{Signed Distance Field} (SDF) defines a function that maps a target location to its minimum distance to the surface of the object. SDF utilizes an analytical continuous scalar function to encode geometry by assigning the shortest distance to the surface at each point, positive outside, negative inside, and zero directly on the boundary, and DeepSDF~\cite{park2019deepsdf} extends it into a learning-based paradigm. SDF provides an efficient option for applications concentrating on geometric while neglecting texture, such as collision calculation or geometry regularization of other representations.

\textbf{Neural Radiance Field} (NeRF)~\cite{nerf} maps target position and viewing direction to the volume density and radiance of the target point. NeRF pioneers optimizing neural representations to synthesize photorealistic novel views in a differentiable manner. However, NeRF suffers from aliasing artifacts and large inference latency arising from the substantial computational overhead in volumetric sampling. These limitations critically hinder real-time rendering, yet drive intensive researches~\cite{zhang2020nerf++, Barron2021MipNeRF, Barron2021MipNeRF360, Mller2022InstantNGP}.
\subsubsection{Explicit Representation}\hfill

\textbf{Point Cloud} is an unordered set of 3D points that samples the surface of an object or scene. Each point consists of coordinates and additional attributes such as color, intensity~\cite{li2024dgnr}, or normal vectors compared with the concept in data modality. Its unstructured nature offers flexibility for aggregation and partitioning, but also incurs computational challenges due to the lack of explicit topology.

\textbf{Voxel} consists of grid-aligned 3D cubic units that store attributes such as density or color, forming a structured volumetric representation. Voxel captures both surface and internal information, making them well-suited for biomedical and spatial reasoning tasks. Voxel grids allow flexible attribute encoding, from occupancy probabilities and semantic labels~\cite{zhang2023occformer, wang2024panoocc, tang2024sparseocc, yang2024adaptiveocc, xiao2024semantic} to learnable features in neural networks~\cite{omniscene, zhang2024geolrm}.

\textbf{Mesh} is a structured 3D surface representation composed of vertices, edges, and faces, typically 2D triangles or quadrilaterals, connected in a graph-like topology. This structured connectivity enables efficient projection and high-fidelity rendering, making meshes widely used in graphics and engineering. However, their interdependent components increase the complexity of construction and modification.

\textbf{3D Gaussians}~\cite{3dgs} represent a scene using a set of learnable anisotropic Gaussian ellipsoids, each parameterized by position, covariance (defining shape and orientation), and appearance attributes such as color and opacity. Leveraging the projection-friendly properties of Gaussian functions, 3D Gaussian achieves high rendering fidelity with low computational overhead, making it significantly more efficient than traditional surface- or volume-based methods. While it trades off some GPU memory to support large numbers of Gaussians, this enables real-time, high-quality rendering~\cite{lu2024scaffold, yu2024mipsplatting, Szymanowicz2023SplatterImage} and broad applicability, establishing 3D Gaussians as a state-of-the-art representation for both research and practical deployment.\vspace{-1.0em}

\subsection{Differentiable Rendering}\label{sec:rendering}
Unlike traditional rendering, which only simulates image formation, differentiable rendering enables gradient-based optimization of 3D representations by computing the derivatives of reconstruction loss with respect to representation parameters. This approach bridges image supervision and learnable representations in a differentiable pipeline, accelerating convergence and reducing reliance on ground-truth 3D data.

\begin{figure}[ht]
 \subfloat[NeRF with volume rendering~\cite{nerf}\label{1a}]{%
       \includegraphics[width=0.95\linewidth]{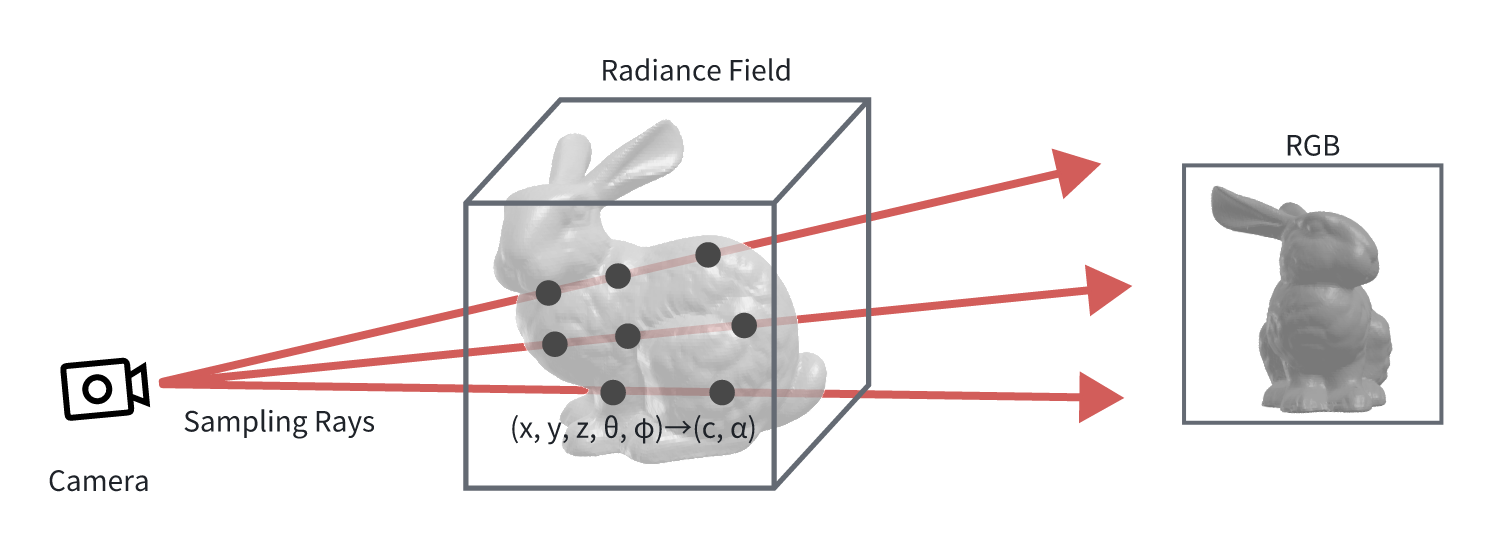}}
       \\
  \subfloat[Gaussian Splatting and rasterization~\cite{3dgs}\label{1b}]{%
        \includegraphics[width=0.95\linewidth]{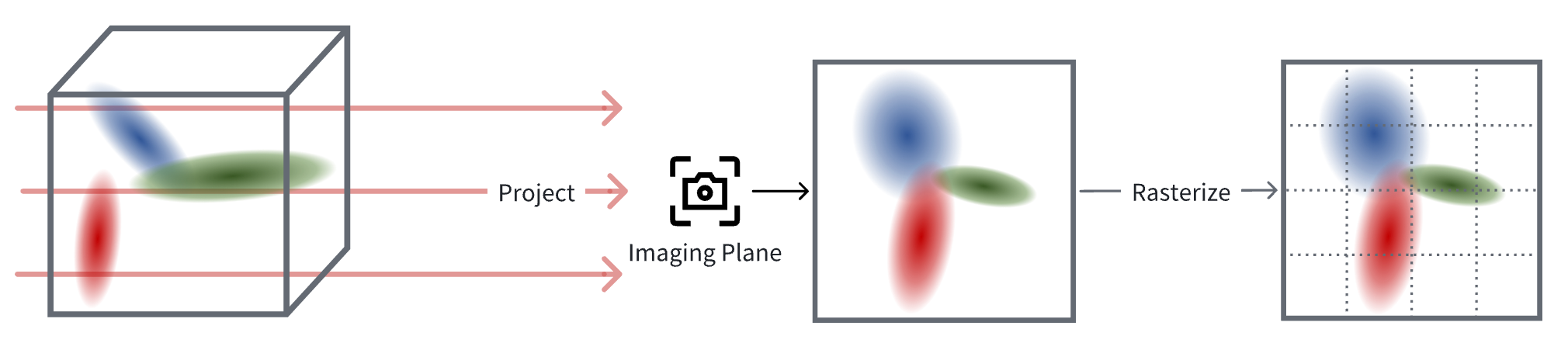}}
\caption{Rendering Pipeline of NeRF~\cite{nerf} and 3D Gaussian Splatting~\cite{3dgs}}\label{fig::milestone}
\vspace{-1.0em}
\end{figure}

The choice of rendering method has a critical impact on both reconstruction quality and efficiency. Implicit representations typically rely on volume rendering due to projection constraints, while explicit representations support more flexible strategies, with rasterization favored for its efficiency. In the following, we introduce two predominant differentiable rendering approaches as illustrated in Figure~\ref{fig::milestone}.


\textbf{Volume Rendering} imitates camera to sample the volume through ray-tracing actively. Rays cast from the viewpoint traverse the volume, where each ray represents a pixel, sampling points at discrete positions with parameterized color and opacity. Pixel color is obtained by alpha-blending all sampling points along the ray through opacity-weighted accumulation. Volumetric rendering achieves photorealistic results through physically grounded simulation of light transport, while maintaining compatibility with diverse camera models via parameterized projection matrices. The dual challenges of suboptimal sampling efficiency and the computational complexity of physically accurate material simulations impose a substantial computational cost that hinders real-time rendering.

\textbf{Rasterization} projects 3D geometric primitives into discrete pixel-level elements for 2D display. The pipeline projects 3D spatial representations onto the imaging plane, followed by pixel-aligned discretization. Rasterization enables real-time rendering with specific primitives, such as meshes or 3D Gaussians, making it suitable for latency-sensitive tasks. However, the simplified lighting models in rasterization restrict the capacity to model light transport, demonstrating compensated rendering fidelity for advanced materials such as specular surfaces and translucent materials.

\section{Datasets \& Metrics}\label{sec:data}

\subsection{Datasets}\label{sec:dataset}
To reconstruct 3D driving scenes, datasets are essential as they provide synchronized, multimodal inputs paired with rich annotations, such as 3D bounding boxes, semantic labels and HD map. These datasets can be broadly categorized into full-scene collections that capture diverse driving scenarios and object-centric collections that offer detailed modeling of individual traffic participants. Together, both types establish the foundational training and evaluation benchmark for holistic scene understanding and precise object reconstruction in autonomous driving.

\subsubsection{Object-centric Dataset}\label{sec:vehicle_dataset}
Object-centric datasets provide support for high-precision reconstruction of specific traffic participants. Based on the objects within the datasets, they can be categorized into vehicle datasets and human datasets.

\textbf{Vehicle datasets} provide multi-view observations of vehicles, offering essential priors for sparse-view reconstruction of vehicles. Multi-view Marketplace Car (MVMC)~\cite{zhang2021ners} contains approximately 6,000 images of 600 vehicle models, while CarPatch~\cite{liu2024car} collects 530,101 images of varying quality from the internet. PandaCar~\cite{neusim} and 3DRealCar~\cite{du20243drealcar} further include LiDAR point clouds, providing precise geometric information. Additionally, 3DRealCar~\cite{du20243drealcar} provides detailed vehicle component segmentation masks, laying a data foundation for future component-level reconstruction.

\textbf{Human Datasets}
To date, a dedicated pedestrian-specific 3D reconstruction dataset remains absent, and existing research relies on human digitization datasets. Series such as Human3.6M~\cite{h36m_pami}, CHI3D~\cite{human_interaction}, Fit3D~\cite{fit3d}, FlickrSC3D~\cite{sc3d}, HumanSC3D~\cite{sc3d}, and 3DPW~\cite{3dpw} provide indoor-captured human motion data, enabling posture reconstruction, but their predefined synthetic human geometry and lack of appearance details fail to meet 3D reconstruction requirements. RenderPeople~\cite{Renderpeople2025} offers high-quality, paid 3D human models with diverse genders, ethnicities, and clothing, some of which include motion and facial expressions, providing rich and detailed human data. However, real-world pedestrian characteristics differ significantly from these datasets, primarily due to severe occlusions, diverse behavioral patterns, and varying outdoor lighting conditions. All these differences jointly pose substantial challenges for pedestrian data collection.

\subsubsection{Full-scene collections} 
\begin{table*}[ht]
\caption{Overall comparison between full-scene data collections.}\label{tab:dataset}
\rowcolors{7}{lightgray}{white}
\centering
\resizebox{\textwidth}{!}{%
\begin{tabular}{ccccccccccccccccc}
\toprule
\multirow{2}{*}{Dataset} & \multirow{2}{*}{Year} & \multirow{2}{*}{Type} & \multicolumn{3}{c}{Modality}         & \multicolumn{2}{c}{Annotation} & \multicolumn{2}{c}{Volume} & \multirow{2}{*}{Reolution} & \multicolumn{2}{c}{Frame Rate} & \multirow{2}{*}{\begin{tabular}[c]{@{}c@{}}Sem.\\ Types\end{tabular}} & \multirow{2}{*}{\begin{tabular}[c]{@{}c@{}}Weather\\ Types\end{tabular}} & \multirow{2}{*}{\begin{tabular}[c]{@{}c@{}}Time \\ of Day\end{tabular}} & \multirow{2}{*}{\begin{tabular}[c]{@{}c@{}}Scenario\\ Types\end{tabular}} \\ \cmidrule(lr){4-6}\cmidrule(lr){7-8}\cmidrule(lr){9-10} \cmidrule(lr){12-13}
                         &                       &                       & Image      & Depth      & PCL        & BBox        & Semantic         & Scenes       & Image       &                            & Camera                              & LiDAR                              &                                                                           &                                                                          &                                                                         &                                                                           \\ \midrule
KITTI~\cite{Geiger2012kitti}                    & 2012                  & Real                  & \checkmark & \checkmark & \checkmark & 3D          & -                & 22           & 15K         & 1242$\times$376                   & 10                                  & 10                                 & 8                                                                         & 1                                                                        & 1                                                                       & 2                                                                         \\
CityScapes~\cite{Cordts2016Cityscapes}                  & 2016                  & Real                  & \checkmark & \checkmark          & -          & -          & \checkmark       & 50         & 25K         & 2048$\times$1024                   & -                                  & -                                  & 30                                                                        & 2                                                                        & 11                                                                       & 3                                                                         \\
BDD100K~\cite{yu2020bdd100k}                  & 2018                  & Real                  & \checkmark & -          & -          & 2D          & \checkmark       & 100k         & 12M         & 1280$\times$720                   & 30                                  & -                                  & 40                                                                        & 5                                                                        & 2                                                                       & 4                                                                         \\
SemanticKITTI~\cite{behley2019semankitti}            & 2019                  & Real                  & -          & -          & \checkmark & -           & \checkmark       & 22           & 43552       & -                          & -                                   & 10                                 & 28                                                                        & 1                                                                        & 1                                                                       & 4                                                                         \\
NuScenes~\cite{nuscenes2019}                 & 2019                  & Real                  & \checkmark & -          & \checkmark & 3D          & \checkmark       & 1000         & 1.4M        & 2048$\times$1536                  & 12                                  & 10                                 & 23                                                                        & 3                                                                        & 2                                                                       & 4                                                                         \\
Waymo~\cite{sun2020waymo}                     & 2019                  & Real                  & \checkmark & -          & \checkmark & 3D          & \checkmark       & 1150         & 390K        & 1920$\times$1080                  & 10                                  & 10                                 & 23                                                                        & 2                                                                        & 3                                                                       & 5                                                                         \\
VirtualKITTI2~\cite{cabon2020vkitti2}            & 2020                  & Syn.             & \checkmark & \checkmark & -          & 3D          & \checkmark       & 5            & 20992       & 1242$\times$375                   & \#                                   & -                                  & 8                                                                         & 3                                                                        & 2                                                                       & 5                                                                         \\
Argoverse2~\cite{Argoverse2}               & 2021                  & Real                  & \checkmark & -          & \checkmark & 3D          & \checkmark       & 1000         & 6M          & 2048$\times$1550                  & 20                                  & 10                                 & 30                                                                        & 2                                                                        & 1                                                                       & 1                                                                         \\
KITTI-360~\cite{Liao2022kitti360}                & 2022                  & Real                  & \checkmark & -          & \checkmark & 3D          & \checkmark       & 11           & 150K        & 1408$\times$376                          & \#                                   & \#                                  & 37                                                                        & 1                                                                        & 1                                                                       & 1                                                                         \\
NOTR~\cite{yang2023emernerf}                     & 2023                  & Real                  & \checkmark & -          & \checkmark & 3D          & \checkmark       & 120          & \#           & 1920$\times$1080                  & 10                                  & 10                                 & 23                                                                        & \#                                                                        & \#                                                                       & \#                                                                         \\\bottomrule
\end{tabular}%
}\vspace{0.3em}\\
\justifying\hfill\\
{\scriptsize {\bf All Column}: ``-''---Absence of corresponding modality; ``\#''---Not mentioned.}\\
{\scriptsize {\bf Type}: ``Real''---Collected in real world; ``\#''---Synthesized.}\\
{\scriptsize {\bf Annotation-BBox}: 3D---\textbf{3D} Bouding Box; 2D---\textbf{2D} Bounding Box}\\
\vspace{-2.0em}
\end{table*}

Full-scene collections are particularly valuable for autonomous driving as they capture the complete spatial and temporal context of driving environments, enabling models to understand complex inter-object relationships, scene dynamics, and the holistic geometric structure necessary for comprehensive autonomous driving perception. Full-scene collections can be categorized into multimodal datasets obtained by professional sensor suites and unimodal datasets collected from public sources. Detailed comparison among full-scene data collections is shown in Table~\ref{tab:dataset}.

Multimodal datasets provide rich sensor data, including images, calibration, and LiDAR point clouds, along with comprehensive annotations such as 2D/3D bounding boxes, semantic segmentation, and HD maps. Commonly used multimodal datasets include the KITTI series~\cite{Geiger2012kitti, behley2019semankitti, cabon2020virtual, cabon2020vkitti2, Liao2022kitti360}, Cityscapes~\cite{Cordts2016Cityscapes}, NuScenes~\cite{nuscenes2019}, Waymo Open Dataset~\cite{sun2020waymo}, and Argoverse~\cite{argoverse, Argoverse2}. Table~\ref{tab:dataset} provides a horizontal comparison of datasets across several key aspects, including data modalities, annotations, resolution, and diversity to highlight their distinct advantages. KITTI~\cite{Geiger2012kitti}, as the first driving scenario dataset, has been widely applied but is limited by lower data resolution and diversity, leading to the development of several derivative or iterative datasets. NuScenes~\cite{nuscenes2019} and Argoverse2~\cite{Argoverse2} offer surround-view perspectives, providing the most comprehensive image observations. Waymo~\cite{sun2020waymo} delivers high-quality LiDAR point clouds utilizing two high-precision LiDARs with an expanded field of view compared with NuScenes. NOTR~\cite{yang2023emernerf} is a subset selected from the Waymo Open Dataset containing 120 challenging driving scenes, enriched with additional annotations including 2D bounding boxes for dynamic objects, ground truth 3D scene flow, and 3D semantic occupancy. Unimodal datasets, primarily represented by BDD100K~\cite{yu2020bdd100k}, collect an unparalleled scale of driving scenario videos from the internet but are constrained by simple GPS location information and variable video quality.\vspace{-1.0em}

\subsection{Metrics}\label{sec:metric}
\subsubsection{Pixel-wise Metric}
Pixel-wise metrics evaluate reconstruction quality by directly comparing pixel values between generated and reference images using measures like Mean Squared Error (MSE), Mean Absolute Error (MAE), and Mean Absolute Percentage Error (MAPE). While computationally efficient, these metrics often fail to capture perceptual quality.

\subsubsection{Perceptual Metric}
Perceptual metrics evaluate reconstruction quality by modeling human visual perception rather than direct pixel comparisons. These metrics assess structural and textural similarity, including edge preservation and overall visual coherence. Typical perceptual metrics include Structural Similarity Index Measure (SSIM) for structural assessment, as well as feature-based metrics like Learned Perceptual Image Patch Similarity (LPIPS)~\cite{Zhang2018lpips} and VGG loss~\cite{vgg}, which operate in learned representation spaces. While these metrics correlate better with human quality judgments, they require greater computational resources than pixel-wise alternatives.

\subsubsection{Generative Metric}
Generative metrics evaluate the quality of synthetic content by assessing how closely generated data distributions align with real-world distributions. Key metrics include Inception Score (IS)~\cite{Barratt2018inceptionscore} for evaluating image quality and diversity, Fréchet Inception Distance (FID)~\cite{Heusel2017fid} for comparing feature distributions between generated and authentic images, Fréchet Video Distance (FVD) ~\cite{unterthiner2019fvd} for temporal content evaluation, and Maximum Mean Discrepancy (MMD)~\cite{mmd} for quantifying distribution divergence. These metrics specifically measure the statistical alignment between generated and reference data distributions.

\section{Traffic Elements Reconstruction}\label{sec:element}
The driving scenario comprises diverse complex elements that can be functionally categorized into static backgrounds and traffic agents. Their distinct characteristics, described in Section~\ref{sec:static} and \ref{sec:agent}, pose different challenges for 3D reconstruction. Table~\ref{tab:static_methods} exhibits detailed information and performance of static background reconstruction methods.

\begin{itemize}
    \item \textbf{Static background} Static background forms the environmental foundation of driving scene. Section~\ref{sec:static} reviews research on static background reconstruction, focusing on two main aspects: 1) improving geometric fidelity and achieving photorealistic rendering, 2) addressing large-scale optimization challenge.
    \item \textbf{Traffic agents} Traffic agents are the primary interactive elements in driving scenes. Section~\ref{sec:agent} surveys common methods for two main categories of traffic agents, rigid and non-rigid, and summarizes relevant studies based on their distinct characteristics.\vspace{-1.0em}
\end{itemize}

\subsection{Static Background Reconstruction}\label{sec:static}

\begin{table*}[ht]
\caption{Comparison among static background reconstruction methods.}\label{tab:static_methods}
\centering
\resizebox{\textwidth}{!}{%
\begin{tabular}{cccccccccccccccccccccc}
\toprule
 &  & \multicolumn{3}{c}{Input} &  &  & \multicolumn{6}{c}{Output} & \multicolumn{2}{c}{Training} &  & \multicolumn{5}{c}{Dataset} &  \\ \cmidrule(lr){3-5} \cmidrule(lr){8-15} \cmidrule(lr){17-21}
\multirow{-2}{*}{\begin{tabular}[c]{@{}c@{}}Scene\\ Repre.\end{tabular}} & \multirow{-2}{*}{Method} & \rotatebox{90}{Image} & \rotatebox{90}{Calibration} & \rotatebox{90}{Point Cloud} & \multirow{-2}{*}{\begin{tabular}[c]{@{}c@{}}Occl.\\ Iden.\end{tabular}} & \multirow{-2}{*}{\begin{tabular}[c]{@{}c@{}}Par.\\ Cri.\end{tabular}} & \rotatebox{90}{Image} & \rotatebox{90}{Depth} & \rotatebox{90}{Normal} & \rotatebox{90}{Point Cloud} & \rotatebox{90}{Semantic} & \rotatebox{90}{Mesh} & Devices & \multirow{-2}{*}{\begin{tabular}[c]{@{}c@{}}GPU\\ Hours\end{tabular}} & \multirow{-2}{*}{FPS} & Waymo & \rotatebox{90}{KITTI} & \rotatebox{90}{NuScenes} & \rotatebox{90}{KITTI-360} & \rotatebox{90}{Others} & \multirow{-2}{*}{\begin{tabular}[c]{@{}c@{}}Open\\ Source\end{tabular}} \\ \midrule
 & StreetSurf~\cite{guo2023streetsurf} & \checkmark & \checkmark & A &  & Dist. & \checkmark & \checkmark & \checkmark & \checkmark &  & \checkmark &  RTX3090 & 1.5 &  & 26.66 & \checkmark & \checkmark &  &  & \checkmark \\
 & \cellcolor[HTML]{EAEAEA}PlaNeRF~\cite{wang2023planerf} & \cellcolor[HTML]{EAEAEA}\checkmark & \cellcolor[HTML]{EAEAEA}\checkmark & \cellcolor[HTML]{EAEAEA}- & \cellcolor[HTML]{EAEAEA}Sem. & \cellcolor[HTML]{EAEAEA} & \cellcolor[HTML]{EAEAEA}\checkmark & \cellcolor[HTML]{EAEAEA}\checkmark & \cellcolor[HTML]{EAEAEA} & \cellcolor[HTML]{EAEAEA}\checkmark & \cellcolor[HTML]{EAEAEA}\checkmark & \cellcolor[HTML]{EAEAEA}\checkmark & \cellcolor[HTML]{EAEAEA} RTX3090 & \cellcolor[HTML]{EAEAEA}2.5 & \cellcolor[HTML]{EAEAEA} & \cellcolor[HTML]{EAEAEA} & \cellcolor[HTML]{EAEAEA} & \cellcolor[HTML]{EAEAEA} & \cellcolor[HTML]{EAEAEA}\checkmark & \cellcolor[HTML]{EAEAEA} & \cellcolor[HTML]{EAEAEA} \\
\multirow{-3}{*}{NeRF} & AlignMiF~\cite{tao2024alignmif} & \checkmark & \checkmark & I & Sem. &  & \checkmark & \checkmark &  & \checkmark &  &  &  RTX3090 & 2.5 &  & 29.78 &  &  & \checkmark & \cite{weng2021aiodrive} & \checkmark \\ \midrule
 & \cellcolor[HTML]{EAEAEA}Neural Point Light Field~\cite{ost2022neurallightfield} & \cellcolor[HTML]{EAEAEA}\checkmark & \cellcolor[HTML]{EAEAEA}\checkmark & \cellcolor[HTML]{EAEAEA}I & \cellcolor[HTML]{EAEAEA} & \cellcolor[HTML]{EAEAEA} & \cellcolor[HTML]{EAEAEA}\checkmark & \cellcolor[HTML]{EAEAEA}\checkmark & \cellcolor[HTML]{EAEAEA} & \cellcolor[HTML]{EAEAEA}\checkmark & \cellcolor[HTML]{EAEAEA} & \cellcolor[HTML]{EAEAEA} & \cellcolor[HTML]{EAEAEA} V100 & \cellcolor[HTML]{EAEAEA}48 & \cellcolor[HTML]{EAEAEA} & \cellcolor[HTML]{EAEAEA}31.25 & \cellcolor[HTML]{EAEAEA}\checkmark & \cellcolor[HTML]{EAEAEA}\checkmark & \cellcolor[HTML]{EAEAEA} & \cellcolor[HTML]{EAEAEA} & \cellcolor[HTML]{EAEAEA}\checkmark \\
\multirow{-2}{*}{PCL} & DGNR~\cite{li2024dgnr} & \checkmark & \checkmark & - &  & Spa. & \checkmark &  &  & \checkmark &  &  &  V100 &  & 16.67 &  & \checkmark & \checkmark &  & \cite{Cordts2016Cityscapes}\cite{argoverse} &  \\ \midrule
 & \cellcolor[HTML]{EAEAEA}DNMP~\cite{lu2023urban} & \cellcolor[HTML]{EAEAEA} & \cellcolor[HTML]{EAEAEA}\checkmark & \cellcolor[HTML]{EAEAEA}I & \cellcolor[HTML]{EAEAEA} & \cellcolor[HTML]{EAEAEA} & \cellcolor[HTML]{EAEAEA}\checkmark & \cellcolor[HTML]{EAEAEA}\checkmark & \cellcolor[HTML]{EAEAEA} & \cellcolor[HTML]{EAEAEA}\checkmark & \cellcolor[HTML]{EAEAEA} & \cellcolor[HTML]{EAEAEA}\checkmark & \cellcolor[HTML]{EAEAEA} A100 & \cellcolor[HTML]{EAEAEA} & \cellcolor[HTML]{EAEAEA} & \cellcolor[HTML]{EAEAEA}27.62 & \cellcolor[HTML]{EAEAEA} & \cellcolor[HTML]{EAEAEA} & \cellcolor[HTML]{EAEAEA}\checkmark & \cellcolor[HTML]{EAEAEA} & \cellcolor[HTML]{EAEAEA}\checkmark \\
\multirow{-2}{*}{Mesh} & RoMe~\cite{mei2024rome} & \checkmark & \checkmark & - & Sem. & Smpl. & \checkmark & \checkmark &  & \checkmark & \checkmark & \checkmark & RTX3090 & 2 &  &  & \checkmark & \checkmark &  &  & \checkmark \\
 & \cellcolor[HTML]{EAEAEA}EMIE-MAP~\cite{wu2024emie} & \cellcolor[HTML]{EAEAEA}\checkmark & \cellcolor[HTML]{EAEAEA}\checkmark & \cellcolor[HTML]{EAEAEA}S & \cellcolor[HTML]{EAEAEA} & \cellcolor[HTML]{EAEAEA} & \cellcolor[HTML]{EAEAEA}\checkmark & \cellcolor[HTML]{EAEAEA} & \cellcolor[HTML]{EAEAEA} & \cellcolor[HTML]{EAEAEA} & \cellcolor[HTML]{EAEAEA}\checkmark & \cellcolor[HTML]{EAEAEA}\checkmark & \cellcolor[HTML]{EAEAEA}A100 & \cellcolor[HTML]{EAEAEA} & \cellcolor[HTML]{EAEAEA} & \cellcolor[HTML]{EAEAEA} & \cellcolor[HTML]{EAEAEA}\checkmark & \cellcolor[HTML]{EAEAEA} & \cellcolor[HTML]{EAEAEA} & \cellcolor[HTML]{EAEAEA} & \cellcolor[HTML]{EAEAEA} \\ \midrule
& HGS-Mapping~\cite{wu2024hgs} & \checkmark & \checkmark & I &  &  & \checkmark & \checkmark & \checkmark & \checkmark &  & \checkmark & RTX3090 &  & 271 & 26.45 & \checkmark & \checkmark &  & \cite{cabon2020vkitti2} &  \\
& \cellcolor[HTML]{EAEAEA}StreetSurfGS~\cite{cui2024streetsurfgs} & \cellcolor[HTML]{EAEAEA}\checkmark & \cellcolor[HTML]{EAEAEA}\checkmark & \cellcolor[HTML]{EAEAEA}I & \cellcolor[HTML]{EAEAEA} & \cellcolor[HTML]{EAEAEA}Tem. & \cellcolor[HTML]{EAEAEA}\checkmark & \cellcolor[HTML]{EAEAEA}\checkmark & \cellcolor[HTML]{EAEAEA}\checkmark & \cellcolor[HTML]{EAEAEA} & \cellcolor[HTML]{EAEAEA} & \cellcolor[HTML]{EAEAEA}\checkmark & \cellcolor[HTML]{EAEAEA}RTX3090 & \cellcolor[HTML]{EAEAEA} & \cellcolor[HTML]{EAEAEA} & \cellcolor[HTML]{EAEAEA}28.67 & \cellcolor[HTML]{EAEAEA} & \cellcolor[HTML]{EAEAEA} & \cellcolor[HTML]{EAEAEA}\checkmark & \cellcolor[HTML]{EAEAEA}\cite{wang2023f2} & \cellcolor[HTML]{EAEAEA} \\
\multirow{-3}{*}{G.S.} & StreetUnveiler~\cite{xu2024unveiler} & \checkmark & \checkmark & I & Sem. &  & \checkmark & \checkmark & \checkmark &  &  & \checkmark &  &  &  &  &  &  &  & \cite{xiao2021pandaset} & \checkmark \\ \midrule
& \cellcolor[HTML]{EAEAEA}GVKF~\cite{song2024gvkf} & \cellcolor[HTML]{EAEAEA}\checkmark & \cellcolor[HTML]{EAEAEA}\checkmark & \cellcolor[HTML]{EAEAEA}I & \cellcolor[HTML]{EAEAEA} & \cellcolor[HTML]{EAEAEA} & \cellcolor[HTML]{EAEAEA}\checkmark & \cellcolor[HTML]{EAEAEA}\checkmark & \cellcolor[HTML]{EAEAEA}\checkmark & \cellcolor[HTML]{EAEAEA} & \cellcolor[HTML]{EAEAEA} & \cellcolor[HTML]{EAEAEA}\checkmark & \cellcolor[HTML]{EAEAEA} & \cellcolor[HTML]{EAEAEA}1.5 & \cellcolor[HTML]{EAEAEA}32 & \cellcolor[HTML]{EAEAEA}30.24 & \cellcolor[HTML]{EAEAEA} & \cellcolor[HTML]{EAEAEA} & \cellcolor[HTML]{EAEAEA} & \cellcolor[HTML]{EAEAEA} & \cellcolor[HTML]{EAEAEA}\checkmark \\
\multirow{-2}{*}{Hybrid} & DHGS~\cite{shi2024dhgs} & \checkmark & \checkmark & I & Sem. & Dist. & \checkmark & \checkmark &  &  & \checkmark &  &  &  &  & 28.09 &  &  &  &  &  \\\bottomrule
\end{tabular}%
}\vspace{0.3em}\\
\justifying\hfill\\
{\scriptsize {\bf Input-Point Cloud}: A---Optional \textbf{A}ugmentation; S---\textbf{S}upervision; I---\textbf{I}nput.}\\
{\scriptsize {\bf Occl. Iden.}: Occlusion identification. Sem.---\textbf{Sem}antic Segmentation; Uncer.---\textbf{Uncer}tainty.}\\
{\scriptsize {\bf Par. Cri.}: Spa.---\textbf{Spa}tial; Dist.---\textbf{Dist}ance to Ego; Smpl.---Waypoint \textbf{S}a\textbf{m}\textbf{pl}ing; Tem.---\textbf{Tem}poral Chunk.}\\
{\scriptsize {\bf Dataset}: Num.---PSNR of reconstruction; \checkmark---Evaluated on.}\\
{\scriptsize {\bf Open Source}:\checkmark---released to the public; $\star$---Recognized Method; U---\textbf{U}nofficial implementation.}\hfill\\
\vspace{-2.0em}
\end{table*}

Static background elements, including roads, buildings, vegetation, and traffic infrastructure, constitute the environmental foundation for autonomous driving systems. Reconstructing these elements presents dual challenges: achieving high fidelity reconstruction while managing the computational complexity of large-scale optimization. This section analyzes existing methods addressing these challenges and examines their practical applications in autonomous driving. 

\subsubsection{Fidelity}\label{sec:fidel}
High-fidelity static background reconstruction is fundamentally crucial to alleviate the sim-to-real gap in autonomous driving systems. It can provide a stable and accurate reference for downstream tasks that require precise capture of both spatial geometry and visual appearance of the driving environment. Reconstruction quality fundamentally depends on the choice of 3D representation, with different representations offering unique strengths and weaknesses.

\textbf{Geometry Fidelity} Novel learning-based representations, NeRF and 3D Gaussians, demonstrate significant advantages in visual quality through differentiable rendering, yet the geometric fidelity of these representations remains unsatisfactory.

NeRF~\cite{nerf} enables photorealistic rendering, but the lack of explicit geometric constraints results in limited geometric fidelity. PlaNeRF~\cite{wang2023planerf} introduces an SVD-based regularization term to enhance the geometric quality of road surfaces while StreetSurf~\cite{guo2023streetsurf} integrates SDF as geometry regularization. AlignMiF~\cite{tao2024alignmif} advances deep multimodal alignment research by systematically investigating LiDAR-camera misalignment issues when incorporating LiDAR data. AlignMiF proposes a geometry-aware alignment module to enhance geometry fidelity by utilizing a LiDAR prior.

3DGS struggles to maintain precise geometric reconstruction due to the fixed ellipsoidal shape of Gaussians, particularly for flat surfaces like roads or walls. \cite{wu2024hgs, cui2024streetsurfgs, shi2024dhgs, song2024gvkf} leverage distinct geometric priors of different scene elements, enabling the design of tailored Gaussian variants that enhance the quality of geometry and appearance reconstruction, such as spherical Gaussians for sky modeling~\cite{wu2024hgs}, planar 2D Gaussians for road surface~\cite{wu2024hgs, feng2024rogs}. To alleviate the distortion brought by discrete Gaussian primitives, DHGS~\cite{shi2024dhgs} further incorporates an SDF as auxiliary implicit surface representation to regularize the near-road reconstruction. Gaussian Voxel Kernel Function (GVKF)~\cite{song2024gvkf} combines 3D Gaussian with continuous surface modeling capability of implicit representation, providing a hybrid representation that offers a novel alternative for geometry-sensitive downstream applications.

\textbf{Photorealism} Traditional explicit representations, including point clouds and meshes, demonstrate impressive geometric flexibility in reconstruction, enabling accurate modeling of arbitrary geometry. However, their limited representation capacity fails to encode fine-grained visual details within a tolerable cost. Approaches integrate differentiable neural representations with explicit representations to achieve photorealistic rendering.

Neural point light field~\cite{ost2022neurallightfield} encodes the local light field into each point as neural features, enhancing the spatial capacity of each point. Combined with differentiable volumetric rendering for optimization, it achieves high-quality rendering results. DGNR~\cite{li2024dgnr} extracts feature points from the density field of NeRF, employing depth regularization to eliminate outliers and enhance geometric quality. Additionally, it designs a UNet architecture to decode the feature points into images, achieving improved reconstruction of both geometry and appearance.

RoMe~\cite{mei2024rome} and EMIE-MAP~\cite{wu2024emie} propose specialized neural mesh representations for road surface reconstruction. RoMe~\cite{mei2024rome} assigns each vertex fixed 2D coordinates, learnable elevation, RGB, and semantic features. However, direct optimization of RGB leads to convergence instability due to varying observation direction and lighting conditions. EMIE-MAP~\cite{wu2024emie} builds on RoMe by replacing direct RGB values with implicit RGB features, stabilizing the training process and improving reconstruction quality. In contrast, DNMP~\cite{lu2023urban} enables more flexible reconstruction by learning both the position and appearance of vertices for diverse static components like roads, vegetation, and buildings.

\subsubsection{Large-scale Optimization}\label{sec:largescale}

The static background in autonomous driving scenarios typically manifests as unbounded, long-range environments encompassing multi-scale elements with diverse material properties, creating significant challenges for high-quality reconstruction while maintaining computational efficiency. To address these large-scale optimization challenges, researchers have adopted scene decoupling strategies that can be categorized into two primary approaches based on their underlying principles.

Content-agnostic methods partition scenes using uniform criteria such as spatial location~\cite{li2024dgnr}, temporal data slices~\cite{cui2024streetsurfgs}, or data sampling points~\cite{mei2024rome}. These approaches divide environments into regular partitions and apply consistent spatial representations across all segments.

Conversely, content-aware methods leverage scene-specific characteristics by decoupling based on distance to the camera~\cite{guo2023streetsurf} or element types~\cite{shi2024dhgs}, such as distinguishing road surfaces from non-road elements. This decoupling method enables customized representations tailored to each segment's unique properties. For instance, StreetSurf~\cite{guo2023streetsurf} employs distance-based segmentation to create components with varying detail levels, allowing for optimal NeRF variants for each segment, while DHGS~\cite{shi2024dhgs} separates road surfaces from other elements, proposing specialized planar 2D Gaussians regularized by SDF for nearby road surfaces.\vspace{-1.0em}

\begin{figure}[t]
    \centering
    \includegraphics[width=0.9\linewidth]{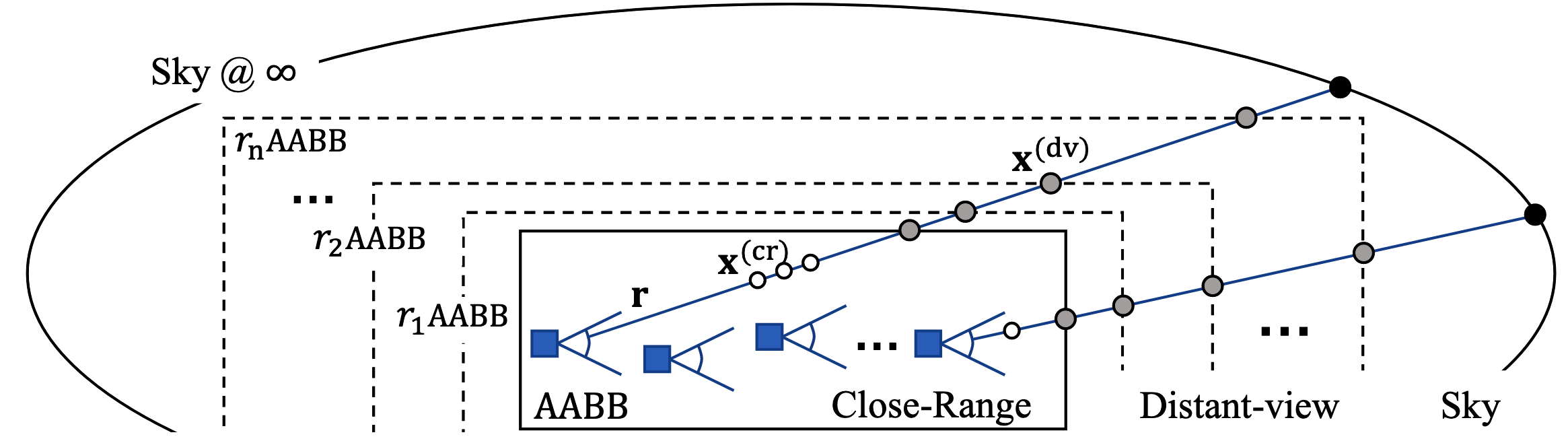}
    \caption{Distance-based Segmentation of StreetSurf~\cite{guo2023streetsurf}}
    \label{fig:streetsurf}\vspace{-1.0em}
\end{figure}

\subsection{Traffic Agents Reconstruction}\label{sec:agent}

While static background reconstruction establishes the environmental foundation, traffic agents, encompassing vehicles, cyclists, and pedestrians, represent the primary interactive elements that autonomous driving systems must accurately perceive and understand. High-quality reconstruction of these dynamic participants is critical for enabling essential capabilities such as object detection and tracking, behavior prediction, and collision avoidance. Traffic agents can be naturally categorized based on their deformation properties, which fundamentally determine the reconstruction approach required: rigid agents maintain a fixed shape during motion, allowing them to be reconstructed as a static object within their local coordinate, while non-rigid agents, including cyclists and pedestrians, exhibit complex articulated movements and shape variations that demand specialized deformation modeling techniques.

\subsubsection{Rigid Agents Reconstruction}\label{sec:rigid} Rigid agents are characterized by non-deformable and symmetric properties, which facilitate their reconstruction. 

Vehicles are typically modeled as rigid agents. Due to the significant color variations within the same model, independently modeling the geometry and appearance of vehicle is a practical approach. CADSim~\cite{wang2022cadsim} introduces predefined CAD models as geometry initialization priors to enhance fidelity and accelerate convergence, subsequently fine-tuning geometry and appearance through differentiable rendering. Car-Studio~\cite{liu2024car} incorporates separate shape and texture components with mip-NeRF architecture while NeuSim~\cite{neusim} leverages point cloud to construct a preliminary geometry using an SDF, then models RGB and shading features with NeRF. DreamCar employs a hierarchical NeRF framework comprising three specialized models~\cite{nerf, Wang2021NeuS, shen2021deep} to achieve coarse-to-fine geometry reconstruction and incorporates generative models to provide high-resolution textures.

Mirror symmetry is a key characteristic of vehicles, effectively mitigating the reconstruction challenges posed by sparse observations. NeuSim~\cite{neusim} applies mirror symmetry when constructing the geometric SDF from point clouds to complete the occluded side, and DreamCar~\cite{du2024dreamcar} employs mirror flipping during data processing to augment the dataset.

\begin{figure}[t]
    \centering
    \subfloat[Vehicles~\cite{GenAssets}]{\includegraphics[width=0.55\linewidth]{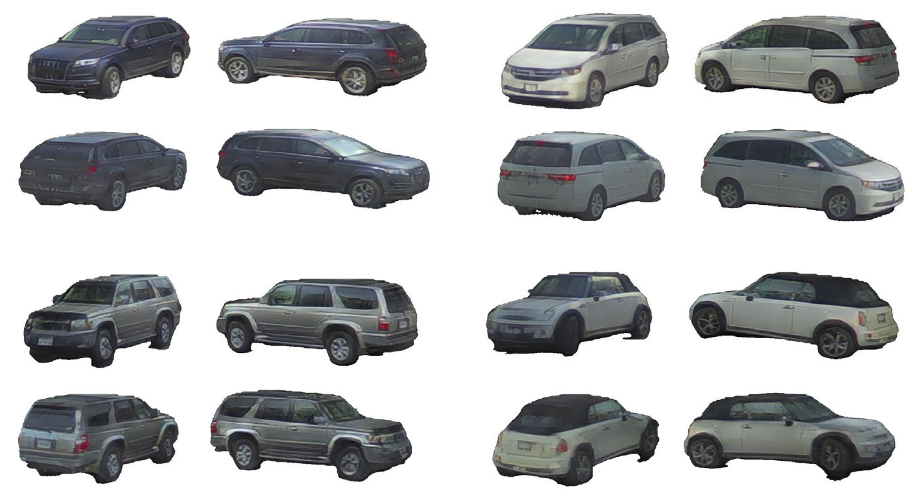}}
    \subfloat[Human~\cite{IDOL}]{\includegraphics[width=0.45\linewidth]{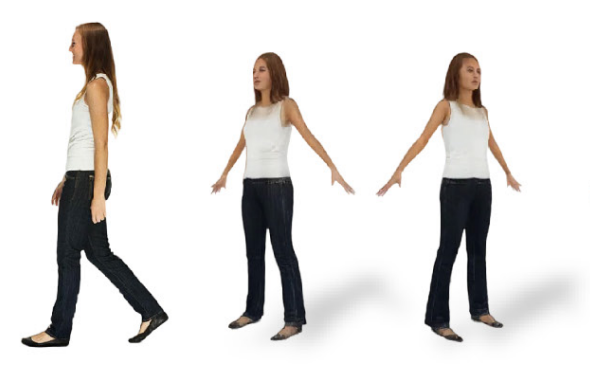}}
    \caption{3D reconstruction of traffic agents.}
    \label{fig:genassets}\vspace{-2.0em}
\end{figure}

While many 3D reconstruction methods exist, GenAssets~\cite{GenAssets} proposes a unified framework that combines reconstruction and generation of rigid 3D assets in a shared latent space, and can also approximate non-rigid traffic participants, such as cyclists and pedestrians, by modeling them as rigid objects (Fig~\ref{fig:genassets}). GenAssets encodes diverse agents into a low-dimensional latent distribution, thereby enabling high-quality 3D asset generation through a diffusion model.

\begin{table*}[ht]
\centering
\caption{Overall comparison among comprehensive scene reconstruction, generation methods and simulators.}\label{tab:comprehensive}
\resizebox{\textwidth}{!}{%
\begin{tabular}{ccccccccccccccccccccccc}
\toprule
 &  &  & \multicolumn{3}{c}{Input} &  & \multicolumn{7}{c}{Rendering} &  &  & \multicolumn{6}{c}{Dataset} &  \\ \cmidrule(lr){4-6} \cmidrule(lr){8-14} \cmidrule(lr){17-22}
\multirow{-2}{*}{Task} & \multirow{-2}{*}{Repre.} & \multirow{-2}{*}{Method} & \rotatebox{90}{Image} & \rotatebox{90}{Pose} & \rotatebox{90}{Point Cloud } & \multirow{-2}{*}{Decomp.} & \rotatebox{90}{Image} & \rotatebox{90}{Depth} & \rotatebox{90}{Point Cloud } & \rotatebox{90}{Veh. Pose} & \rotatebox{90}{Instance} & \rotatebox{90}{Semantic} & \rotatebox{90}{Flow} & \multirow{-2}{*}{Devices} & \multirow{-2}{*}{FPS} & KITTI & Waymo & \rotatebox{90}{NuScenes} & \rotatebox{90}{vKITTI2} & \rotatebox{90}{KITTI-360} & \rotatebox{90}{Others} & \multirow{-2}{*}{\begin{tabular}[c]{@{}c@{}}Open\\ Source\end{tabular}} \\ \midrule
 & PCL & \cellcolor[HTML]{EAEAEA}READ~\cite{li2023read} & \cellcolor[HTML]{EAEAEA}\checkmark & \cellcolor[HTML]{EAEAEA}\checkmark & \cellcolor[HTML]{EAEAEA}Input & \cellcolor[HTML]{EAEAEA}BBox & \cellcolor[HTML]{EAEAEA}\checkmark & \cellcolor[HTML]{EAEAEA} & \cellcolor[HTML]{EAEAEA} & \cellcolor[HTML]{EAEAEA} & \cellcolor[HTML]{EAEAEA} & \cellcolor[HTML]{EAEAEA} & \cellcolor[HTML]{EAEAEA} & \cellcolor[HTML]{EAEAEA}RTX2070 & \cellcolor[HTML]{EAEAEA} & \cellcolor[HTML]{EAEAEA}23.286 & \cellcolor[HTML]{EAEAEA} & \cellcolor[HTML]{EAEAEA} & \cellcolor[HTML]{EAEAEA} & \cellcolor[HTML]{EAEAEA} & \cellcolor[HTML]{EAEAEA}\cite{ligocki2020brno} & \cellcolor[HTML]{EAEAEA}$\star$ \\ \cmidrule(l){2-23} 
 &  & NSG~\cite{ost2021nsg} & \checkmark & \checkmark & - & BBox & \checkmark &  &  &  &  &  &  & RTX6000 & 0.032 & 26.66 &  &  & \checkmark &  &  & \checkmark \\
 &  & \cellcolor[HTML]{EAEAEA}PNF~\cite{kundu2022panoptic} & \cellcolor[HTML]{EAEAEA}\checkmark & \cellcolor[HTML]{EAEAEA}- & \cellcolor[HTML]{EAEAEA}- & \cellcolor[HTML]{EAEAEA}Self & \cellcolor[HTML]{EAEAEA}\checkmark & \cellcolor[HTML]{EAEAEA}\checkmark & \cellcolor[HTML]{EAEAEA} & \cellcolor[HTML]{EAEAEA}\checkmark & \cellcolor[HTML]{EAEAEA}\checkmark & \cellcolor[HTML]{EAEAEA}\checkmark & \cellcolor[HTML]{EAEAEA} & \cellcolor[HTML]{EAEAEA} & \cellcolor[HTML]{EAEAEA} & \cellcolor[HTML]{EAEAEA}27.48 & \cellcolor[HTML]{EAEAEA} & \cellcolor[HTML]{EAEAEA} & \cellcolor[HTML]{EAEAEA}\checkmark & \cellcolor[HTML]{EAEAEA}\checkmark & \cellcolor[HTML]{EAEAEA} & \cellcolor[HTML]{EAEAEA} \\
 &  & NeuRAD~\cite{tonderski2024neurad} & \checkmark & \checkmark & SV. & BBox & \checkmark & \checkmark & \checkmark &  &  &  &  &  &  & 27.91 &  & \checkmark &  &  & \cite{Argoverse2}\cite{zod} & $\star$ \\
 &  & \cellcolor[HTML]{EAEAEA}SUDS~\cite{turki2023suds} & \cellcolor[HTML]{EAEAEA}\checkmark & \cellcolor[HTML]{EAEAEA}- & \cellcolor[HTML]{EAEAEA}SV. & \cellcolor[HTML]{EAEAEA}Self & \cellcolor[HTML]{EAEAEA}\checkmark & \cellcolor[HTML]{EAEAEA}\checkmark & \cellcolor[HTML]{EAEAEA} & \cellcolor[HTML]{EAEAEA}\checkmark & \cellcolor[HTML]{EAEAEA}\checkmark & \cellcolor[HTML]{EAEAEA}\checkmark & \cellcolor[HTML]{EAEAEA}S & \cellcolor[HTML]{EAEAEA}RTX6000 & \cellcolor[HTML]{EAEAEA}0.01 & \cellcolor[HTML]{EAEAEA}28.31 & \cellcolor[HTML]{EAEAEA} & \cellcolor[HTML]{EAEAEA} & \cellcolor[HTML]{EAEAEA}\checkmark & \cellcolor[HTML]{EAEAEA} & \cellcolor[HTML]{EAEAEA} & \cellcolor[HTML]{EAEAEA}\checkmark \\
 &  & EmerNeRF~\cite{yang2023emernerf} & \checkmark & \checkmark & SV. & Self & \checkmark & \checkmark & \checkmark &  &  &  & S & RTX6000 & 0.053 &  & 28.87 &  &  &  &  & $\star$ \\
 &  & \cellcolor[HTML]{EAEAEA}ProSGNeRF~\cite{deng2023prosgnerf} & \cellcolor[HTML]{EAEAEA}\checkmark & \cellcolor[HTML]{EAEAEA}\checkmark & \cellcolor[HTML]{EAEAEA}SV. & \cellcolor[HTML]{EAEAEA}BBox & \cellcolor[HTML]{EAEAEA}\checkmark & \cellcolor[HTML]{EAEAEA}\checkmark & \cellcolor[HTML]{EAEAEA} & \cellcolor[HTML]{EAEAEA} & \cellcolor[HTML]{EAEAEA} & \cellcolor[HTML]{EAEAEA} & \cellcolor[HTML]{EAEAEA} & \cellcolor[HTML]{EAEAEA} & \cellcolor[HTML]{EAEAEA} & \cellcolor[HTML]{EAEAEA}30.31 & \cellcolor[HTML]{EAEAEA} & \cellcolor[HTML]{EAEAEA} & \cellcolor[HTML]{EAEAEA}\checkmark & \cellcolor[HTML]{EAEAEA} & \cellcolor[HTML]{EAEAEA} & \cellcolor[HTML]{EAEAEA} \\
 &  & DiCo-NeRF~\cite{choi2024dico} & \checkmark & \checkmark & - & Self & \checkmark &  &  &  &  &  &  &  &  &  &  &  &  & \checkmark & ~\cite{jbnu} & S \\
 &  & \cellcolor[HTML]{EAEAEA}S-NeRF~\cite{xie2023s} & \cellcolor[HTML]{EAEAEA}\checkmark & \cellcolor[HTML]{EAEAEA}\checkmark & \cellcolor[HTML]{EAEAEA}SV. & \cellcolor[HTML]{EAEAEA}BBox & \cellcolor[HTML]{EAEAEA}\checkmark & \cellcolor[HTML]{EAEAEA}\checkmark & \cellcolor[HTML]{EAEAEA} & \cellcolor[HTML]{EAEAEA} & \cellcolor[HTML]{EAEAEA} & \cellcolor[HTML]{EAEAEA} & \cellcolor[HTML]{EAEAEA}O & \cellcolor[HTML]{EAEAEA}RTX6000 & \cellcolor[HTML]{EAEAEA}0.02 & \cellcolor[HTML]{EAEAEA} & \cellcolor[HTML]{EAEAEA}23.60 & \cellcolor[HTML]{EAEAEA}\checkmark & \cellcolor[HTML]{EAEAEA} & \cellcolor[HTML]{EAEAEA} & \cellcolor[HTML]{EAEAEA} & \cellcolor[HTML]{EAEAEA}\checkmark \\
 & \multirow{-9}{*}{NeRF} & S-NeRF++~\cite{chen2025s} & \checkmark & \checkmark & SV. & BBox & \checkmark & \checkmark &  &  &  & \checkmark & O &  &  &  & 25.78 & \checkmark &  &  &  &  \\ \cmidrule(l){2-23} 
 &  & \cellcolor[HTML]{EAEAEA}DrivingGaussian~\cite{zhou2024drivinggaussian} & \cellcolor[HTML]{EAEAEA}\checkmark & \cellcolor[HTML]{EAEAEA}\checkmark & \cellcolor[HTML]{EAEAEA}AUG. & \cellcolor[HTML]{EAEAEA}BBox & \cellcolor[HTML]{EAEAEA}\checkmark & \cellcolor[HTML]{EAEAEA} & \cellcolor[HTML]{EAEAEA} & \cellcolor[HTML]{EAEAEA} & \cellcolor[HTML]{EAEAEA} & \cellcolor[HTML]{EAEAEA} & \cellcolor[HTML]{EAEAEA} & \cellcolor[HTML]{EAEAEA} & \cellcolor[HTML]{EAEAEA} & \cellcolor[HTML]{EAEAEA} & \cellcolor[HTML]{EAEAEA} & \cellcolor[HTML]{EAEAEA}\checkmark & \cellcolor[HTML]{EAEAEA} & \cellcolor[HTML]{EAEAEA}\checkmark & \cellcolor[HTML]{EAEAEA} & \cellcolor[HTML]{EAEAEA}\checkmark \\
 &  & GGRt~\cite{li2024ggrt} & \checkmark & - & - & - & \checkmark & \checkmark &  &  &  &  &  &  &  & 22.59 & 32.12 &  &  &  &  & \checkmark \\
 &  & \cellcolor[HTML]{EAEAEA}TCLC-GS~\cite{zhao2024tclc} & \cellcolor[HTML]{EAEAEA}\checkmark & \cellcolor[HTML]{EAEAEA}\checkmark & \cellcolor[HTML]{EAEAEA}Input & \cellcolor[HTML]{EAEAEA}- & \cellcolor[HTML]{EAEAEA}\checkmark & \cellcolor[HTML]{EAEAEA}\checkmark & \cellcolor[HTML]{EAEAEA} & \cellcolor[HTML]{EAEAEA} & \cellcolor[HTML]{EAEAEA} & \cellcolor[HTML]{EAEAEA} & \cellcolor[HTML]{EAEAEA} & \cellcolor[HTML]{EAEAEA}RTX3090ti & \cellcolor[HTML]{EAEAEA}90 & \cellcolor[HTML]{EAEAEA} & \cellcolor[HTML]{EAEAEA}28.11 & \cellcolor[HTML]{EAEAEA}\checkmark & \cellcolor[HTML]{EAEAEA} & \cellcolor[HTML]{EAEAEA} & \cellcolor[HTML]{EAEAEA} & \cellcolor[HTML]{EAEAEA} \\
 &  & SGD~\cite{yu2024sgd} & \checkmark & \checkmark & Input & - & \checkmark & \checkmark &  &  &  &  &  &  &  & 23.85* &  &  &  & \checkmark &  &  \\
 &  & \cellcolor[HTML]{EAEAEA}HO-Gaussian~\cite{li2024ho} & \cellcolor[HTML]{EAEAEA}\checkmark & \cellcolor[HTML]{EAEAEA}\checkmark & \cellcolor[HTML]{EAEAEA}- & \cellcolor[HTML]{EAEAEA}- & \cellcolor[HTML]{EAEAEA}\checkmark & \cellcolor[HTML]{EAEAEA} & \cellcolor[HTML]{EAEAEA} & \cellcolor[HTML]{EAEAEA} & \cellcolor[HTML]{EAEAEA} & \cellcolor[HTML]{EAEAEA} & \cellcolor[HTML]{EAEAEA} & \cellcolor[HTML]{EAEAEA}V100 & \cellcolor[HTML]{EAEAEA}71 & \cellcolor[HTML]{EAEAEA} & \cellcolor[HTML]{EAEAEA}28.03 & \cellcolor[HTML]{EAEAEA} & \cellcolor[HTML]{EAEAEA} & \cellcolor[HTML]{EAEAEA} & \cellcolor[HTML]{EAEAEA}\cite{Argoverse2} & \cellcolor[HTML]{EAEAEA} \\
 &  & $S^3$Gaussian~\cite{huang2024s3} & \checkmark & \checkmark & Input & Self & \checkmark & \checkmark &  &  &  & \checkmark &  &  &  &  & 32.14 &  &  &  & \cite{yang2023emernerf} & \checkmark \\
 &  & \cellcolor[HTML]{EAEAEA}PVG~\cite{chen2023periodic} & \cellcolor[HTML]{EAEAEA}\checkmark & \cellcolor[HTML]{EAEAEA}\checkmark & \cellcolor[HTML]{EAEAEA}Input & \cellcolor[HTML]{EAEAEA}Self & \cellcolor[HTML]{EAEAEA}\checkmark & \cellcolor[HTML]{EAEAEA}\checkmark & \cellcolor[HTML]{EAEAEA} & \cellcolor[HTML]{EAEAEA} & \cellcolor[HTML]{EAEAEA} & \cellcolor[HTML]{EAEAEA}\checkmark & \cellcolor[HTML]{EAEAEA}V & \cellcolor[HTML]{EAEAEA}RTX6000 & \cellcolor[HTML]{EAEAEA}59 & \cellcolor[HTML]{EAEAEA}32.83 & \cellcolor[HTML]{EAEAEA}32.46 & \cellcolor[HTML]{EAEAEA} & \cellcolor[HTML]{EAEAEA} & \cellcolor[HTML]{EAEAEA} & \cellcolor[HTML]{EAEAEA} & \cellcolor[HTML]{EAEAEA}\checkmark \\
 &  & Street Gaussians~\cite{yan2024streetgaussian} & \checkmark & - & Input & BBox & \checkmark & \checkmark &  & \checkmark &  & \checkmark &  & RTX4090 & 135 & 25.79* & 34.61 &  &  &  &  & $\star$ \\
 &  & \cellcolor[HTML]{EAEAEA}OmniRe~\cite{chen2024omnire} & \cellcolor[HTML]{EAEAEA}\checkmark & \cellcolor[HTML]{EAEAEA}\checkmark & \cellcolor[HTML]{EAEAEA}Input & \cellcolor[HTML]{EAEAEA}BBox & \cellcolor[HTML]{EAEAEA}\checkmark & \cellcolor[HTML]{EAEAEA}\checkmark & \cellcolor[HTML]{EAEAEA} & \cellcolor[HTML]{EAEAEA} & \cellcolor[HTML]{EAEAEA} & \cellcolor[HTML]{EAEAEA} & \cellcolor[HTML]{EAEAEA} & \cellcolor[HTML]{EAEAEA}RTX4090 & \cellcolor[HTML]{EAEAEA}60 & \cellcolor[HTML]{EAEAEA} & \cellcolor[HTML]{EAEAEA}34.25 & \cellcolor[HTML]{EAEAEA} & \cellcolor[HTML]{EAEAEA} & \cellcolor[HTML]{EAEAEA} & \cellcolor[HTML]{EAEAEA} & \cellcolor[HTML]{EAEAEA}$\star$ \\
 &  & DeSiRe-GS~\cite{peng2024desire} & \checkmark & \checkmark & Input & Self & \checkmark & \checkmark &  &  &  &  & V & RTX4090 & 36 & 33.94 & 33.61 &  &  &  &  & \checkmark \\
 &  & \cellcolor[HTML]{EAEAEA}VDG~\cite{li2024vdg} & \cellcolor[HTML]{EAEAEA}\checkmark & \cellcolor[HTML]{EAEAEA}- & \cellcolor[HTML]{EAEAEA}- & \cellcolor[HTML]{EAEAEA}Self & \cellcolor[HTML]{EAEAEA}\checkmark & \cellcolor[HTML]{EAEAEA}\checkmark & \cellcolor[HTML]{EAEAEA} & \cellcolor[HTML]{EAEAEA} & \cellcolor[HTML]{EAEAEA} & \cellcolor[HTML]{EAEAEA} & \cellcolor[HTML]{EAEAEA} & \cellcolor[HTML]{EAEAEA}V100 & \cellcolor[HTML]{EAEAEA}61 & \cellcolor[HTML]{EAEAEA}31.61 & \cellcolor[HTML]{EAEAEA}31.65 & \cellcolor[HTML]{EAEAEA} & \cellcolor[HTML]{EAEAEA} & \cellcolor[HTML]{EAEAEA} & \cellcolor[HTML]{EAEAEA} & \cellcolor[HTML]{EAEAEA}S \\
 &  & HUGS~\cite{zhou2024hugs} & \checkmark & \checkmark & Input & BBox & \checkmark & \checkmark &  &  & \checkmark & \checkmark & O & RTX4090 & 93 & 28.78 &  &  & \checkmark & \checkmark &  & \checkmark \\
 &  & \cellcolor[HTML]{EAEAEA}VEGS~\cite{hwang2024vegs} & \cellcolor[HTML]{EAEAEA}\checkmark & \cellcolor[HTML]{EAEAEA}\checkmark & \cellcolor[HTML]{EAEAEA}Input & \cellcolor[HTML]{EAEAEA}BBox & \cellcolor[HTML]{EAEAEA}\checkmark & \cellcolor[HTML]{EAEAEA}\checkmark & \cellcolor[HTML]{EAEAEA} & \cellcolor[HTML]{EAEAEA} & \cellcolor[HTML]{EAEAEA} & \cellcolor[HTML]{EAEAEA} & \cellcolor[HTML]{EAEAEA} & \cellcolor[HTML]{EAEAEA}RTX3090 & \cellcolor[HTML]{EAEAEA}144 & \cellcolor[HTML]{EAEAEA}24.77 & \cellcolor[HTML]{EAEAEA} & \cellcolor[HTML]{EAEAEA} & \cellcolor[HTML]{EAEAEA} & \cellcolor[HTML]{EAEAEA}\checkmark & \cellcolor[HTML]{EAEAEA} & \cellcolor[HTML]{EAEAEA}\checkmark \\
 &  & AutoSplat~\cite{khan2024autosplat} & \checkmark & \checkmark & Input & BBox & \checkmark &  &  &  &  &  &  &  &  & 26.59* &  &  &  &  & \cite{xiao2021pandaset} &  \\
 &  & \cellcolor[HTML]{EAEAEA}GGS~\cite{han2024ggs} & \cellcolor[HTML]{EAEAEA}\checkmark & \cellcolor[HTML]{EAEAEA}\checkmark & \cellcolor[HTML]{EAEAEA}Input & \cellcolor[HTML]{EAEAEA}- & \cellcolor[HTML]{EAEAEA}\checkmark & \cellcolor[HTML]{EAEAEA}\checkmark & \cellcolor[HTML]{EAEAEA} & \cellcolor[HTML]{EAEAEA} & \cellcolor[HTML]{EAEAEA} & \cellcolor[HTML]{EAEAEA} & \cellcolor[HTML]{EAEAEA} & \cellcolor[HTML]{EAEAEA} & \cellcolor[HTML]{EAEAEA} & \cellcolor[HTML]{EAEAEA}29.12 & \cellcolor[HTML]{EAEAEA} & \cellcolor[HTML]{EAEAEA} & \cellcolor[HTML]{EAEAEA} & \cellcolor[HTML]{EAEAEA} & \cellcolor[HTML]{EAEAEA}\cite{ligocki2020brno} & \cellcolor[HTML]{EAEAEA} \\
 &  & DriveDreamer4D~\cite{zhao2024drivedreamer4d} & \checkmark & \checkmark & SV. & BBox & \checkmark & \checkmark &  &  &  &  &  &  &  &  &  &  &  &  &  & \checkmark \\
 &  & \cellcolor[HTML]{EAEAEA}DrivingForward~\cite{tian2024drivingforward} & \cellcolor[HTML]{EAEAEA}\checkmark & \cellcolor[HTML]{EAEAEA}- & \cellcolor[HTML]{EAEAEA}- & \cellcolor[HTML]{EAEAEA}Self & \cellcolor[HTML]{EAEAEA}\checkmark & \cellcolor[HTML]{EAEAEA}\checkmark & \cellcolor[HTML]{EAEAEA} & \cellcolor[HTML]{EAEAEA} & \cellcolor[HTML]{EAEAEA} & \cellcolor[HTML]{EAEAEA} & \cellcolor[HTML]{EAEAEA} & \cellcolor[HTML]{EAEAEA} & \cellcolor[HTML]{EAEAEA} & \cellcolor[HTML]{EAEAEA} & \cellcolor[HTML]{EAEAEA} & \cellcolor[HTML]{EAEAEA}\checkmark & \cellcolor[HTML]{EAEAEA} & \cellcolor[HTML]{EAEAEA} & \cellcolor[HTML]{EAEAEA} & \cellcolor[HTML]{EAEAEA}\checkmark \\
 &  & SplatAD~\cite{hess2024splatad} & \checkmark & \checkmark & Input & BBox & \checkmark & \checkmark & \checkmark &  &  &  &  &  &  &  &  & \checkmark &  &  & \cite{Argoverse2}\cite{xiao2021pandaset} & S \\
 &  & \cellcolor[HTML]{EAEAEA}DreamDrive~\cite{mao2024dreamdrive} & \cellcolor[HTML]{EAEAEA}\checkmark & \cellcolor[HTML]{EAEAEA}\checkmark & \cellcolor[HTML]{EAEAEA}Input & \cellcolor[HTML]{EAEAEA}- & \cellcolor[HTML]{EAEAEA}\checkmark & \cellcolor[HTML]{EAEAEA} & \cellcolor[HTML]{EAEAEA} & \cellcolor[HTML]{EAEAEA} & \cellcolor[HTML]{EAEAEA} & \cellcolor[HTML]{EAEAEA} & \cellcolor[HTML]{EAEAEA} & \cellcolor[HTML]{EAEAEA} & \cellcolor[HTML]{EAEAEA} & \cellcolor[HTML]{EAEAEA} & \cellcolor[HTML]{EAEAEA} & \cellcolor[HTML]{EAEAEA}\checkmark & \cellcolor[HTML]{EAEAEA} & \cellcolor[HTML]{EAEAEA} & \cellcolor[HTML]{EAEAEA} & \cellcolor[HTML]{EAEAEA} \\
 &  & STORM~\cite{yang2024storm} & \checkmark & \checkmark & SV. & Self & \checkmark & \checkmark &  &  &  &  & S &  &  &  & 26.38 &  &  &  & \cite{Argoverse2} & $\star$ \\
 &  & \cellcolor[HTML]{EAEAEA}EMD~\cite{wei2024emd} & \cellcolor[HTML]{EAEAEA}\checkmark & \cellcolor[HTML]{EAEAEA}\checkmark & \cellcolor[HTML]{EAEAEA}Input & \cellcolor[HTML]{EAEAEA}B/S & \cellcolor[HTML]{EAEAEA}\checkmark & \cellcolor[HTML]{EAEAEA} & \cellcolor[HTML]{EAEAEA} & \cellcolor[HTML]{EAEAEA} & \cellcolor[HTML]{EAEAEA} & \cellcolor[HTML]{EAEAEA} & \cellcolor[HTML]{EAEAEA}S & \cellcolor[HTML]{EAEAEA} & \cellcolor[HTML]{EAEAEA} & \cellcolor[HTML]{EAEAEA} & \cellcolor[HTML]{EAEAEA}32.5 & \cellcolor[HTML]{EAEAEA} & \cellcolor[HTML]{EAEAEA} & \cellcolor[HTML]{EAEAEA} & \cellcolor[HTML]{EAEAEA} & \cellcolor[HTML]{EAEAEA}S \\
 &  & Omni-Scene~\cite{omniscene} & \checkmark & \checkmark & - & - & \checkmark & \checkmark &  &  &  &  &  &  &  &  &  & \checkmark &  &  &  & \checkmark \\
\multirow{-33}{*}{Recon.} & \multirow{-23}{*}{G.S.} & \cellcolor[HTML]{EAEAEA}Uni-Gaussians~\cite{yuan2025uni} & \cellcolor[HTML]{EAEAEA}\checkmark & \cellcolor[HTML]{EAEAEA}\checkmark & \cellcolor[HTML]{EAEAEA}Input & \cellcolor[HTML]{EAEAEA}BBox & \cellcolor[HTML]{EAEAEA}\checkmark & \cellcolor[HTML]{EAEAEA}\checkmark & \cellcolor[HTML]{EAEAEA}\checkmark & \cellcolor[HTML]{EAEAEA} & \cellcolor[HTML]{EAEAEA} & \cellcolor[HTML]{EAEAEA} & \cellcolor[HTML]{EAEAEA} & \cellcolor[HTML]{EAEAEA} & \cellcolor[HTML]{EAEAEA} & \cellcolor[HTML]{EAEAEA} & \cellcolor[HTML]{EAEAEA}29.62 & \cellcolor[HTML]{EAEAEA} & \cellcolor[HTML]{EAEAEA} & \cellcolor[HTML]{EAEAEA} & \cellcolor[HTML]{EAEAEA} & \cellcolor[HTML]{EAEAEA} \\ \midrule
 &  & UniScene~\cite{li2024uniscene} & - & - & - & BBox & \checkmark & \checkmark & \checkmark &  &  & \checkmark &  &  &  &  &  & \checkmark &  &  &  & S \\
\multirow{-2}{*}{Gen.} & \multirow{-2}{*}{G.S.} & \cellcolor[HTML]{EAEAEA}MagicDrive3D~\cite{gao2024magicdrive3d} & \cellcolor[HTML]{EAEAEA}- & \cellcolor[HTML]{EAEAEA}- & \cellcolor[HTML]{EAEAEA}- & \cellcolor[HTML]{EAEAEA}BBox & \cellcolor[HTML]{EAEAEA}\checkmark & \cellcolor[HTML]{EAEAEA}\checkmark & \cellcolor[HTML]{EAEAEA} & \cellcolor[HTML]{EAEAEA}\checkmark & \cellcolor[HTML]{EAEAEA} & \cellcolor[HTML]{EAEAEA} & \cellcolor[HTML]{EAEAEA} & \cellcolor[HTML]{EAEAEA} & \cellcolor[HTML]{EAEAEA} & \cellcolor[HTML]{EAEAEA} & \cellcolor[HTML]{EAEAEA} & \cellcolor[HTML]{EAEAEA}\checkmark & \cellcolor[HTML]{EAEAEA} & \cellcolor[HTML]{EAEAEA} & \cellcolor[HTML]{EAEAEA} & \cellcolor[HTML]{EAEAEA}S \\ \midrule
 & NeRF & UniSim~\cite{yang2023unisim} & \checkmark & \checkmark & Input & BBox & \checkmark &  & \checkmark & \checkmark &  &  &  &  &  &  &  &  &  &  & \cite{xiao2021pandaset} &  \\
 & NeRF & \cellcolor[HTML]{EAEAEA}MARS~\cite{wu2023mars} & \cellcolor[HTML]{EAEAEA}\checkmark & \cellcolor[HTML]{EAEAEA}\checkmark & \cellcolor[HTML]{EAEAEA}- & \cellcolor[HTML]{EAEAEA}BBox & \cellcolor[HTML]{EAEAEA}\checkmark & \cellcolor[HTML]{EAEAEA}\checkmark & \cellcolor[HTML]{EAEAEA} & \cellcolor[HTML]{EAEAEA} & \cellcolor[HTML]{EAEAEA} & \cellcolor[HTML]{EAEAEA}\checkmark & \cellcolor[HTML]{EAEAEA} & \cellcolor[HTML]{EAEAEA}RTX6000 & \cellcolor[HTML]{EAEAEA}0.03 & \cellcolor[HTML]{EAEAEA}29.06 & \cellcolor[HTML]{EAEAEA} & \cellcolor[HTML]{EAEAEA} & \cellcolor[HTML]{EAEAEA}\checkmark & \cellcolor[HTML]{EAEAEA} & \cellcolor[HTML]{EAEAEA} & \cellcolor[HTML]{EAEAEA}$\star$ \\
 & NeRF & NeuroNCap~\cite{ljungbergh2024neuroncap} & \checkmark & \checkmark & - & BBox & \checkmark &  &  & \checkmark &  &  &  &  &  &  &  & \checkmark &  &  &  & \checkmark \\
 & NeRF & \cellcolor[HTML]{EAEAEA}OASim~\cite{yan2024oasim} & \cellcolor[HTML]{EAEAEA}\checkmark & \cellcolor[HTML]{EAEAEA}\checkmark & \cellcolor[HTML]{EAEAEA}AUG. & \cellcolor[HTML]{EAEAEA}BBox & \cellcolor[HTML]{EAEAEA}\checkmark & \cellcolor[HTML]{EAEAEA} & \cellcolor[HTML]{EAEAEA}\checkmark & \cellcolor[HTML]{EAEAEA}\checkmark & \cellcolor[HTML]{EAEAEA} & \cellcolor[HTML]{EAEAEA} & \cellcolor[HTML]{EAEAEA} & \cellcolor[HTML]{EAEAEA} & \cellcolor[HTML]{EAEAEA} & \cellcolor[HTML]{EAEAEA} & \cellcolor[HTML]{EAEAEA} & \cellcolor[HTML]{EAEAEA}\checkmark & \cellcolor[HTML]{EAEAEA} & \cellcolor[HTML]{EAEAEA} & \cellcolor[HTML]{EAEAEA} & \cellcolor[HTML]{EAEAEA}\checkmark \\
 & NeRF & ChatSim~\cite{wei2024editable} & \checkmark & \checkmark & Input & BBox & \checkmark & \checkmark &  & \checkmark &  &  &  &  &  &  & 25.82 &  &  &  &  & \checkmark \\ \cmidrule(l){2-23} 
 &  & \cellcolor[HTML]{EAEAEA}HUGSIM~\cite{zhou2024hugsim} & \cellcolor[HTML]{EAEAEA}\checkmark & \cellcolor[HTML]{EAEAEA}\checkmark & \cellcolor[HTML]{EAEAEA}- & \cellcolor[HTML]{EAEAEA}BBox & \cellcolor[HTML]{EAEAEA}\checkmark & \cellcolor[HTML]{EAEAEA}\checkmark & \cellcolor[HTML]{EAEAEA} & \cellcolor[HTML]{EAEAEA}\checkmark & \cellcolor[HTML]{EAEAEA} & \cellcolor[HTML]{EAEAEA}\checkmark & \cellcolor[HTML]{EAEAEA}O & \cellcolor[HTML]{EAEAEA}RTX3090 & \cellcolor[HTML]{EAEAEA}89 & \cellcolor[HTML]{EAEAEA}27.40 & \cellcolor[HTML]{EAEAEA}28.79 & \cellcolor[HTML]{EAEAEA}\checkmark & \cellcolor[HTML]{EAEAEA}\checkmark & \cellcolor[HTML]{EAEAEA}\checkmark & \cellcolor[HTML]{EAEAEA} & \cellcolor[HTML]{EAEAEA} \\
\multirow{-7}{*}{Sim.} & G.S. & DrivingGaussian++~\cite{xiong2025drivinggaussian++} & \checkmark & \checkmark & Input & BBox & \checkmark & \checkmark &  & \checkmark &  &  &  & RTX8000 &  &  &  & \checkmark &  & \checkmark &  & $\star$ \\ \bottomrule 
\end{tabular}%
}\vspace{0.3em}\\
\justifying\hfill\\
{\scriptsize {\bf Task}: Recon.---Comprehensive Scene \textbf{Recon}struction; Gen.---\textbf{Gen}eration; Sim.---\textbf{Sim}ulator.}\\
{\scriptsize {\bf Repre.} Representation: PCL---\textbf{P}oint \textbf{Cl}oud.}\hfill\\
{\scriptsize {\bf Input-Point Cloud}: SV.---\textbf{S}uper\textbf{v}ision; AUG.---Optional \textbf{Aug}mentation.}\\
{\scriptsize {\bf Decomp.} Decomposition: BBox---3D \textbf{B}ounding \textbf{Box}; Self---\textbf{Self}-supervised; B/S---3D \textbf{B}ounding Boxes or \textbf{S}elf-supervised.}\hfill\\
{\scriptsize {\bf Rendering-Flow}: S---3D \textbf{S}cene Flow; O---2D \textbf{O}ptical Flow; V---\textbf{V}elocity Map.}\hfill\\
{\scriptsize {\bf Dataset}:\checkmark---Evaluated on; Num.---PSNR of reconstruction in dB; *---PSNR of Novel View Synthesis.}\hfill\\
{\scriptsize {\bf Open Source}:\checkmark---Released to the Public; $\star$---Recognized Method; S---Officially Announced Coming \textbf{S}oon.}\hfill\\
\vspace{-2.0em}
\end{table*}

\subsubsection{Non-rigid Agents Reconstruction}\label{sec:nonrigid}

Pedestrians and cyclists present significant reconstruction challenges due to their complex articulated movements and continuous shape variations during motion. Linear Blend Skinning (LBS) serves as a popular technique for 3D human body reconstruction, which first reconstructs the body in canonical space and parameterizes body deformations as skeletal pose variations, enabling reconstruction under arbitrary poses. Early LBS-based approaches~\cite{yangrecovering, Yanghumans3, Xiephysical} integrate pose estimation with spatial representations such as occupancy grids or meshes, enabling dynamic reconstruction of human geometry and motion from video sequences. However, these methods fail to recover visual details due to the inherent complexity of modeling both deformation and photometric properties simultaneously. Recent advances in human body reconstruction~\cite{li2022tava, guo2023vid2avatar, jiang2023instantavatar, qian20243dgsavatar, li2024animatable, lee2025geoavatar, IDOL} have introduced novel spatial representations into the LBS framework, achieving dual improvements in both photorealism and geometric fidelity. Notably, 3DGS-based methods~\cite{qian20243dgsavatar, lee2025geoavatar, li2024animatable, IDOL} further enable real-time rendering capabilities within this enhanced framework. Although existing methods demonstrate impressive fidelity on close-range, high-quality human digitization datasets, extending them to in-the-wild autonomous driving scenarios remains two critical issues which necessitate further research:
\begin{itemize}
    \item \textbf{Degradation at Long Range}. Distant pedestrians appear with low resolution, which significantly compromises the accuracy of both human pose estimation and relative camera pose estimation, leading to unstable reconstruction and loss of geometric detail.
    \item \textbf{Complex Occlusions}. In crowded traffic scenes, pedestrians are frequently occluded by other traffic participants or static infrastructure. Reconstructing complete geometries from such partial observations remains a significant hurdle for current algorithms.
\end{itemize} 

\section{Dynamic Driving Scene Reconstruction}\label{sec:scene_recon}

Beyond individual element reconstruction, dynamic driving scene reconstruction faces two unique fundamental challenges caused by ego and agent motion. The following analysis examines existing methods across these two challenges, highlighting how different approaches balance reconstruction quality, computational requirements, and task-specific applicability. Table~\ref{tab:comprehensive} exhibits detailed information and performance of dynamic scene reconstruction methods.

\begin{itemize}
    \item \textbf{Geometry Reconstruction}. The continuous movement of both the ego vehicle and surrounding agents leads to minimal overlap among observations, which makes traditional geometric reconstruction methods based on epipolar geometry ineffective. Section~\ref{sec:geometry} reviews relevant studies on reconstructing geometric details under different input modalities in dynamic driving scene.
    \item \textbf{Spatial-temporal Modeling}. Dynamic driving scenes involve intricate spatial-temporal relationships that go far beyond static 3D reconstruction. Section~\ref{sec:spatialtemporal} summarizes three representative strategies for modeling the spatial-temporal dependencies to achieve coherent and temporally consistent reconstruction.
\end{itemize}

\subsection{Geometry Reconstruction}\label{sec:geometry}

Dynamic driving scenarios captured from the ego-view perspective exhibit minimal cross-view overlap, posing increased difficulty to geometry reconstruction. Based on the different dependencies on input data modalities, methods can be categorized into LiDAR-centric methods that leverage point cloud inputs for geometric recovery explicitly and vision-centric methods that estimate geometry solely from images.

\subsubsection{LiDAR-centric}\label{sec:lidarcentric} RGB images provide rich visual details but lack geometric information, particularly in dynamic driving scenes with limited multi-view observations. LiDAR point clouds, on the other hand, offer precise geometric information but often exhibit sparse characteristics. LiDAR-centric methods utilize both to achieve complementary advantages.

Point clouds in LiDAR-centric frameworks serve two distinct technical functions: as supervisory signals to guide the optimization process or as initialization priors within direct inputs. As supervisory signals, point clouds generate ground-truth depth maps through imaging-plane projection, which supervise rendered depth maps to optimize either sampled points~\cite{tonderski2024neurad, turki2023suds, deng2023prosgnerf, yang2023emernerf, xie2023s, chen2025s, guo2023streetsurf, zhao2024drivedreamer4d}. Alternatively, other approaches directly leverage the inherent geometric information within point clouds to achieve near-optimal initialization, thereby avoiding sub-optimal solutions and accelerating convergence. Point cloud-based~\cite{ost2022neurallightfield, lu2023urban, zhu2024rpbg } and part of 3DGS-based methods require high-precision LiDAR point clouds~\cite{huang2024s3, chen2023periodic, chen2024omnire, peng2024desire, hess2024splatad, yan2024streetgaussian, hwang2024vegs, khan2024autosplat, wei2024emd } while others~\cite{zhou2024hugs, han2024ggs, mao2024dreamdrive} utilize SfM or COLMAP point clouds as an approximate substitute. \cite{yu2024sgd} further utilizes images and LiDAR point clouds as conditions for a video diffusion model to synthesize 3D Gaussians directly.

\subsubsection{Vision-centric}\label{sec:visioncentric}
Early vision-centric approaches~\cite{ost2021nsg, choi2024dico} primarily assessed rendering fidelity while neglecting geometric accuracy, resulting in incomplete 3D scene reconstruction. PNF~\cite{kundu2022panoptic} undergoes pre-training on ShapeNet~\cite{shapenet2015} to acquire vehicle shape priors, which facilitate geometric structure initialization in the model. \cite{li2024ho, li2024ggrt} enhance geometric reconstruction by strengthening epipolar correspondence across consecutive frames. Specifically, HO-Gaussian~\cite{li2024ho} generates virtual viewpoints with minimal displacements through linear transformations, while GGRt~\cite{li2024ggrt} employs cross-attention mechanisms to establish correlations among sampled points along epipolar rays. DrivingForward~\cite{tian2024drivingforward} constrains depth estimation accuracy by enforcing consistency of rendered depth map across multi-perspective surround-view images, thereby enhancing geometric reconstruction quality. VDG~\cite{li2024vdg} employs off-the-shelf visual odometry for monocular depth estimation, subsequently achieving geometric reconstruction through inverse projection.

Accurate 3D geometry recovery from 2D images necessitates the construction of precise camera models, where the correctness of calibration parameters plays a critical role. However, calibrations obtained during high-speed motion, particularly for extrinsic parameters, are typically noisy and rely on costly manual corrections. This has driven some methods to seek independence from calibration inputs. VDG~\cite{li2024vdg} estimates and refines extrinsics through a pre-trained visual odometry, while DrivingForward\cite{tian2024drivingforward} uses fixed vehicle-to-camera and inter-frame camera motion to eliminate dependency on camera extrinsics.

\subsection{Spatial-temporal Modeling}\label{sec:spatialtemporal}
The dynamic behavior of agents introduces complex spatial-temporal relationships that transform driving scenarios from static 3D reconstruction into dynamic 4D modeling challenges. Existing approaches for capturing these spatial-temporal dynamics can be categorized into three primary strategies: per-frame reconstruction methods that independently process temporal snapshots, scene graph-based approaches that simplify dynamics as object bounding box movements, and native 4D representation methods that uniformly encode time-varying scene properties across the entire spatial-temporal domain.

\subsubsection{Per-frame Reconstruction}\label{sec:perframe}
Per-frame reconstruction decouples temporal frames and computes an independent 3D representation for each frame, which shrinks the temporal scale to a single or a few frames. This pipeline significantly simplifies the complexity of spatial-temporal relationships, facilitating feedforward methods~\cite{omniscene, tian2024drivingforward} in straightforward design of models and losses. \cite{omniscene, tian2024drivingforward} perform per-frame per-pixel Gaussian predictions while \cite{tian2024drivingforward} enhances spatial-temporal consistency by leveraging multi-view and multi-frame inputs as prediction context. Nonetheless, this pipeline overlooks cross-frame consistency, particularly within static elements, introducing significant redundancy in 3D representations and resulting in unnecessary computational and storage overhead.

\subsubsection{Scene Graph}\label{sec:scenegraph}

\begin{figure}[t]
    \centering
    \subfloat[3D Visualization]{\includegraphics[width=0.45\columnwidth]{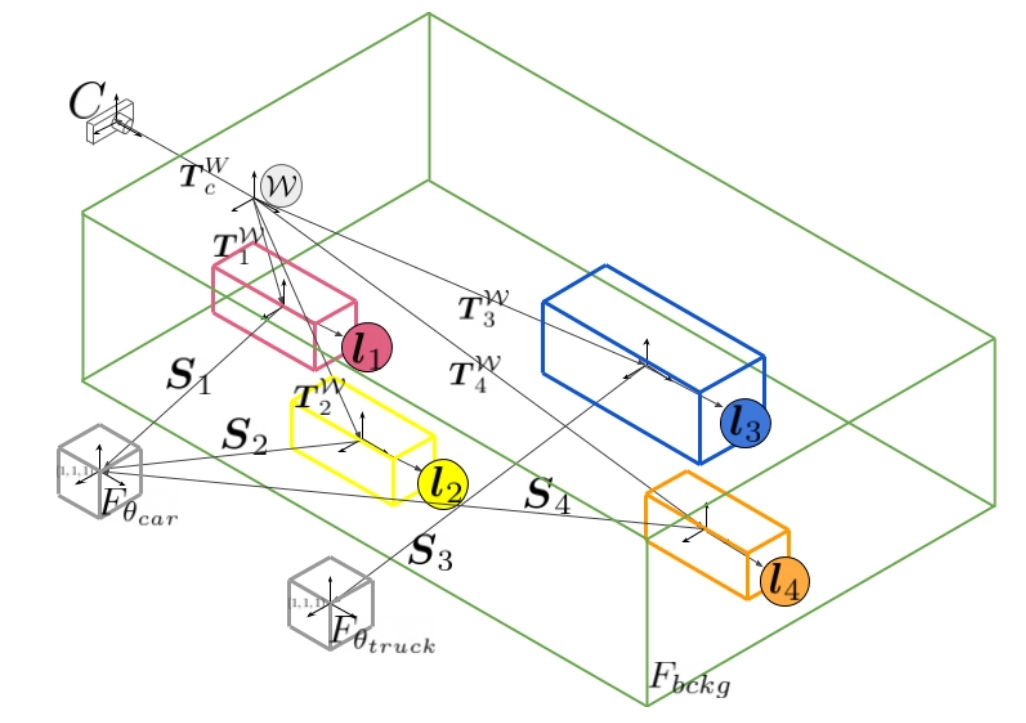}}
    \subfloat[Ego View]{\includegraphics[width=0.55\columnwidth]{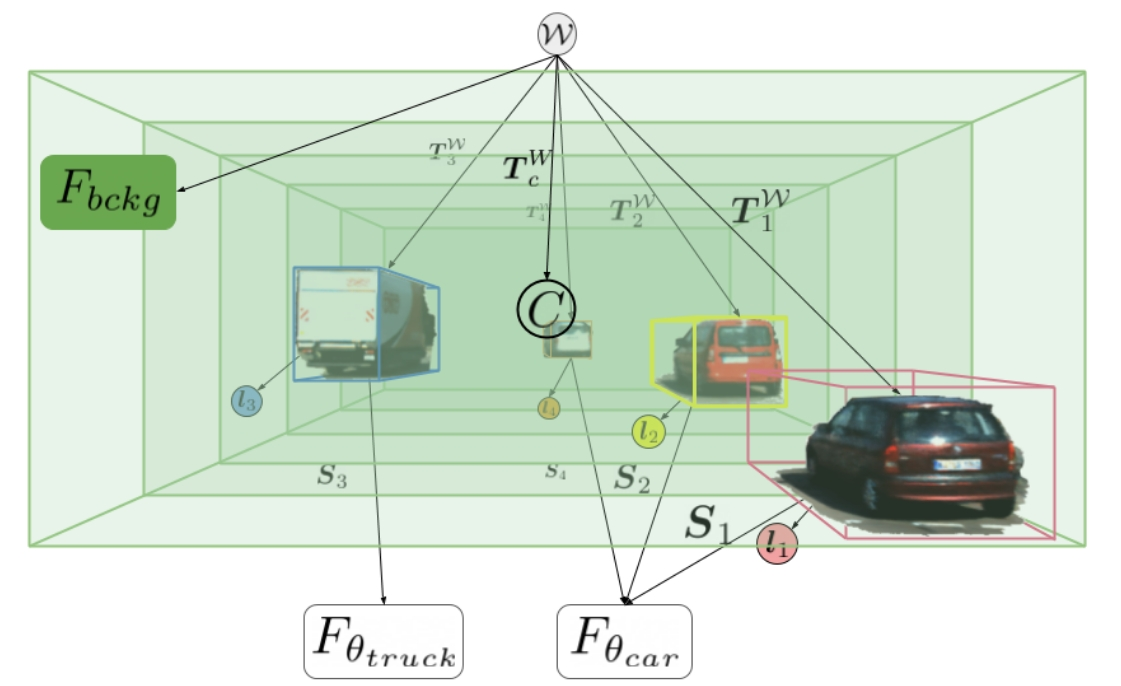}}
    \caption{Decoupling Paradigm of Neural Scene Graph~\cite{ost2021nsg}. Agents are modeled in corresponding local coordinates and resembled into complete scene in global coordinates with transformation and scaling (visualized in arrow).}
    \label{fig:nsg}\vspace{-1.0em}
\end{figure}

As a pioneering work in dynamic driving scene reconstruction, NSG~\cite{ost2021nsg} proposes \emph{Scene Graph} to model spatial-temporal relationships by decomposing dynamic driving scenes into individual agents with their associated bounding boxes and radiance properties (Fig~\ref{fig:nsg}). This approach simplifies temporal dynamics to object-level transformations of bounding boxes over time, enabling the reconstruction of agents in their own canonical frame, and has been adopted in subsequent methods~\cite{tonderski2024neurad, deng2023prosgnerf, li2023read, zhou2024hugs, xie2023s, chen2025s, chen2024omnire, zhou2024drivinggaussian, hwang2024vegs, khan2024autosplat, hess2024splatad, yuan2025uni}. Beyond reconstruction, Scene Graph enables users to readily achieve parameterized control of agents within the scene with different object arrangements and the creation of novel scene compositions by integrating agent pose control modules, giving rise to a series of open-loop and closed-loop simulators~\cite{yang2023unisim, yan2024oasim, ljungbergh2024neuroncap, wu2023mars, zhou2024hugsim}.

However, scene graph modeling faces two significant limitations in driving scenarios. First, scene graph methods critically depend on accurate 3D bounding boxes, which impose substantial computational costs whether obtained through manual annotation or automated detection and tracking systems. The inherent noise and inaccuracies in 3D bounding box estimation further degrade scene graph fidelity. To address these challenges, HUGS~\cite{zhou2024hugs} introduces learnable kinematic parameters combined with unicycle model-based regularization to improve 3D bounding box estimation while reducing reliance on external detection systems. Second, the rigid body assumption inherently fails to capture non-rigid deformations in dynamic agents such as pedestrians and cyclists. While some approaches~\cite{chen2024omnire, yuan2025uni} attempt to address this by integrating parametric models like SMPL for human representation, the resulting appearance artifacts limit practical deployment. Despite these advances, research in scene graph-based dynamic reconstruction remains active, with ongoing efforts to further address the limitations mentioned above.

\subsubsection{4D Representations}\label{sec:4drepre}
\begin{figure}[t]
    \centering
    \subfloat[PVG]{\includegraphics[height=0.78in]{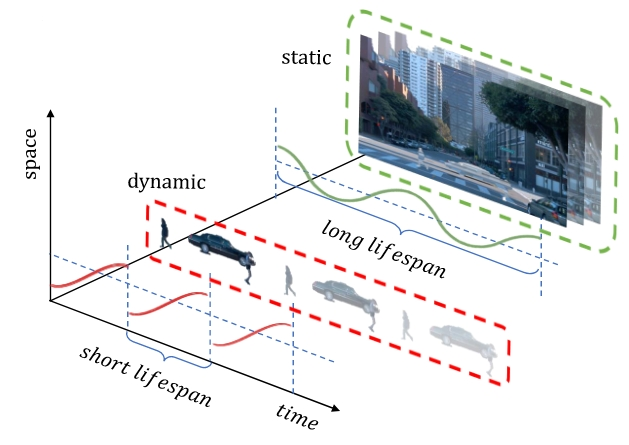}}
    \subfloat[Training Pipeline]{\includegraphics[height=0.78in]{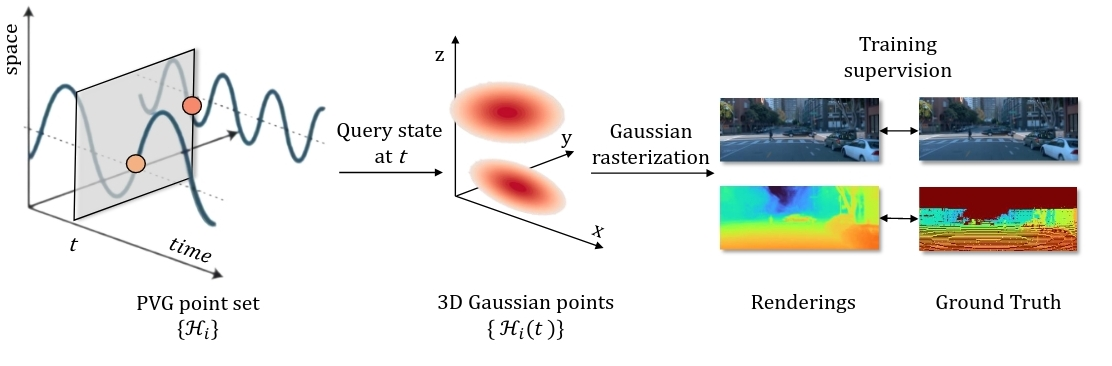}}
    \caption{Illustration of Periodic Vibration Gaussian (PVG)~\cite{chen2023periodic}. (a): Static background with long lifespan and dynamic agents with short ones; (b): Pipeline of PVG. Zoom-in for details.}\vspace{-1.0em}
    \label{fig:pvg}
\end{figure}

By elevating static 3D representation into dynamic 4D space via introducing an additional temporal dimension, 4D representations enable native modeling of spatial-temporal relationships for both rigid and non-rigid agents without compromises~\cite{turki2023suds, yang2023emernerf, choi2024dico, huang2024s3, chen2023periodic, peng2024desire, li2024vdg, mao2024dreamdrive, yang2024storm, zhao2024drivedreamer4d}. NeRF-based methods~\cite{turki2023suds, yang2023emernerf} utilize a separate dynamic component to represent dynamic agents within the scene, which accepts time as an additional input to the radiance field, reformulating the color and density as time-varying attributes. Some Gaussian-based works~\cite{mao2024dreamdrive, zhao2024drivedreamer4d} model 4D Gaussians as a combination of a 3D Gaussian representing the initial state and a time-varying offset of each attribute, forming a basic form of 4D Gaussian. Periodic Vibration Gaussian (PVG)~\cite{chen2023periodic} proposes a novel spatial-temporally unified representation based on 3DGS (Fig.~\ref{fig:pvg}) for modeling time-varying dynamic elements and is adopted by \cite{peng2024desire, li2024vdg}. PVG reformulates the mean and opacity of traditional 3D Gaussian into continuous time-dependent functions with learnable lifespan, extending 3D Gaussian from reconstruction to an analytical primitive.

Beyond natively representing scene dynamics, 4D representation-based methods also eliminate reliance on external bounding box annotations by achieving self-supervised dynamic-static decomposition during reconstruction. These approaches employ various strategies for automatic scene segmentation. DiCoNeRF~\cite{choi2024dico} identifies dynamic objects by computing cosine similarity between ground truth and rendered features, while $S^3$Gaussian~\cite{huang2024s3} utilizes spatial-temporal Hexplane features to separate static and dynamic elements, with dynamic agents naturally captured on temporally varying planes. \cite{mao2024dreamdrive} employs a two-stage approach, first constructing a time-invariant static model with 3D Gaussians, then identifying dynamic agents as regions exhibiting large rendering errors. PVG-based methods offer additional flexibility through learnable attributes as PVG~\cite{chen2023periodic} dynamically allocates learnable lifespans and velocities to scene elements during training, enabling subsequent works~\cite{peng2024desire, li2024vdg} to derive motion masks from velocity attributes for static-dynamic decoupling explicitly.

4D representation demonstrates notable advantages in processing complex driving scenarios due to its capacity to simultaneously handle both rigid and non-rigid agents without relying on any a priori assumptions or compromises. Furthermore, the self-supervised dynamic-static decomposition based on 4D representation eliminates the dependency on costly 3D bounding box annotations, thereby establishing it as the most promising spatiotemporal paradigm for future applications.

\section{Applications}\label{sec:application}
\subsection{Data Collection \& Augmentation}\label{sec::data_collection}
\begin{figure}[t]
    \centering
    \subfloat[Multi-modal rendering]{\includegraphics[width=0.45\linewidth]{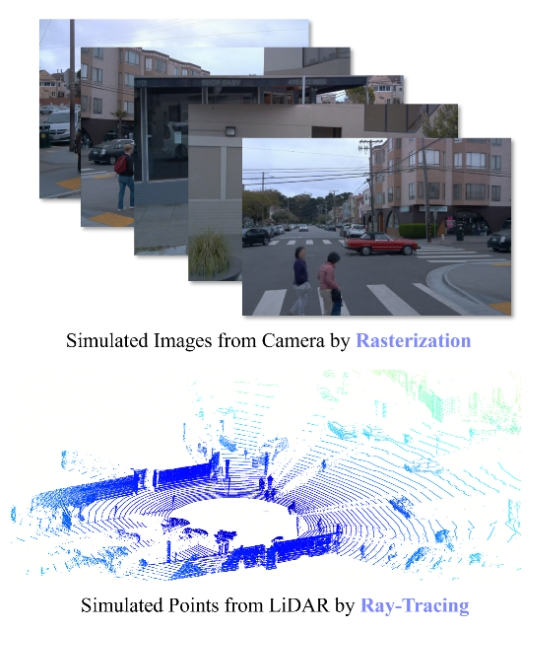}}
    \subfloat[Semantic Annotations]{\includegraphics[width=0.5\linewidth]{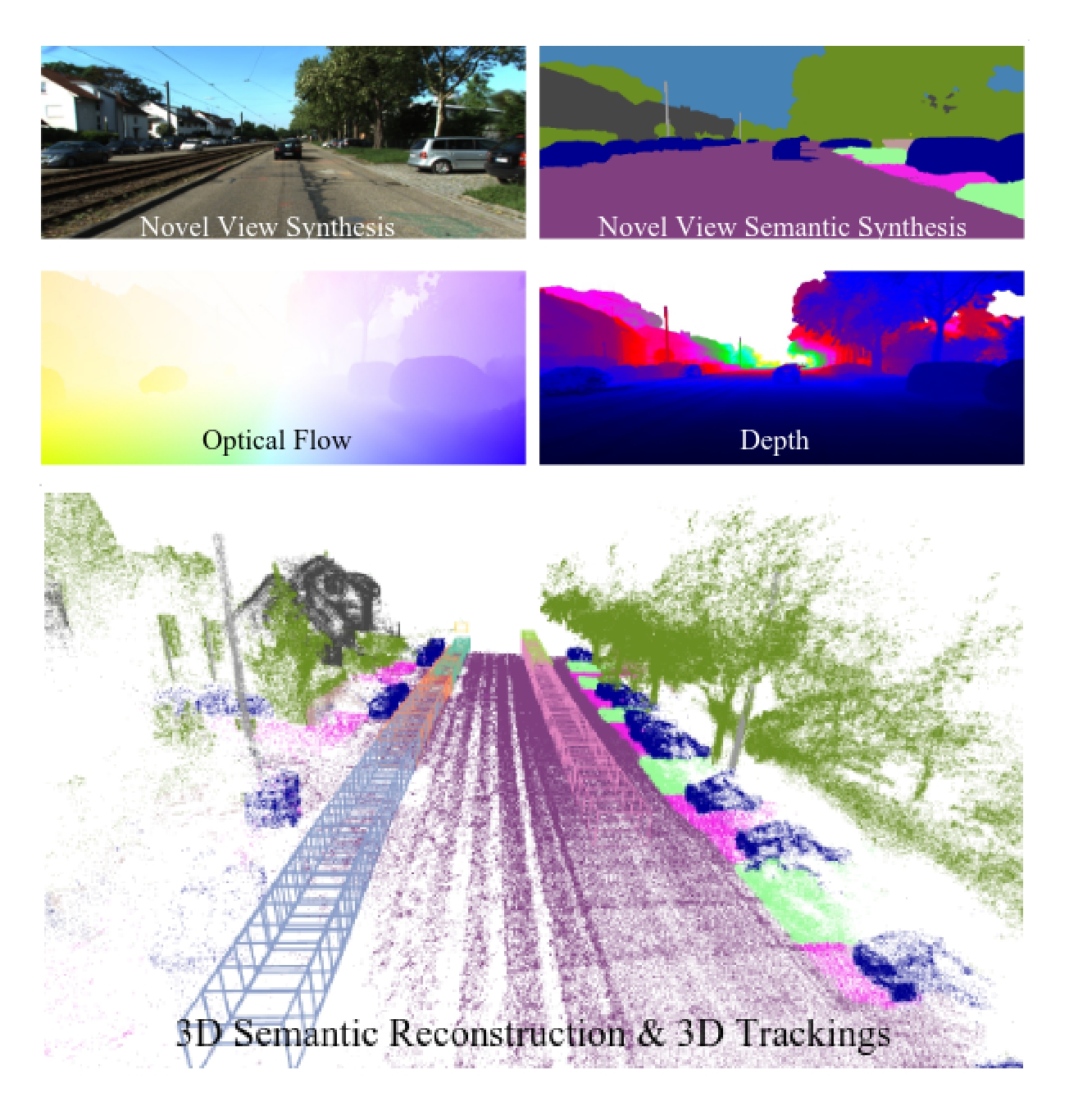}}
    \caption{Visualization of Rendering Modalities. (a): Synchronized image and point cloud rendering from Uni-Gaussians~\cite{yuan2025uni}; (b): Comprehensive annotations from HUGS~\cite{zhou2024hugs}, including semantic segmentation for both image and point cloud, optical flow and depth image.}\vspace{-1.0em}
    \label{fig:modalvis}
\end{figure}

Reliable perception and decision-making in complex traffic environments rely heavily on large-scale, high-quality multi-modal datasets. However, acquiring such data in the real world suffers from high cost and safety risks. 3D reconstruction offers an efficient alternative by generating rich, physically consistent data from limited inputs. For example, reconstruction can be achieved using only 2D images to generate accurate depth images~\cite{li2024ggrt,tian2024drivingforward,li2024vdg}, or simulating LiDAR point clouds~\cite{tonderski2024neurad, yang2023emernerf, hess2024splatad, yuan2025uni}. This substantially reduces the reliance on expensive sensor suites and enables scalable data collection under configurable settings.
Furthermore, 3D reconstruction frameworks integrate multi-modal rendering pipelines to synthesize realistic and fine-grained annotations, including semantic segmentation maps~\cite{zhou2024hugs, huang2024s3}, agent pose~\cite{kundu2022panoptic, yan2024streetgaussian}, and optical or scene flows~\cite{zhou2024hugs, yang2024storm, wei2024emd}, effectively bridging the gap between simulation and real-world data. Figure~\ref{fig:modalvis} illustrates representative examples of these rendering modalities and annotations within autonomous driving contexts.


\textbf{Image} constitutes an indispensable component across all methods as the most fundamental modality in 3D reconstruction. Notably, DiCo-NeRF~\cite{choi2024dico} stands out as one of the few works implementing fisheye-view rendering, providing critical support for panoramic perception systems and solutions.

\textbf{Depth Image} is adopted by most methodologies~\cite{kundu2022panoptic, tonderski2024neurad, turki2023suds, yang2023emernerf, deng2023prosgnerf, xie2023s, chen2025s, li2024ggrt, zhao2024tclc, yu2024sgd, huang2024s3, chen2023periodic, yan2024streetgaussian, chen2024omnire, peng2024desire, li2024vdg, zhou2024hugs, hwang2024vegs, han2024ggs, zhao2024drivedreamer4d, tian2024drivingforward, hess2024splatad, yang2024storm, omniscene } as 2D geometric proxies for spatial information, where 3D scene geometry is projected onto the 2D imaging plane via perspective transformation. Although depth images retain partial spatial cues (e.g., relative distances), they inherently incur lossy spatial encoding due to dimension reduction, discarding critical 3D structural details (e.g., surface normals, multi-view consistency) and introducing ambiguity in occluded regions.

\textbf{Point Cloud} has gradually garnered significant attention with the increasing adoption of LiDAR in autonomous driving systems. It can be synthesized with LiDAR simulation in scene reconstruction. Due to the uniqueness of the imaging method, LiDAR point clouds exhibit distinct characteristics compared with 2D images. LiDAR ray drop probability and rolling shutter effects from sequential scanning during rapid vehicle movement are explicitly modeled by \cite{tonderski2024neurad, hess2024splatad}, enabling more realistic multimodal simulation. EmerNeRF~\cite{yang2023emernerf} models asynchronous LiDAR sampling via depth and line-of-sight constrained volumetric rendering. SplatAD~\cite{hess2024splatad} develops a projection algorithm in spherical coordinates and customizes a tiling, sorting, and rasterization pipeline tailored for imitating the imaging process of LiDAR. Uni-Gaussians~\cite{yuan2025uni} combines ray-tracing and rasterization into a high-fidelity, efficient framework for joint camera-LiDAR simulation, using bounding volume hierarchy and custom CUDA kernels to accelerate LiDAR ray-tracing.

\textbf{Segmentation} plays a vital role in perception modules. PNF~\cite{kundu2022panoptic} represents a pioneering extension of NeRF to semantic radiance fields, which augments rendered outputs with instance segmentation, thereby supporting diverse downstream perception tasks. \cite{turki2023suds, zhou2024hugs} follows PNF to additionally incorporate instance segmentation annotations, and \cite{turki2023suds, zhou2024hugs, chen2025s, huang2024s3, chen2023periodic, yan2024streetgaussian } provide semantic segmentation results. 

\textbf{Vehicle pose}, which provides positional information of surrounding vehicles, constitutes a critical component in scene understanding, with standard formats including 3D bounding boxes and waypoints. While conventional approaches require pose as explicit input, emerging methods not only eliminate this dependency but also pose during 3D reconstruction. SUDS~\cite{turki2023suds} derives bounding boxes via PCA from instance segmentation results of 3D point clouds, while~\cite{kundu2022panoptic, yan2024streetgaussian} optimizes parameterized poses within their training frameworks.

\textbf{Motion fields}, including 2D optical flow, 3D scene flow and velocity field, reflect the motion patterns of the scene.

\emph{Optical flow} refers to the 2D motion field of pixels between consecutive frames in an image sequence, enhancing spatial-temporal coherence in rendering~\cite{xie2023s, chen2025s, zhou2024hugs, zhou2024hugsim}. \cite{xie2023s, chen2025s} introduces optical flow as a cross-view geometric consistency regularization to mitigate aliasing artifacts. HUGS and HUGSIM~\cite{zhou2024hugs, zhou2024hugsim} leverage supervision over optical flow to improve the rendered depth image.

\emph{Scene flow} extends this concept to 3D, describing the 3D motion of objects in the scene. Several approaches~\cite{turki2023suds, yang2023emernerf, yang2024storm} leverage scene flow to aggregate temporal information and enforce spatial correspondence in adjacent frames, effectively enhancing the temporal coherence of dynamic agents.

\emph{Velocity field} normalizes the scene flow along the temporal dimension, thereby providing a quantitative ground truth for downstream tasks such as object tracking. The reformulation of 3D Gaussian in PVG~\cite{chen2023periodic, peng2024desire, li2024vdg} inherently incorporates velocity attributes, while EMD~\cite{wei2024emd} learns a deformation field of the scene at each time step, both of them enable direct calculation of velocity fields.

\subsection{Localization \& Mapping}\label{sec:slam}
Precise localization and high-fidelity mapping form the foundational spatial intelligence for autonomous vehicles, enabling safe navigation and context-aware decision-making. 3D reconstruction, particularly through 3DGS, delivers a groundbreaking spatial representation for SLAM systems via real-time photorealistic rendering capabilities, significantly enhancing robustness in complex outdoor environments. \cite{xiao2024liv, zhou2025gsgvins, xie2024gs} jointly advance online dense mapping through continuous scene representation optimization and achieve centimeter-level localization accuracy via iterative pose refinement. BEV-GS~\cite{wu2025bevgs} recovers fine-grained road surface details such as road surface materials, markings, and damages, enhancing the smoothness and comfort of motion planning.

\begin{figure}[t]
    \centering
    \includegraphics[width=\linewidth]{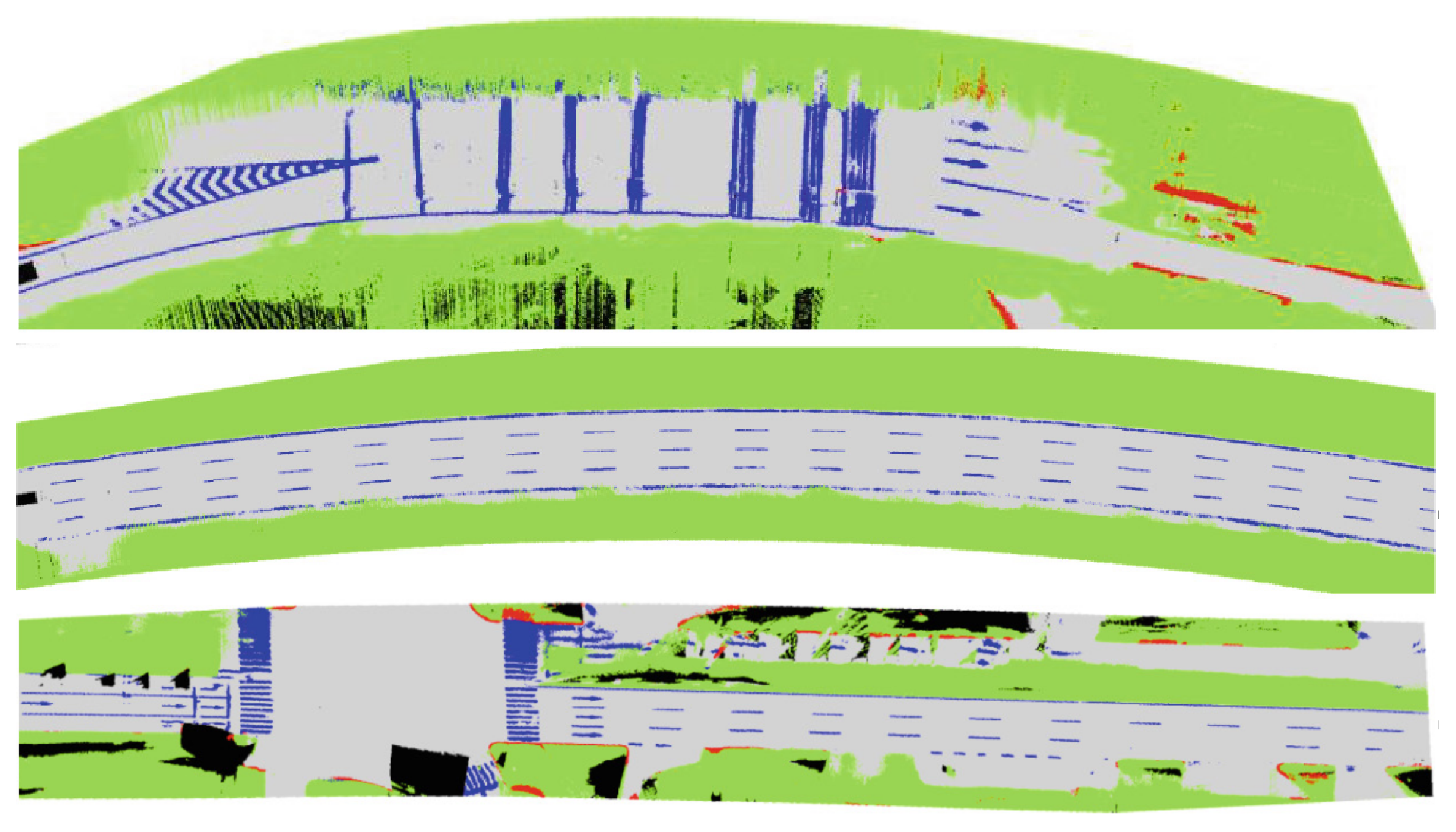}
    \caption{Semantic map from EMIE-MAP~\cite{wu2024emie}.}
    \label{fig:mesh_recon}\vspace{-1.0em}
\end{figure}

Semantic map (Fig.~\ref{fig:mesh_recon}) provides a structured representation of the environment, encoding semantic information like lane markings, traffic signs, and road types. Previous semantic map extraction relied on manual annotation, incurring significant labor and financial costs while struggling to maintain map freshness. Road surface reconstruction methods~\cite{mei2024rome, wu2024emie, zhang2024vision, wang2024nero, chen2024camav2} incorporate semantic information modeling during reconstruction, enabling online extraction of semantic maps. These maps support tasks such as navigation and path planning, while also providing pseudo-ground truth for scene understanding tasks like lane line detection.

However, all these SLAM and semantic map extraction methods have only been validated on high-end GPUs, with \cite{xiao2024liv} achieving 8 FPS on RTX 4090 and \cite{wu2025bevgs} reaching 26 FPS on A100. A substantial disparity in computational capability exists between mobile platforms and these high-end GPUs. Consequently, performance metrics for 3D reconstruction-based SLAM methods, specifically computational resource consumption and latency, warrant further testing and validation.\vspace{-1.0em}

\subsection{Scene Understanding}\label{sec:understand}

Beyond its role in geometric reconstruction, recent studies have explored using 3D representations as a unified latent feature for understanding and interpreting complex traffic scenarios. By augmenting with a compact neural representation, primitives can encapsulate the essential multimodal information required by autonomous driving systems into a single latent feature that integrates visual, geometric, motion, and semantic cues. Such features provide autonomous driving systems with richer environmental context, thereby enhancing perception accuracy as well as spatial understanding and reasoning capability.

StreetUnveiler~\cite{xu2024unveiler} extends static background reconstruction by identifying and filtering stationary traffic participants (e.g., parked vehicles) to isolate unobstructed infrastructure. This facilitates a more stable environmental modeling and supports downstream tasks such as high-definition map generation and long-term scene understanding.

3D reconstruction enhances perception modules of autonomous driving systems by encoding persistent scene knowledge and completing missing geometry~\cite{yuan2024presight, yang2024unipad}. PreSight~\cite{yuan2024presight} leverages NeRF to construct static prior memories based on historical observations, which encodes city-scale NeRF reconstructions in memory, enabling efficient environmental context augmentation for online perception systems. UniPAD~\cite{yang2024unipad} innovatively integrates 3D reconstruction to address partial observation in driving scenarios, enhancing scene perception and understanding. It integrates masked autoencoders to complete geometry with partially observed images and point clouds, improving performance on tasks like object detection and semantic segmentation.

GaussianAD~\cite{zheng2024gaussianad} presents an end-to-end framework that employs 3D Gaussians as unified scene features, thereby bridging dense tasks (3D semantic occupancy) and sparse tasks (detection/motion prediction). It introduces Gaussian Flow, a 3D Gaussian motion field jointly predicted via neural scene features and ego-motion trajectory estimation, to enable efficient future scene computation through affine transforms for robust spatial-temporal forecasting.

RAD~\cite{gao2025rad} introduces a simulator-in-loop reinforcement learning framework integrating 3D reconstruction-based simulation for iterative policy optimization. A 3DGS simulator dynamically reconstructs high-risk scenarios through log-replay of traffic participants, enabling adversarial training via edge case generation. The planner is optimized via simulator-derived rewards, achieving progressive performance gains through interaction with a cyclical environment.\vspace{-0.5em}

\subsection{Simulation}\label{sec:simulation}

Simulators employ modular architectures and user-friendly interfaces to encapsulate driving scene reconstruction methods, delivering both high-fidelity multimodal sensor simulation and parameterized traffic agent control via scene graphs for open-loop~\cite{wu2023mars} or closed-loop~\cite{yang2023unisim, yan2024oasim, ljungbergh2024neuroncap, zhou2024hugsim} verification.

Early game-engine-based simulators~\cite{dosovitskiy2017carla, wang2022learning, viworldsim} failed to simulate realistic driving scenarios, with rendered images exhibiting distinctly artificial features. This incurred a significant distribution discrepancy between the simulated and real-world driving scenarios, which in turn led to a evident performance degradation for autonomous driving systems due to the sim-to-real gap~\cite{elmquist2024methodology, pahk2023effects, pasios2025carla2real}. Consequently, many simulators~\cite{wu2023mars, yang2023unisim, yan2024oasim, ljungbergh2024neuroncap, zhou2024hugsim} have focused on improving their rendering fidelity. MARS~\cite{wu2023mars} provides a highly configurable open-loop simulator platform that supports switching between diverse NeRF variants for the backend renderer, multiple ray sampling strategies, and multimodal inputs, yet it only supports simulating scene variations through recorded playback. The open-loop simulation mode fails to provide counterfactual or long-term simulation, significantly limiting its effectiveness in training autonomous driving systems.

Closed-loop simulators~\cite{yang2023unisim, yan2024oasim, ljungbergh2024neuroncap, zhou2024hugsim, wei2024editable} enable traffic participants to react to environmental changes, thereby better approximating real-world traffic dynamics and substantially expanding their application scope. To model diverse traffic participants in driving scenarios, UniSim~\cite{yang2023unisim} employs a hypernetwork to dynamically generate their features, whereas OASim~\cite{yan2024oasim} predefines an asset library of agents. However, these simulators are naturally constrained by the complexity of NeRF, which limits their ability to support real-time rendering.

HUGSIM~\cite{zhou2024hugsim} extends HUGS as a closed-loop simulator, establishing the first 3DGS-based simulator. It outperforms NeRF-based methods in terms of real-time rendering while supporting a diverse range of rendering modalities. HUGSIM integrates a trajectory generator to eliminate manual vehicle trajectory design and incorporates a vehicle asset library reconstructed from 3DRealCar~\cite{du20243drealcar} to enhance rendering fidelity.

Beyond simulators primarily focus on enhancing perceptual reconstruction quality, research on scene editing has begun to emerge. ChatSim~\cite{wei2024editable} integrates a Large Language Model (LLM) to adjust scenes based on text prompts, enabling functionalities such as adding or removing agents, modifying agent behaviors, and separating foreground and background, providing rich simulation capabilities through a user-friendly method. DrivingGaussian++~\cite{xiong2025drivinggaussian++} extends DrivingGaussian~\cite{zhou2024drivinggaussian} from a reconstruction method to a simulation method, and enables text-based scene editing via a Large Language Model (LLM), including agent trajectories, weather conditions, and object textures.\vspace{-1.0em}

\subsection{Scene Generation}\label{sec:scenegen}

\begin{figure}[t]
 \subfloat[Night]{%
       \includegraphics[width=0.95\linewidth]{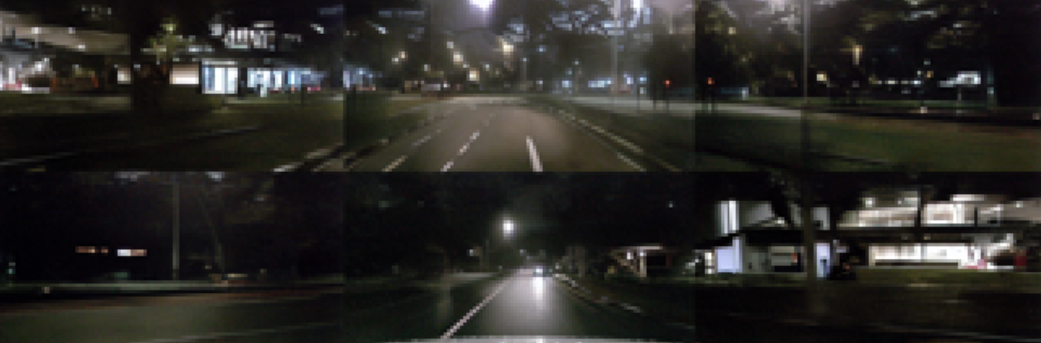}}
       \\
  \subfloat[Rainy]{%
        \includegraphics[width=0.95\linewidth]{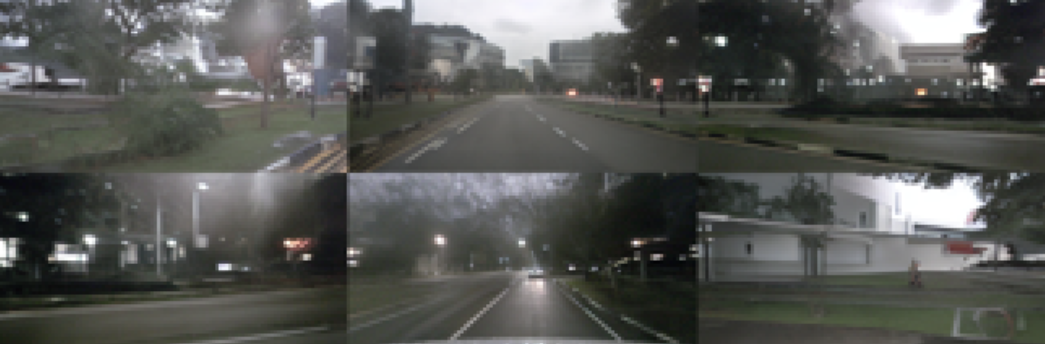}}
    \caption{Synthesized scene under different environmental conditions by UniScene~\cite{yuan2025uni}.}
    \label{fig:uniscene}\vspace{-2.0em}
\end{figure} 

Building upon advances in 3D reconstruction, scene generation techniques extend these capabilities from passive reconstruction to active synthesis. With photorealistic and geometry consistent 3D representation, they enable the creation of realistic and controllable virtual driving scenes that capture complex spatial structures and sensor characteristics under the guidance from high-level semantic conditions, such as BEV maps or text prompts. Such synthetic data, particularly from emergency scenarios and extreme weather conditions, is vital for improving the robustness of autonomous driving systems and mitigating long-tail challenges that are difficult to capture in real-world datasets. As an integral component of methodological paradigm evolution, generative approaches are comparatively analyzed in Table~\ref{tab:comprehensive}.

Existing generative approaches~\cite{gao2024magicdrive3d, li2024uniscene, yan2025drivingsphere} both condition on semantic inputs and leverage video generation models to synthesize perceptual details, albeit through distinct technical pathways, pre-generation and post-generation. MagicDrive3D~\cite{gao2024magicdrive3d} adopts a pre-generation paradigm that first synthesizes visual details in 2D space before 3D reconstruction. The framework initially employs a video diffusion model to generate surrounding-view videos compliant with BEV layouts from the ego's perspective, then elevates 2D video sequences to 3D point clouds through monocular depth estimation and inverse projection, ultimately fusing both modalities for 3D driving scene generation. \cite{yan2025drivingsphere,li2024uniscene} adopt post-generation paradigm that synthesizes semantic occupancy fields as guidance for visual details generation. UniScene~\cite{li2024uniscene} innovatively employs 3DGS as semantic carriers rather than visual information carriers, utilizing an occupancy diffusion model to produce semantic 3D Gaussians conforming to BEV inputs. These semantic 3D Gaussians are subsequently rendered into semantic maps, serving as guidance for video diffusion model to synthesize the final driving video.

While both paradigms achieve novel view synthesis from BEV layouts to arbitrary perspectives under various environmental conditions specified by text prompts (Fig.~\ref{fig:uniscene}), the pre-generation approach consolidates visual detail generation and viewpoint transformation within a single video diffusion model, exacerbating inherent inter-frame inconsistencies. In contrast, post-generation leverages explicit 3D representations to enable consistent multi-view rendering, significantly improving both temporal coherence and cross-view consistency.

\section{Retrospect \& Prospect}\label{sec:retro}

\subsection{Technical Evolutions} \label{sec:techevo}
Learning-based 3D reconstruction techniques for autonomous driving have undergone significant advancements across multiple dimensions, including input modality requirements, methodological paradigms and representation schemes.

\textbf{1) Representation Evolutions} \label{sec:repreevo}
As the cornerstone of 3D reconstruction, the evolution of representation has fundamentally driven the progress of the field.
The evolution of learning-based 3D representations began with NeRF~\cite{nerf}, a pioneer that offered a memory-efficient and high-fidelity pipeline but was hindered by slow volume rendering due to its implicit nature. This limitation led to a renewed focus on explicit representations, which leverage efficient rasterization. While point clouds offered geometric precision and meshes provided high-fidelity appearance, they suffered from limited image quality and intensive optimization, respectively. The breakthrough came with 3D Gaussian Splatting~\cite{3dgs}, which achieved an excellent balance of photorealism and high efficiency in both training and rendering, despite a slight lack of geometric accuracy. Subsequently, to address the unique limitations of autonomous driving, the field has progressed to the next stage. This includes hybrid representations that synergize the strengths of different representations, such as combining 3D Gaussians with geometric primitives~\cite{song2024gvkf, shi2024dhgs} to improve accuracy, and the development of novel 4D representations~\cite{turki2023suds, yang2023emernerf, chen2023periodic, mao2024dreamdrive, zhao2024drivedreamer4d}, which efficiently model dynamic scenes by incorporating time-varying parameters. The evolution of representations originated from three fundamental challenges in driving scene reconstruction, visual fidelity, geometric accuracy, and spatial-temporal relationship modeling. While significant progress has been made in rigid object reconstruction, non-rigid agents remain an underexplored frontier.

\textbf{2) Input Data Requirements}\label{sec:inputreq} Early learning-based 3D reconstruction in autonomous driving~\cite{ost2021nsg, choi2024dico} focused solely on photorealism while neglecting geometric accuracy. Subsequent multimodal approaches~\cite{tonderski2024neurad, turki2023suds, zhao2024tclc, chen2023periodic, xiao2024liv, xie2024gs} achieved high-quality geometric reconstruction by incorporating LiDAR point clouds, but incur significantly increased costs. The dependency of LiDAR substantially limited the practicality of 3D reconstruction, thus driving research toward relaxing input data requirements. Unimodal methods first eliminated dependence on LiDAR point clouds~\cite{li2024ggrt, li2024ho}, and some approaches~\cite{tian2024drivingforward, li2024vdg, zhou2025gsgvins} further enabled 3D reconstruction with pose-free visual inputs. Emerging techniques like VGGT~\cite{wang2025vggt} have exhibited potential for pose-free 3D reconstruction of driving scenes in a feedforward manner.

\textbf{3) Methodology Paradigm}\label{sec:methpara} The methodological paradigm has evolved from the earliest per-scene optimization~\cite{ost2021nsg, kundu2022panoptic, yang2023emernerf} to feedforward reconstruction~\cite{tian2024drivingforward, omniscene}, and further to generation~\cite{gao2024magicdrive3d, li2024uniscene}, significantly streamlining the data acquisition process. Per-scene optimization requires substantial computational resources to reconstruct a scene offline, which fails to meet the data volume demands of contemporary data-driven models and thus lacks practicality. In contrast, feedforward models establish a universal 2D-to-3D mapping to predict scene representations in real-time online, significantly enhancing the scalability of 3D reconstruction methods. Generative models further revolutionize this paradigm by enabling the direct generation of 3D driving scenes from abstract inputs (e.g., text prompts or BEV maps), completely bypassing traditional sensor data acquisition pipelines while maintaining reconstruction quality, propelling data scalability to the next level. This methodological paradigm evolution reflects the field's progression toward efficient, large-scale 3D content creation for autonomous systems.
\vspace{-1.0em}

\subsection{Challenges \& Future Directions}\label{sec:challenge}
While learning-based 3D reconstruction methods have shown superior capabilities in replicating real-world scenarios and potential in autonomous driving applications, several critical challenges still remain to be addressed in future research.

\textbf{1) Closed-loop Simulation}\label{sec:closesim} Current 3D reconstruction-based simulators have achieved significant advances in reconstruction fidelity. Although minor issues such as edge blurring and aliasing persist, they have substantially alleviated the sim-to-real gap associated with game-engine-based simulators. However, the agent behavior in closed-loop simulation is oversimplified~\cite{yan2024oasim, zhou2024hugsim} or reliant on human intervention~\cite{yang2023unisim, ljungbergh2024neuroncap, wei2024editable}. This defect will lead to the overfitting of autonomous driving systems to simple scenarios, compromising their ability to handle real-world complexities. Agent behavior simulation models~\cite{zhou2024behaviorgpt, wu2024smart} simulate realistic agent interactions based on the historical motion of all agents in the scene and predict future poses, offering authentic scenario evolution. Agent behavior simulation models can be integrated as a plug-and-play module, enabling seamless coupling with scene graph-based simulators. In addition, the development of world models~\cite{bruce2024genie, rtfm, zhou2025hermesunifiedselfdrivingworld} is advancing rapidly. By learning the principles of how the world evolves from vast amounts of data, world models can generate highly realistic and interactive scenes. This capability extends beyond just traffic participants to include all objects within the environment, such as traffic signs, infrastructures, and weather variations. Integrating world models with closed-loop simulation not only enhances the diversity of simulated scenarios but, more importantly, helps identify and improve the key aspects of 3D reconstruction that are critical to autonomous driving, thereby providing a robust training foundation for current and future autonomous driving systems.

\textbf{2) Weather/Lighting Editing}\label{sec:weather} Autonomous vehicles require robustness in all conditions, but existing datasets and simulators predominantly feature ideal weather and lighting, lacking reflection of adverse weather or low-light scenarios. Weather editing methodologies~\cite{sang2025weather, qian2025weatheredit} facilitate the simulation of diverse atmospheric conditions, including rain, snow, and fog, through injecting dynamic particles within scene representations, while driving scene generative models~\cite{gao2024magicdrive3d, li2024uniscene} employ video diffusion models to synthesize diverse environmental conditions. While these methods realize a preliminary visual simulation of weather or lighting editing, they are incapable of replicating the optical phenomena arising from environmental changes, such as scattering in foggy conditions or additional reflections generated during rainfall. Physics-based rendering emerges as a promising research direction to address these limitations(e.g., inverse rendering~\cite{bi2020neural, gardner2024sky, gs-ir, kaleta2024lumigauss, chen2025gigs, zhang2025scaling}). Such integration would allow the simulation of complex weather-induced effects with physical realism, such as scattering in fog, or reflections of the road surface on a rainy day. Bridging this gap is crucial for creating all-round simulators to enhance the robustness of autonomous driving systems under adverse weather and lighting conditions.

\textbf{3) Comprehensive Scene Feature}\label{sec:integration} Most current autonomous driving systems~\cite{wu2025bevgs,chen2025multi,wang2024towards,lang2024bev} choose Bird's-Eye View (BEV) as the comprehensive scene feature. However, as a lossy representation projected onto a 2D space, BEV discards spatial and visual details. As the architecture of autonomous driving systems has evolved from modular designs to end-to-end networks~\cite{hu2023planning, jiang2023vad}, and now to the mainstream Vision-Language-Action (VLA) models~\cite{zhou2025opendrivevla, zhou2025autovla}, the training paradigm has shifted towards multi-modal, multi-task learning. This requires the system not only to recognize-segment-track objects at a visual level, but also to understand, analyze, and reason about the driving scene in order to make timely and safe decisions. The BEV representation is clearly insufficient to meet the requirements for comprehensive scene features in autonomous driving systems. With advancements in 3D reconstruction, modern 3D representations can not only perform reconstruction but also serve directly as a compact and unified scene feature with not only visual details but also semantic features for system-level integration into autonomous driving systems, as demonstrated by \cite{yuan2024presight, yang2024unipad, zheng2024gaussianad}. As discussed in Section \ref{sec::data_collection}, 3D representations have expanded beyond merely capturing the high-fidelity geometry and appearance details required for rendering, now simultaneously encoding high-level information vital to autonomous driving systems, including semantics and flows. This fulfills the demands of modern autonomous driving architectures for vision-based scene understanding and multi-task training frameworks, enhancing the capability to handle complex and interactive scenarios, consequently promoting the intelligence, traffic efficiency, and safety of autonomous driving systems to reach new heights.

\textbf{4) On-board Validation}\label{sec:onboard} 3D reconstruction in autonomous driving has undergone a dramatic evolution in computational efficiency. Early NeRFs~\cite{ost2021nsg, yang2023emernerf} required several to tens of seconds to render a single frame, progressing to near real-time rendering with later neural point clouds~\cite{li2024dgnr}. Finally, with the advent of 3DGS, rendering speeds can reach hundreds of FPS~\cite{wu2024hgs, hwang2024vegs}. The increased computational efficiency creates a speed margin, allowing the rendering process to remain responsive even under higher workloads or limited computational resources.

However, the high rendering efficiency of these methods is built upon the substantial computational capacity of high-performance GPUs, deployment on on-board platform presents unique challenges, including limited computational resources and strict latency requirements. Bridging this gap demands representations and rendering paradigms custom-built for lightweight, heterogeneous hardware. Emerging rendering acceleration methods, such as Gaussian importance filtering~\cite{huang2025seele, liu2025mobilegaussian}, primitive compression~\cite{hu2024lowlatency}, and cloud-rendering~\cite{liu2025voyager}, provide a foundation for lightweight on-board deployment. However, existing 3D reconstruction methods, applications, and associated lightweighting techniques are all at the independent proof-of-concept stage. Integrating them into genuinely practical on-board autonomous driving systems demands future research efforts.

\textbf{5) Evaluating Safety Influence}\label{sec:safety}
3D reconstruction introduces new solutions for autonomous driving and consequently brings complex safety implications. On one hand, data acquisition and simulators based on 3D reconstruction~\cite{wu2023mars, ljungbergh2024neuroncap, zhou2024hugsim} have achieved significant improvements in fidelity, yet they still fall within the category of synthetic data and inevitably exhibit epistemic uncertainties, such as aliasing artifacts or edge blurring, reflecting the same lack of comprehensive safety validation arising from the sim-to-real gap inherent to autonomous driving systems trained on synthetic data. To address this, researchers should quantitatively evaluate their proposed methods, including reconstruction metrics (PNSR, Chamfer Distance, etc.) and distributional fidelity metrics (FID, MMD, etc.), to fundamentally measure and further reduce the discrepancy with the real world and avoid safety degradation. Regrettably, evaluation and validation of the safety impact of proposed methods are notably neglected in existing research, with only a very small number of works providing related reports~\cite{gao2025rad, zheng2024gaussianad, ljungbergh2024neuroncap, yan2025drivingsphere}. Despite pursuing technological advancements, future research and applications should strengthen their focus on safety validation and assessment.

On the other hand, 3D reconstruction technology has a comprehensive and heterogeneous impact on the safety of autonomous driving systems. For instance, the inherent epistemic uncertainty may disrupt perception accuracy, creating potential risks of misidentification. Conversely, 3D reconstruction-based closed-loop simulation~\cite{gao2025rad, ljungbergh2024neuroncap, yan2025drivingsphere} and comprehensive scene features~\cite{zheng2024gaussianad} enhance system robustness in rare scenarios by mitigating long-tail problems. Therefore, a holistic assessment of the autonomous driving system's overall safety and robustness should be conducted at the system level across various driving scenarios under the guidance from existing safety standards~\cite{ISO26262, ISO21448, koopman2023ul}. A special focus should be placed on the safety risks arising from functional insufficiency defined by ISO 21448~\cite{ISO21448} (SOTIF), using metrics such as Miles per Intervention or collision rates. Ultimately, mandatory comprehensive real-vehicle testing prior to the large-scale release of autonomous driving systems will serve as the most solid guarantee of safety, as stipulated by regulation~\cite{eu, japan}.

\bibliographystyle{IEEEtran}
\bibliography{IEEEabrv, main.bib}

\begin{thebibliography}{100}
\providecommand{\url}[1]{#1}
\csname url@samestyle\endcsname
\providecommand{\newblock}{\relax}
\providecommand{\bibinfo}[2]{#2}
\providecommand{\BIBentrySTDinterwordspacing}{\spaceskip=0pt\relax}
\providecommand{\BIBentryALTinterwordstretchfactor}{4}
\providecommand{\BIBentryALTinterwordspacing}{\spaceskip=\fontdimen2\font plus
\BIBentryALTinterwordstretchfactor\fontdimen3\font minus
  \fontdimen4\font\relax}
\providecommand{\BIBforeignlanguage}[2]{{%
\expandafter\ifx\csname l@#1\endcsname\relax
\typeout{** WARNING: IEEEtran.bst: No hyphenation pattern has been}%
\typeout{** loaded for the language `#1'. Using the pattern for}%
\typeout{** the default language instead.}%
\else
\language=\csname l@#1\endcsname
\fi
#2}}
\providecommand{\BIBdecl}{\relax}
\BIBdecl

\bibitem{nerf}
B.~Mildenhall, P.~P. Srinivasan, M.~Tancik, J.~T. Barron, R.~Ramamoorthi, and
  R.~Ng, ``Nerf: Representing scenes as neural radiance fields for view
  synthesis,'' in \emph{Proceedings of the European conference on computer
  vision (ECCV)}, 2020, pp. 405--421.

\bibitem{Mller2022InstantNGP}
T.~M{\"u}ller, A.~Evans, C.~Schied, and A.~Keller, ``Instant neural graphics
  primitives with a multiresolution hash encoding,'' \emph{ACM Transactions on
  Graphics (TOG)}, vol.~41, pp. 1 -- 15, 2022.

\bibitem{Barron2021MipNeRF}
J.~T. Barron, B.~Mildenhall, M.~Tancik, P.~Hedman, R.~Martin-Brualla, and P.~P.
  Srinivasan, ``Mip-nerf: A multiscale representation for anti-aliasing neural
  radiance fields,'' \emph{Proceedings of the IEEE/CVF International Conference
  on Computer Vision (ICCV)}, pp. 5835--5844, 2021.

\bibitem{Barron2021MipNeRF360}
J.~T. Barron, B.~Mildenhall, D.~Verbin, P.~P. Srinivasan, and P.~Hedman,
  ``Mip-nerf 360: Unbounded anti-aliased neural radiance fields,'' \emph{2022
  IEEE/CVF Conference on Computer Vision and Pattern Recognition (CVPR)}, pp.
  5460--5469, 2021.

\bibitem{feng2022neuralpoints}
W.~Feng, J.~Li, H.~Cai, X.~Luo, and J.~Zhang, ``Neural points: Point cloud
  representation with neural fields for arbitrary upsampling,'' in
  \emph{Proceedings of the Computer Vision and Pattern Recognition Conference
  (CVPR)}, 2022, pp. 18\,633--18\,642.

\bibitem{xu2022point}
Q.~Xu, Z.~Xu, J.~Philip, S.~Bi, Z.~Shu, K.~Sunkavalli, and U.~Neumann,
  ``Point-nerf: Point-based neural radiance fields,'' in \emph{Proceedings of
  the Computer Vision and Pattern Recognition Conference (CVPR)}, 2022, pp.
  5438--5448.

\bibitem{yang2022neumesh}
B.~Yang, C.~Bao, J.~Zeng, H.~Bao, Y.~Zhang, Z.~Cui, and G.~Zhang, ``Neumesh:
  Learning disentangled neural mesh-based implicit field for geometry and
  texture editing,'' in \emph{Proceedings of the European conference on
  computer vision (ECCV)}, 2022, pp. 597--614.

\bibitem{3dgs}
B.~Kerbl, G.~Kopanas, T.~Leimk{\"u}hler, and G.~Drettakis, ``3d gaussian
  splatting for real-time radiance field rendering.'' \emph{ACM Trans. Graph.},
  vol.~42, no.~4, pp. 139--1, 2023.

\bibitem{turki2023suds}
H.~Turki, J.~Y. Zhang, F.~Ferroni, and D.~Ramanan, ``Suds: Scalable urban
  dynamic scenes,'' in \emph{Proceedings of the Computer Vision and Pattern
  Recognition Conference (CVPR)}, 2023, pp. 12\,375--12\,385.

\bibitem{yuan2025uni}
Z.~Yuan, Y.~Pu, H.~Luo, F.~Lang, C.~Chi, T.~Li, Y.~Shen, H.~Sun, B.~Wang, and
  X.~Yang, ``Uni-gaussians: Unifying camera and lidar simulation with gaussians
  for dynamic driving scenarios,'' \emph{arXiv preprint arXiv:2503.08317},
  2025.

\bibitem{gao2025rad}
H.~Gao, S.~Chen, B.~Jiang, B.~Liao, Y.~Shi, X.~Guo, Y.~Pu, H.~Yin, X.~Li,
  X.~Zhang \emph{et~al.}, ``Rad: Training an end-to-end driving policy via
  large-scale 3dgs-based reinforcement learning,'' \emph{arXiv preprint
  arXiv:2502.13144}, 2025.

\bibitem{yuan2024presight}
T.~Yuan, Y.~Mao, J.~Yang, Y.~Liu, Y.~Wang, and H.~Zhao, ``Presight: Enhancing
  autonomous vehicle perception with city-scale nerf priors,'' in
  \emph{Proceedings of the European conference on computer vision (ECCV)},
  2024, pp. 323--339.

\bibitem{yang2024unipad}
H.~Yang, S.~Zhang, D.~Huang, X.~Wu, H.~Zhu, T.~He, S.~Tang, H.~Zhao, Q.~Qiu,
  B.~Lin \emph{et~al.}, ``Unipad: A universal pre-training paradigm for
  autonomous driving,'' in \emph{Proceedings of the Computer Vision and Pattern
  Recognition Conference (CVPR)}, 2024, pp. 15\,238--15\,250.

\bibitem{wu2024emie}
W.~Wu, Q.~Wang, G.~Wang, J.~Wang, T.~Zhao, Y.~Liu, D.~Gao, Z.~Liu, and H.~Wang,
  ``Emie-map: Large-scale road surface reconstruction based on explicit mesh
  and implicit encoding,'' in \emph{Proceedings of the European conference on
  computer vision (ECCV)}, 2024, pp. 370--386.

\bibitem{zheng2024gaussianad}
W.~Zheng, J.~Wu, Y.~Zheng, S.~Zuo, Z.~Xie, L.~Yang, Y.~Pan, Z.~Hao, P.~Jia,
  X.~Lang \emph{et~al.}, ``Gaussianad: Gaussian-centric end-to-end autonomous
  driving,'' \emph{arXiv preprint arXiv:2412.10371}, 2024.

\bibitem{yang2023unisim}
Z.~Yang, Y.~Chen, J.~Wang, S.~Manivasagam, W.-C. Ma, A.~J. Yang, and
  R.~Urtasun, ``Unisim: A neural closed-loop sensor simulator,'' in
  \emph{Proceedings of the Computer Vision and Pattern Recognition Conference
  (CVPR)}, 2023, pp. 1389--1399.

\bibitem{yan2024oasim}
G.~Yan, J.~Pi, J.~Guo, Z.~Luo, M.~Dou, N.~Deng, Q.~Huang, D.~Fu, L.~Wen, P.~Cai
  \emph{et~al.}, ``Oasim: An open and adaptive simulator based on neural
  rendering for autonomous driving,'' \emph{arXiv preprint arXiv:2402.03830},
  2024.

\bibitem{wu2023mars}
Z.~Wu, T.~Liu, L.~Luo, Z.~Zhong, J.~Chen, H.~Xiao, C.~Hou, H.~Lou, Y.~Chen,
  R.~Yang \emph{et~al.}, ``Mars: An instance-aware, modular and realistic
  simulator for autonomous driving,'' in \emph{CAAI International Conference on
  Artificial Intelligence}, 2023, pp. 3--15.

\bibitem{ljungbergh2024neuroncap}
W.~Ljungbergh, A.~Tonderski, J.~Johnander, H.~Caesar, K.~{\AA}str{\"o}m,
  M.~Felsberg, and C.~Petersson, ``Neuroncap: Photorealistic closed-loop safety
  testing for autonomous driving,'' in \emph{Proceedings of the European
  conference on computer vision (ECCV)}, 2024, pp. 161--177.

\bibitem{zhou2024hugsim}
H.~Zhou, L.~Lin, J.~Wang, Y.~Lu, D.~Bai, B.~Liu, Y.~Wang, A.~Geiger, and
  Y.~Liao, ``Hugsim: A real-time, photo-realistic and closed-loop simulator for
  autonomous driving,'' \emph{arXiv preprint arXiv:2412.01718}, 2024.

\bibitem{wei2024editable}
Y.~Wei, Z.~Wang, Y.~Lu, C.~Xu, C.~Liu, H.~Zhao, S.~Chen, and Y.~Wang,
  ``Editable scene simulation for autonomous driving via collaborative
  llm-agents,'' in \emph{Proceedings of the Computer Vision and Pattern
  Recognition Conference (CVPR)}, 2024, pp. 15\,077--15\,087.

\bibitem{he2024nerfsurvey}
L.~He, L.~Li, W.~Sun, Z.~Han, Y.~Liu, S.~Zheng, J.~Wang, and K.~Li, ``Neural
  radiance field in autonomous driving: A survey,'' \emph{arXiv preprint
  arXiv:2404.13816}, 2024.

\bibitem{bao20253dgssurvey}
Y.~Bao, T.~Ding, J.~Huo, Y.~Liu, Y.~Li, W.~Li, Y.~Gao, and J.~Luo, ``3d
  gaussian splatting: Survey, technologies, challenges, and opportunities,''
  \emph{IEEE Transactions on Circuits and Systems for Video Technology}, 2025.

\bibitem{zhu20243dgssurvey}
H.~Zhu, Z.~Zhang, J.~Zhao, H.~Duan, Y.~Ding, X.~Xiao, and J.~Yuan, ``Scene
  reconstruction techniques for autonomous driving: a review of 3d gaussian
  splatting,'' \emph{Artificial Intelligence Review}, vol.~58, no.~1, p.~30,
  2024.

\bibitem{bsurface}
A.~Arnal and J.~Monterde, ``Bézier-smart surfaces of arbitrary degree,''
  \emph{Journal of Computational and Applied Mathematics}, vol. 457, p. 116253,
  2025.

\bibitem{bsurface2}
Y.-X. Hao and W.-Q. Fei, ``Construction of bézier surfaces with minimal
  quadratic energy for given diagonal curves,'' \emph{Journal of Computational
  and Applied Mathematics}, vol. 446, p. 115854, 2024.

\bibitem{Rosli_Zulkifly_2023bsurface}
S.~N.~I. Rosli and M.~I.~E. Zulkifly, ``Neutrosophic bicubic bezier surface
  approximationmodel for uncertainty data,'' \emph{MATEMATIKA}, vol.~39, no.~3,
  p. 281–291, Dec. 2023.

\bibitem{nurbs1}
X.~Zou, S.~B. Lo, R.~Sevilla, O.~Hassan, and K.~Morgan, ``The generation of 3d
  surface meshes for nurbs-enhanced fem,'' \emph{Computer-Aided Design}, vol.
  168, p. 103653, 2024.

\bibitem{nurbs2}
N.~Grillanda, A.~Chiozzi, G.~Milani, and A.~Tralli, ``Nurbs solid modeling for
  the three-dimensional limit analysis of curved rigid block structures,''
  \emph{Computer Methods in Applied Mechanics and Engineering}, vol. 399, p.
  115304, 2022.

\bibitem{park2019deepsdf}
J.~J. Park, P.~Florence, J.~Straub, R.~Newcombe, and S.~Lovegrove, ``Deepsdf:
  Learning continuous signed distance functions for shape representation,'' in
  \emph{Proceedings of the Computer Vision and Pattern Recognition Conference
  (CVPR)}, 2019, pp. 165--174.

\bibitem{zhang2020nerf++}
K.~Zhang, G.~Riegler, N.~Snavely, and V.~Koltun, ``Nerf++: Analyzing and
  improving neural radiance fields,'' \emph{arXiv preprint arXiv:2010.07492},
  2020.

\bibitem{li2024dgnr}
Z.~Li, C.~Wu, L.~Zhang, and J.~Zhu, ``Dgnr: Density-guided neural point
  rendering of large driving scenes,'' \emph{IEEE Transactions on Automation
  Science and Engineering}, 2024.

\bibitem{zhang2023occformer}
Y.~Zhang, Z.~Zhu, and D.~Du, ``Occformer: Dual-path transformer for
  vision-based 3d semantic occupancy prediction,'' in \emph{Proceedings of the
  IEEE/CVF International Conference on Computer Vision (ICCV)}, 2023, pp.
  9433--9443.

\bibitem{wang2024panoocc}
Y.~Wang, Y.~Chen, X.~Liao, L.~Fan, and Z.~Zhang, ``Panoocc: Unified occupancy
  representation for camera-based 3d panoptic segmentation,'' in
  \emph{Proceedings of the Computer Vision and Pattern Recognition Conference
  (CVPR)}, 2024, pp. 17\,158--17\,168.

\bibitem{tang2024sparseocc}
P.~Tang, Z.~Wang, G.~Wang, J.~Zheng, X.~Ren, B.~Feng, and C.~Ma, ``Sparseocc:
  Rethinking sparse latent representation for vision-based semantic occupancy
  prediction,'' in \emph{Proceedings of the Computer Vision and Pattern
  Recognition Conference (CVPR)}, 2024, pp. 15\,035--15\,044.

\bibitem{yang2024adaptiveocc}
T.~Yang, Y.~Qian, W.~Yan, C.~Wang, and M.~Yang, ``Adaptiveocc: Adaptive
  octree-based network for multi-camera 3d semantic occupancy prediction in
  autonomous driving,'' \emph{IEEE Transactions on Circuits and Systems for
  Video Technology}, 2024.

\bibitem{xiao2024semantic}
H.~Xiao, W.~Kang, H.~Liu, Y.~Li, and Y.~He, ``Semantic scene completion via
  semantic-aware guidance and interactive refinement transformer,'' \emph{IEEE
  Transactions on Circuits and Systems for Video Technology}, 2024.

\bibitem{omniscene}
D.~Wei, Z.~Li, and P.~Liu, ``Omni-scene: Omni-gaussian representation for
  ego-centric sparse-view scene reconstruction,'' \emph{arXiv preprint
  arXiv:2412.06273}, 2024.

\bibitem{zhang2024geolrm}
C.~Zhang, H.~Song, Y.~Wei, C.~Yu, J.~Lu, and Y.~Tang, ``Geolrm: Geometry-aware
  large reconstruction model for high-quality 3d gaussian generation,''
  \emph{Advances in Neural Information Processing Systems}, vol.~37, pp.
  55\,761--55\,784, 2024.

\bibitem{lu2024scaffold}
T.~Lu, M.~Yu, L.~Xu, Y.~Xiangli, L.~Wang, D.~Lin, and B.~Dai, ``Scaffold-gs:
  Structured 3d gaussians for view-adaptive rendering,'' in \emph{Proceedings
  of the Computer Vision and Pattern Recognition Conference (CVPR)}, 2024, pp.
  20\,654--20\,664.

\bibitem{yu2024mipsplatting}
Z.~Yu, A.~Chen, B.~Huang, T.~Sattler, and A.~Geiger, ``Mip-splatting:
  Alias-free 3d gaussian splatting,'' in \emph{Proceedings of the Computer
  Vision and Pattern Recognition Conference (CVPR)}, 2024, pp.
  19\,447--19\,456.

\bibitem{Szymanowicz2023SplatterImage}
S.~Szymanowicz, C.~Rupprecht, and A.~Vedaldi, ``Splatter image: Ultra-fast
  single-view 3d reconstruction,'' \emph{2024 IEEE/CVF Conference on Computer
  Vision and Pattern Recognition (CVPR)}, pp. 10\,208--10\,217, 2023.

\bibitem{enwiki:1197924651}
{Wikipedia contributors}, ``Bézier surface --- {Wikipedia}{,} the free
  encyclopedia,''
  \url{https://en.wikipedia.org/w/index.php?title=B%C3%A9zier_surface&oldid=1197924651},
  2024, [Online; accessed 21-March-2025].

\bibitem{li2023diffusionsdf}
M.~Li, Y.~Duan, J.~Zhou, and J.~Lu, ``Diffusion-sdf: Text-to-shape via
  voxelized diffusion,'' in \emph{Proceedings of the Computer Vision and
  Pattern Recognition Conference (CVPR)}, 2023, pp. 12\,642--12\,651.

\bibitem{zhang2021ners}
J.~Zhang, G.~Yang, S.~Tulsiani, and D.~Ramanan, ``Ners: Neural reflectance
  surfaces for sparse-view 3d reconstruction in the wild,'' \emph{Advances in
  Neural Information Processing Systems}, vol.~34, pp. 29\,835--29\,847, 2021.

\bibitem{liu2024car}
T.~Liu, H.~Zhao, Y.~Yu, G.~Zhou, and M.~Liu, ``Car-studio: learning car
  radiance fields from single-view and unlimited in-the-wild images,''
  \emph{IEEE Robotics and Automation Letters}, vol.~9, no.~3, 2024.

\bibitem{neusim}
Z.~Yang, S.~Manivasagam, Y.~Chen, J.~Wang, R.~Hu, and R.~Urtasun,
  ``Reconstructing objects in-the-wild for realistic sensor simulation,'' in
  \emph{2023 IEEE International Conference on Robotics and Automation
  (ICRA)}.\hskip 1em plus 0.5em minus 0.4em\relax IEEE, 2023, pp.
  11\,661--11\,668.

\bibitem{du20243drealcar}
X.~Du, H.~Sun, S.~Wang, Z.~Wu, H.~Sheng, J.~Ying, M.~Lu, T.~Zhu, K.~Zhan, and
  X.~Yu, ``3drealcar: An in-the-wild rgb-d car dataset with 360-degree views,''
  \emph{arXiv preprint arXiv:2406.04875}, 2024.

\bibitem{h36m_pami}
C.~Ionescu, D.~Papava, V.~Olaru, and C.~Sminchisescu, ``Human3.6m: Large scale
  datasets and predictive methods for 3d human sensing in natural
  environments,'' \emph{IEEE Transactions on Pattern Analysis and Machine
  Intelligence}, vol.~36, no.~7, pp. 1325--1339, jul 2014.

\bibitem{human_interaction}
M.~Fieraru, M.~Zanfir, E.~Oneata, A.-I. Popa, V.~Olaru, and C.~Sminchisescu,
  ``Reconstructing three-dimensional models of interacting humans,''
  \emph{arXiv preprint arXiv:2308.01854}, 2023.

\bibitem{fit3d}
M.~Fieraru, M.~Zanfir, S.-C. Pirlea, V.~Olaru, and C.~Sminchisescu, ``Aifit:
  Automatic 3d human-interpretable feedback models for fitness training,'' in
  \emph{The IEEE/CVF Conference on Computer Vision and Pattern Recognition
  (CVPR)}, June 2021.

\bibitem{sc3d}
M.~Fieraru, M.~Zanfir, E.~Oneata, A.-I. Popa, V.~Olaru, and C.~Sminchisescu,
  ``Learning complex 3d human self-contact,'' in \emph{Proceedings of the AAAI
  Conference on Artificial Intelligence}, 2021.

\bibitem{3dpw}
T.~von Marcard, R.~Henschel, M.~Black, B.~Rosenhahn, and G.~Pons-Moll,
  ``Recovering accurate 3d human pose in the wild using imus and a moving
  camera,'' in \emph{Proceedings of the European conference on computer vision
  (ECCV)}, sep 2018.

\bibitem{Renderpeople2025}
\BIBentryALTinterwordspacing
{Renderpeople}, 2025. [Online]. Available: \url{https://renderpeople.com/}
\BIBentrySTDinterwordspacing

\bibitem{Geiger2012kitti}
A.~Geiger, P.~Lenz, and R.~Urtasun, ``Are we ready for autonomous driving? the
  kitti vision benchmark suite,'' in \emph{Conference on Computer Vision and
  Pattern Recognition (CVPR)}, 2012.

\bibitem{Cordts2016Cityscapes}
M.~Cordts, M.~Omran, S.~Ramos, T.~Rehfeld, M.~Enzweiler, R.~Benenson,
  U.~Franke, S.~Roth, and B.~Schiele, ``The cityscapes dataset for semantic
  urban scene understanding,'' in \emph{Proc. of the IEEE Conference on
  Computer Vision and Pattern Recognition (CVPR)}, 2016.

\bibitem{yu2020bdd100k}
F.~Yu, H.~Chen, X.~Wang, W.~Xian, Y.~Chen, F.~Liu, V.~Madhavan, and T.~Darrell,
  ``Bdd100k: A diverse driving dataset for heterogeneous multitask learning,''
  in \emph{Proceedings of the Computer Vision and Pattern Recognition
  Conference (CVPR)}, 2020, pp. 2636--2645.

\bibitem{behley2019semankitti}
J.~Behley, M.~Garbade, A.~Milioto, J.~Quenzel, S.~Behnke, C.~Stachniss, and
  J.~Gall, ``{SemanticKITTI: A Dataset for Semantic Scene Understanding of
  LiDAR Sequences},'' in \emph{Proc. of the IEEE/CVF International Conf.~on
  Computer Vision (ICCV)}, 2019.

\bibitem{nuscenes2019}
H.~Caesar, V.~Bankiti, A.~H. Lang, S.~Vora, V.~E. Liong, Q.~Xu, A.~Krishnan,
  Y.~Pan, G.~Baldan, and O.~Beijbom, ``nuscenes: A multimodal dataset for
  autonomous driving,'' \emph{arXiv preprint arXiv:1903.11027}, 2019.

\bibitem{sun2020waymo}
P.~Sun, H.~Kretzschmar, X.~Dotiwalla, A.~Chouard, V.~Patnaik, P.~Tsui, J.~Guo,
  Y.~Zhou, Y.~Chai, B.~Caine \emph{et~al.}, ``Scalability in perception for
  autonomous driving: Waymo open dataset,'' in \emph{Proceedings of the
  Computer Vision and Pattern Recognition Conference (CVPR)}, 2020, pp.
  2446--2454.

\bibitem{cabon2020vkitti2}
Y.~Cabon, N.~Murray, and M.~Humenberger, ``Virtual kitti 2,'' 2020.

\bibitem{Argoverse2}
B.~Wilson, W.~Qi, T.~Agarwal, J.~Lambert, J.~Singh, S.~Khandelwal, B.~Pan,
  R.~Kumar, A.~Hartnett, J.~K. Pontes \emph{et~al.}, ``Argoverse 2: Next
  generation datasets for self-driving perception and forecasting,''
  \emph{arXiv preprint arXiv:2301.00493}, 2023.

\bibitem{Liao2022kitti360}
Y.~Liao, J.~Xie, and A.~Geiger, ``{KITTI}-360: A novel dataset and benchmarks
  for urban scene understanding in 2d and 3d,'' \emph{Pattern Analysis and
  Machine Intelligence (PAMI)}, 2022.

\bibitem{yang2023emernerf}
J.~Yang, B.~Ivanovic, O.~Litany, X.~Weng, S.~W. Kim, B.~Li, T.~Che, D.~Xu,
  S.~Fidler, M.~Pavone \emph{et~al.}, ``Emernerf: Emergent spatial-temporal
  scene decomposition via self-supervision,'' \emph{arXiv preprint
  arXiv:2311.02077}, 2023.

\bibitem{cabon2020virtual}
Y.~Cabon, N.~Murray, and M.~Humenberger, ``Virtual kitti 2,'' \emph{arXiv
  preprint arXiv:2001.10773}, 2020.

\bibitem{argoverse}
M.-F. Chang, J.~Lambert, P.~Sangkloy, J.~Singh, S.~Bak, A.~Hartnett, D.~Wang,
  P.~Carr, S.~Lucey, D.~Ramanan \emph{et~al.}, ``Argoverse: 3d tracking and
  forecasting with rich maps,'' in \emph{Proceedings of the Computer Vision and
  Pattern Recognition Conference (CVPR)}, 2019, pp. 8748--8757.

\bibitem{Zhang2018lpips}
R.~Zhang, P.~Isola, A.~A. Efros, E.~Shechtman, and O.~Wang, ``The unreasonable
  effectiveness of deep features as a perceptual metric,'' \emph{Proceedings of
  the Computer Vision and Pattern Recognition Conference (CVPR)}, pp. 586--595,
  2018.

\bibitem{vgg}
K.~Simonyan and A.~Zisserman, ``Very deep convolutional networks for
  large-scale image recognition,'' \emph{arXiv preprint arXiv:1409.1556}, 2014.

\bibitem{Barratt2018inceptionscore}
S.~T. Barratt and R.~Sharma, ``A note on the inception score,'' \emph{ArXiv},
  vol. abs/1801.01973, 2018.

\bibitem{Heusel2017fid}
M.~Heusel, H.~Ramsauer, T.~Unterthiner, B.~Nessler, and S.~Hochreiter, ``Gans
  trained by a two time-scale update rule converge to a local nash
  equilibrium,'' in \emph{Neural Information Processing Systems}, 2017.

\bibitem{unterthiner2019fvd}
T.~Unterthiner, S.~Van~Steenkiste, K.~Kurach, R.~Marinier, M.~Michalski, and
  S.~Gelly, ``Fvd: A new metric for video generation,'' 2019.

\bibitem{mmd}
A.~Gretton, K.~M. Borgwardt, M.~J. Rasch, B.~Sch{\"o}lkopf, and A.~Smola, ``A
  kernel two-sample test,'' \emph{The Journal of Machine Learning Research},
  vol.~13, no.~1, pp. 723--773, 2012.

\bibitem{guo2023streetsurf}
J.~Guo, N.~Deng, X.~Li, Y.~Bai, B.~Shi, C.~Wang, C.~Ding, D.~Wang, and Y.~Li,
  ``Streetsurf: Extending multi-view implicit surface reconstruction to street
  views,'' \emph{arXiv preprint arXiv:2306.04988}, 2023.

\bibitem{wang2023planerf}
F.~Wang, A.~Louys, N.~Piasco, M.~Bennehar, L.~Rold{\~a}o, and D.~Tsishkou,
  ``Planerf: Svd unsupervised 3d plane regularization for nerf large-scale
  scene reconstruction,'' \emph{arXiv preprint arXiv:2305.16914}, 2023.

\bibitem{tao2024alignmif}
T.~Tao, G.~Wang, Y.~Lao, P.~Chen, J.~Liu, L.~Lin, K.~Yu, and X.~Liang,
  ``Alignmif: Geometry-aligned multimodal implicit field for lidar-camera joint
  synthesis,'' in \emph{Proceedings of the Computer Vision and Pattern
  Recognition Conference (CVPR)}, 2024, pp. 21\,230--21\,240.

\bibitem{weng2021aiodrive}
X.~Weng, Y.~Man, J.~Park, Y.~Yuan, M.~O'Toole, and K.~M. Kitani, ``All-in-one
  drive: A comprehensive perception dataset with high-density long-range point
  clouds,'' 2021.

\bibitem{ost2022neurallightfield}
J.~Ost, I.~Laradji, A.~Newell, Y.~Bahat, and F.~Heide, ``Neural point light
  fields,'' in \emph{Proceedings of the Computer Vision and Pattern Recognition
  Conference (CVPR)}, 2022, pp. 18\,419--18\,429.

\bibitem{lu2023urban}
F.~Lu, Y.~Xu, G.~Chen, H.~Li, K.-Y. Lin, and C.~Jiang, ``Urban radiance field
  representation with deformable neural mesh primitives,'' in \emph{Proceedings
  of the IEEE/CVF International Conference on Computer Vision (ICCV)}, 2023,
  pp. 465--476.

\bibitem{mei2024rome}
R.~Mei, W.~Sui, J.~Zhang, X.~Qin, G.~Wang, T.~Peng, T.~Chen, and C.~Yang,
  ``Rome: Towards large scale road surface reconstruction via mesh
  representation,'' \emph{IEEE Transactions on Intelligent Vehicles}, 2024.

\bibitem{wu2024hgs}
K.~Wu, K.~Zhang, Z.~Zhang, M.~Tie, S.~Yuan, J.~Zhao, Z.~Gan, and W.~Ding,
  ``Hgs-mapping: Online dense mapping using hybrid gaussian representation in
  urban scenes,'' \emph{IEEE Robotics and Automation Letters}, 2024.

\bibitem{cui2024streetsurfgs}
X.~Cui, W.~Ye, Y.~Wang, G.~Zhang, W.~Zhou, and H.~Li, ``Streetsurfgs: Scalable
  urban street surface reconstruction with planar-based gaussian splatting,''
  \emph{arXiv preprint arXiv:2410.04354}, 2024.

\bibitem{wang2023f2}
P.~Wang, Y.~Liu, Z.~Chen, L.~Liu, Z.~Liu, T.~Komura, C.~Theobalt, and W.~Wang,
  ``F2-nerf: Fast neural radiance field training with free camera
  trajectories,'' in \emph{Proceedings of the Computer Vision and Pattern
  Recognition Conference (CVPR)}, 2023, pp. 4150--4159.

\bibitem{xu2024unveiler}
J.~Xu, Y.~Wang, Y.~Zhao, Y.~Fu, and S.~Gao, ``3d streetunveiler with
  semantic-aware 2dgs,'' \emph{arXiv preprint arXiv:2405.18416}, 2024.

\bibitem{xiao2021pandaset}
P.~Xiao, Z.~Shao, S.~Hao, Z.~Zhang, X.~Chai, J.~Jiao, Z.~Li, J.~Wu, K.~Sun,
  K.~Jiang \emph{et~al.}, ``Pandaset: Advanced sensor suite dataset for
  autonomous driving,'' in \emph{2021 IEEE international intelligent
  transportation systems conference (ITSC)}.\hskip 1em plus 0.5em minus
  0.4em\relax IEEE, 2021, pp. 3095--3101.

\bibitem{song2024gvkf}
G.~Song, C.~Cheng, and H.~Wang, ``Gvkf: Gaussian voxel kernel functions for
  highly efficient surface reconstruction in open scenes,'' \emph{Advances in
  Neural Information Processing Systems}, vol.~37, pp. 104\,792--104\,815,
  2024.

\bibitem{shi2024dhgs}
X.~Shi, L.~Chen, P.~Wei, X.~Wu, T.~Jiang, Y.~Luo, and L.~Xie, ``Dhgs: Decoupled
  hybrid gaussian splatting for driving scene,'' \emph{arXiv preprint
  arXiv:2407.16600}, 2024.

\bibitem{feng2024rogs}
Z.~Feng, W.~Wu, T.~Deng, and H.~Wang, ``Rogs: Large scale road surface
  reconstruction with meshgrid gaussian,'' 2024.

\bibitem{wang2022cadsim}
J.~Wang, S.~Manivasagam, Y.~Chen, Z.~Yang, I.~A. B{\^a}rsan, A.~J. Yang, W.-C.
  Ma, and R.~Urtasun, ``Cadsim: Robust and scalable in-the-wild 3d
  reconstruction for controllable sensor simulation,'' in \emph{6th Annual
  Conference on Robot Learning}, 2022.

\bibitem{Wang2021NeuS}
P.~Wang, L.~Liu, Y.~Liu, C.~Theobalt, T.~Komura, and W.~Wang, ``Neus: Learning
  neural implicit surfaces by volume rendering for multi-view reconstruction,''
  \emph{ArXiv}, vol. abs/2106.10689, 2021.

\bibitem{shen2021deep}
T.~Shen, J.~Gao, K.~Yin, M.-Y. Liu, and S.~Fidler, ``Deep marching tetrahedra:
  a hybrid representation for high-resolution 3d shape synthesis,''
  \emph{Advances in Neural Information Processing Systems}, vol.~34, pp.
  6087--6101, 2021.

\bibitem{du2024dreamcar}
X.~Du, H.~Sun, M.~Lu, T.~Zhu, and X.~Yu, ``Dreamcar: Leveraging car-specific
  prior for in-the-wild 3d car reconstruction,'' \emph{IEEE Robotics and
  Automation Letters}, 2024.

\bibitem{GenAssets}
Z.~Yang, J.~Wang, H.~Zhang, S.~Manivasagam, Y.~Chen, and R.~Urtasun,
  ``Genassets: Generating in-the-wild 3d assets in latent space,'' in
  \emph{Proceedings of the Computer Vision and Pattern Recognition Conference
  (CVPR)}, June 2025, pp. 22\,392--22\,403.

\bibitem{IDOL}
Y.~Zhuang, J.~Lv, H.~Wen, Q.~Shuai, A.~Zeng, H.~Zhu, S.~Chen, Y.~Yang, X.~Cao,
  and W.~Liu, ``Idol: Instant photorealistic 3d human creation from a single
  image,'' in \emph{Proceedings of the Computer Vision and Pattern Recognition
  Conference (CVPR)}, June 2025, pp. 26\,308--26\,319.

\bibitem{li2023read}
Z.~Li, L.~Li, and J.~Zhu, ``Read: Large-scale neural scene rendering for
  autonomous driving,'' in \emph{Proceedings of the AAAI Conference on
  Artificial Intelligence}, vol.~37, no.~2, 2023, pp. 1522--1529.

\bibitem{ligocki2020brno}
A.~Ligocki, A.~Jelinek, and L.~Zalud, ``Brno urban dataset-the new data for
  self-driving agents and mapping tasks,'' in \emph{2020 IEEE International
  Conference on Robotics and Automation (ICRA)}.\hskip 1em plus 0.5em minus
  0.4em\relax IEEE, 2020, pp. 3284--3290.

\bibitem{ost2021nsg}
J.~Ost, F.~Mannan, N.~Thuerey, J.~Knodt, and F.~Heide, ``Neural scene graphs
  for dynamic scenes,'' in \emph{Proceedings of the Computer Vision and Pattern
  Recognition Conference (CVPR)}, 2021, pp. 2856--2865.

\bibitem{kundu2022panoptic}
A.~Kundu, K.~Genova, X.~Yin, A.~Fathi, C.~Pantofaru, L.~J. Guibas,
  A.~Tagliasacchi, F.~Dellaert, and T.~Funkhouser, ``Panoptic neural fields: A
  semantic object-aware neural scene representation,'' in \emph{Proceedings of
  the Computer Vision and Pattern Recognition Conference (CVPR)}, 2022, pp.
  12\,871--12\,881.

\bibitem{tonderski2024neurad}
A.~Tonderski, C.~Lindstr{\"o}m, G.~Hess, W.~Ljungbergh, L.~Svensson, and
  C.~Petersson, ``Neurad: Neural rendering for autonomous driving,'' in
  \emph{Proceedings of the Computer Vision and Pattern Recognition Conference
  (CVPR)}, 2024, pp. 14\,895--14\,904.

\bibitem{zod}
M.~Alibeigi, W.~Ljungbergh, A.~Tonderski, G.~Hess, A.~Lilja, C.~Lindstr{\"o}m,
  D.~Motorniuk, J.~Fu, J.~Widahl, and C.~Petersson, ``Zenseact open dataset: A
  large-scale and diverse multimodal dataset for autonomous driving,'' in
  \emph{Proceedings of the IEEE/CVF International Conference on Computer Vision
  (ICCV)}, 2023, pp. 20\,178--20\,188.

\bibitem{deng2023prosgnerf}
T.~Deng, S.~Liu, X.~Wang, Y.~Liu, D.~Wang, and W.~Chen, ``Prosgnerf:
  Progressive dynamic neural scene graph with frequency modulated auto-encoder
  in urban scenes,'' \emph{arXiv preprint arXiv:2312.09076}, 2023.

\bibitem{choi2024dico}
J.~Choi, G.~Hwang, and S.~J. Lee, ``Dico-nerf: Difference of cosine similarity
  for neural rendering of fisheye driving scenes,'' in \emph{Proceedings of the
  Computer Vision and Pattern Recognition Conference (CVPR)}, 2024, pp.
  7850--7858.

\bibitem{jbnu}
E.~Son, J.~Choi, J.~Song, Y.~Jin, and S.~J. Lee, ``Monocular depth estimation
  from a fisheye camera based on knowledge distillation,'' \emph{Sensors},
  vol.~23, no.~24, p. 9866, 2023.

\bibitem{xie2023s}
Z.~Xie, J.~Zhang, W.~Li, F.~Zhang, and L.~Zhang, ``S-nerf: Neural radiance
  fields for street views,'' \emph{arXiv preprint arXiv:2303.00749}, 2023.

\bibitem{chen2025s}
Y.~Chen, J.~Zhang, Z.~Xie, W.~Li, F.~Zhang, J.~Lu, and L.~Zhang, ``S-nerf++:
  Autonomous driving simulation via neural reconstruction and generation,''
  \emph{IEEE Transactions on Pattern Analysis and Machine Intelligence}, 2025.

\bibitem{zhou2024drivinggaussian}
X.~Zhou, Z.~Lin, X.~Shan, Y.~Wang, D.~Sun, and M.-H. Yang, ``Drivinggaussian:
  Composite gaussian splatting for surrounding dynamic autonomous driving
  scenes,'' in \emph{Proceedings of the Computer Vision and Pattern Recognition
  Conference (CVPR)}, 2024, pp. 21\,634--21\,643.

\bibitem{li2024ggrt}
H.~Li, Y.~Gao, C.~Wu, D.~Zhang, Y.~Dai, C.~Zhao, H.~Feng, E.~Ding, J.~Wang, and
  J.~Han, ``Ggrt: Towards pose-free generalizable 3d gaussian splatting in
  real-time,'' in \emph{Proceedings of the European conference on computer
  vision (ECCV)}, 2024, pp. 325--341.

\bibitem{zhao2024tclc}
C.~Zhao, S.~Sun, R.~Wang, Y.~Guo, J.-J. Wan, Z.~Huang, X.~Huang, Y.~V. Chen,
  and L.~Ren, ``Tclc-gs: Tightly coupled lidar-camera gaussian splatting for
  autonomous driving: Supplementary materials,'' in \emph{Proceedings of the
  European conference on computer vision (ECCV)}, 2024, pp. 91--106.

\bibitem{yu2024sgd}
Z.~Yu, H.~Wang, J.~Yang, H.~Wang, Z.~Xie, Y.~Cai, J.~Cao, Z.~Ji, and M.~Sun,
  ``Sgd: Street view synthesis with gaussian splatting and diffusion prior,''
  \emph{arXiv preprint arXiv:2403.20079}, 2024.

\bibitem{li2024ho}
Z.~Li, Y.~Zhang, C.~Wu, J.~Zhu, and L.~Zhang, ``Ho-gaussian: Hybrid
  optimization of 3d gaussian splatting for urban scenes,'' in
  \emph{Proceedings of the European conference on computer vision (ECCV)},
  2024, pp. 19--36.

\bibitem{huang2024s3}
N.~Huang, X.~Wei, W.~Zheng, P.~An, M.~Lu, W.~Zhan, M.~Tomizuka, K.~Keutzer, and
  S.~Zhang, ``\textit{$S^3$} gaussian: Self-supervised street gaussians for
  autonomous driving,'' \emph{arXiv preprint arXiv:2405.20323}, 2024.

\bibitem{chen2023periodic}
Y.~Chen, C.~Gu, J.~Jiang, X.~Zhu, and L.~Zhang, ``Periodic vibration gaussian:
  Dynamic urban scene reconstruction and real-time rendering,'' \emph{arXiv
  preprint arXiv:2311.18561}, 2023.

\bibitem{yan2024streetgaussian}
Y.~Yan, H.~Lin, C.~Zhou, W.~Wang, H.~Sun, K.~Zhan, X.~Lang, X.~Zhou, and
  S.~Peng, ``Street gaussians: Modeling dynamic urban scenes with gaussian
  splatting,'' in \emph{Proceedings of the European conference on computer
  vision (ECCV)}, 2024, pp. 156--173.

\bibitem{chen2024omnire}
Z.~Chen, J.~Yang, J.~Huang, R.~de~Lutio, J.~M. Esturo, B.~Ivanovic, O.~Litany,
  Z.~Gojcic, S.~Fidler, M.~Pavone \emph{et~al.}, ``Omnire: Omni urban scene
  reconstruction,'' \emph{arXiv preprint arXiv:2408.16760}, 2024.

\bibitem{peng2024desire}
C.~Peng, C.~Zhang, Y.~Wang, C.~Xu, Y.~Xie, W.~Zheng, K.~Keutzer, M.~Tomizuka,
  and W.~Zhan, ``Desire-gs: 4d street gaussians for static-dynamic
  decomposition and surface reconstruction for urban driving scenes,''
  \emph{arXiv preprint arXiv:2411.11921}, 2024.

\bibitem{li2024vdg}
H.~Li, J.~Li, D.~Zhang, C.~Wu, J.~Shi, C.~Zhao, H.~Feng, E.~Ding, J.~Wang, and
  J.~Han, ``Vdg: vision-only dynamic gaussian for driving simulation,''
  \emph{arXiv preprint arXiv:2406.18198}, 2024.

\bibitem{zhou2024hugs}
H.~Zhou, J.~Shao, L.~Xu, D.~Bai, W.~Qiu, B.~Liu, Y.~Wang, A.~Geiger, and
  Y.~Liao, ``Hugs: Holistic urban 3d scene understanding via gaussian
  splatting,'' in \emph{Proceedings of the Computer Vision and Pattern
  Recognition Conference (CVPR)}, 2024, pp. 21\,336--21\,345.

\bibitem{hwang2024vegs}
S.~Hwang, M.-J. Kim, T.~Kang, J.~Kang, and J.~Choo, ``Vegs: View extrapolation
  of urban scenes in 3d gaussian splatting using learned priors,'' in
  \emph{Proceedings of the European conference on computer vision (ECCV)},
  2024, pp. 1--18.

\bibitem{khan2024autosplat}
M.~Khan, H.~Fazlali, D.~Sharma, T.~Cao, D.~Bai, Y.~Ren, and B.~Liu,
  ``Autosplat: Constrained gaussian splatting for autonomous driving scene
  reconstruction,'' \emph{arXiv preprint arXiv:2407.02598}, 2024.

\bibitem{han2024ggs}
H.~Han, K.~Zhou, X.~Long, Y.~Wang, and C.~Xiao, ``Ggs: Generalizable gaussian
  splatting for lane switching in autonomous driving,'' \emph{arXiv preprint
  arXiv:2409.02382}, 2024.

\bibitem{zhao2024drivedreamer4d}
G.~Zhao, C.~Ni, X.~Wang, Z.~Zhu, X.~Zhang, Y.~Wang, G.~Huang, X.~Chen, B.~Wang,
  Y.~Zhang \emph{et~al.}, ``Drivedreamer4d: World models are effective data
  machines for 4d driving scene representation,'' \emph{arXiv preprint
  arXiv:2410.13571}, 2024.

\bibitem{tian2024drivingforward}
Q.~Tian, X.~Tan, Y.~Xie, and L.~Ma, ``Drivingforward: Feed-forward 3d gaussian
  splatting for driving scene reconstruction from flexible surround-view
  input,'' \emph{arXiv preprint arXiv:2409.12753}, 2024.

\bibitem{hess2024splatad}
G.~Hess, C.~Lindstr{\"o}m, M.~Fatemi, C.~Petersson, and L.~Svensson, ``Splatad:
  Real-time lidar and camera rendering with 3d gaussian splatting for
  autonomous driving,'' \emph{arXiv preprint arXiv:2411.16816}, 2024.

\bibitem{mao2024dreamdrive}
J.~Mao, B.~Li, B.~Ivanovic, Y.~Chen, Y.~Wang, Y.~You, C.~Xiao, D.~Xu,
  M.~Pavone, and Y.~Wang, ``Dreamdrive: Generative 4d scene modeling from
  street view images,'' \emph{arXiv preprint arXiv:2501.00601}, 2024.

\bibitem{yang2024storm}
J.~Yang, J.~Huang, Y.~Chen, Y.~Wang, B.~Li, Y.~You, A.~Sharma, M.~Igl,
  P.~Karkus, D.~Xu \emph{et~al.}, ``Storm: Spatio-temporal reconstruction model
  for large-scale outdoor scenes,'' \emph{arXiv preprint arXiv:2501.00602},
  2024.

\bibitem{wei2024emd}
X.~Wei, Q.~Wuwu, Z.~Zhao, Z.~Wu, N.~Huang, M.~Lu, N.~Ma, and S.~Zhang, ``Emd:
  Explicit motion modeling for high-quality street gaussian splatting,''
  \emph{arXiv preprint arXiv:2411.15582}, 2024.

\bibitem{li2024uniscene}
B.~Li, J.~Guo, H.~Liu, Y.~Zou, Y.~Ding, X.~Chen, H.~Zhu, F.~Tan, C.~Zhang,
  T.~Wang \emph{et~al.}, ``Uniscene: Unified occupancy-centric driving scene
  generation,'' \emph{arXiv preprint arXiv:2412.05435}, 2024.

\bibitem{gao2024magicdrive3d}
R.~Gao, K.~Chen, Z.~Li, L.~Hong, Z.~Li, and Q.~Xu, ``Magicdrive3d: Controllable
  3d generation for any-view rendering in street scenes,'' \emph{arXiv preprint
  arXiv:2405.14475}, 2024.

\bibitem{xiong2025drivinggaussian++}
Y.~Xiong, X.~Zhou, Y.~Wan, D.~Sun, and M.-H. Yang, ``Drivinggaussian++: Towards
  realistic reconstruction and editable simulation for surrounding dynamic
  driving scenes,'' \emph{arXiv preprint arXiv:2508.20965}, 2025.

\bibitem{yangrecovering}
Z.~Yang, S.~Manivasagam, M.~Liang, B.~Yang, W.-C. Ma, and R.~Urtasun,
  ``Recovering and simulating pedestrians in the wild,'' in \emph{Proceedings
  of the 2020 Conference on Robot Learning}, ser. Proceedings of Machine
  Learning Research, 16--18 Nov 2021, pp. 419--431.

\bibitem{Yanghumans3}
Z.~Yang, S.~Wang, S.~Manivasagam, Z.~Huang, W.-C. Ma, X.~Yan, E.~Yumer, and
  R.~Urtasun, ``S3: Neural shape, skeleton, and skinning fields for 3d human
  modeling,'' in \emph{Proceedings of the Computer Vision and Pattern
  Recognition Conference (CVPR)}, June 2021, pp. 13\,284--13\,293.

\bibitem{Xiephysical}
K.~Xie, T.~Wang, U.~Iqbal, Y.~Guo, S.~Fidler, and F.~Shkurti, ``Physics-based
  human motion estimation and synthesis from videos,'' in \emph{Proceedings of
  the IEEE/CVF International Conference on Computer Vision (ICCV)}, October
  2021, pp. 11\,532--11\,541.

\bibitem{li2022tava}
R.~Li, J.~Tanke, M.~Vo, M.~Zollh{\"o}fer, J.~Gall, A.~Kanazawa, and C.~Lassner,
  ``Tava: Template-free animatable volumetric actors,'' in \emph{Proceedings of
  the European conference on computer vision (ECCV)}, 2022, pp. 419--436.

\bibitem{guo2023vid2avatar}
C.~Guo, T.~Jiang, X.~Chen, J.~Song, and O.~Hilliges, ``Vid2avatar: 3d avatar
  reconstruction from videos in the wild via self-supervised scene
  decomposition,'' in \emph{Proceedings of the Computer Vision and Pattern
  Recognition Conference (CVPR)}, 2023, pp. 12\,858--12\,868.

\bibitem{jiang2023instantavatar}
T.~Jiang, X.~Chen, J.~Song, and O.~Hilliges, ``Instantavatar: Learning avatars
  from monocular video in 60 seconds,'' in \emph{Proceedings of the Computer
  Vision and Pattern Recognition Conference (CVPR)}, 2023, pp.
  16\,922--16\,932.

\bibitem{qian20243dgsavatar}
Z.~Qian, S.~Wang, M.~Mihajlovic, A.~Geiger, and S.~Tang, ``3dgs-avatar:
  Animatable avatars via deformable 3d gaussian splatting,'' in
  \emph{Proceedings of the Computer Vision and Pattern Recognition Conference
  (CVPR)}, 2024, pp. 5020--5030.

\bibitem{li2024animatable}
Z.~Li, Z.~Zheng, L.~Wang, and Y.~Liu, ``Animatable gaussians: Learning
  pose-dependent gaussian maps for high-fidelity human avatar modeling,'' in
  \emph{Proceedings of the Computer Vision and Pattern Recognition Conference
  (CVPR)}, 2024, pp. 19\,711--19\,722.

\bibitem{lee2025geoavatar}
S.~Lee, S.~Kim, H.~Lee, W.-S. Jeong, and J.~H. Lee, ``Geoavatar:
  Geometrically-consistent multi-person avatar reconstruction from sparse
  multi-view videos,'' in \emph{Proceedings of the Computer Vision and Pattern
  Recognition Conference (CVPR)}, 2025, pp. 21\,138--21\,147.

\bibitem{zhu2024rpbg}
Q.~Zhu, Z.~Wei, Z.~Zheng, Y.~Zhan, Z.~Yao, J.~Zhang, K.~Wu, and Y.~Zheng,
  ``Rpbg: Towards robust neural point-based graphics in the wild,'' \emph{arXiv
  preprint arXiv:2405.05663}, 2024.

\bibitem{shapenet2015}
A.~X. Chang, T.~Funkhouser, L.~Guibas, P.~Hanrahan, Q.~Huang, Z.~Li,
  S.~Savarese, M.~Savva, S.~Song, H.~Su, J.~Xiao, L.~Yi, and F.~Yu,
  ``{ShapeNet: An Information-Rich 3D Model Repository},'' Stanford University
  --- Princeton University --- Toyota Technological Institute at Chicago, Tech.
  Rep. arXiv:1512.03012 [cs.GR], 2015.

\bibitem{xiao2024liv}
R.~Xiao, W.~Liu, Y.~Chen, and L.~Hu, ``Liv-gs: Lidar-vision integration for 3d
  gaussian splatting slam in outdoor environments,'' \emph{IEEE Robotics and
  Automation Letters}, 2024.

\bibitem{zhou2025gsgvins}
Z.~Zhou, S.~Uprety, S.~Nie, and H.~Yang, ``Gs-gvins: A tightly-integrated
  gnss-visual-inertial navigation system augmented by 3d gaussian splatting,''
  2025.

\bibitem{xie2024gs}
Y.~Xie, Z.~Huang, J.~Wu, and J.~Ma, ``Gs-livm: Real-time photo-realistic
  lidar-inertial-visual mapping with gaussian splatting,'' \emph{arXiv preprint
  arXiv:2410.17084}, 2024.

\bibitem{wu2025bevgs}
W.~Wu, T.~Zhao, C.~Peng, L.~Yang, Y.~Wei, Z.~Liu, and H.~Wang, ``Bev-gs:
  Feed-forward gaussian splatting in bird's-eye-view for road reconstruction,''
  2025.

\bibitem{zhang2024vision}
J.~Zhang, S.~Chen, H.~Yin, R.~Mei, X.~Liu, C.~Yang, Q.~Zhang, and W.~Sui, ``A
  vision-centric approach for static map element annotation,'' in \emph{2024
  IEEE International Conference on Robotics and Automation (ICRA)}.\hskip 1em
  plus 0.5em minus 0.4em\relax IEEE, 2024, pp. 15\,861--15\,867.

\bibitem{wang2024nero}
R.~Wang, S.~Zhang, P.~Huang, D.~Zhang, and H.~Chen, ``Nero: Neural road surface
  reconstruction,'' 2024.

\bibitem{chen2024camav2}
\BIBentryALTinterwordspacing
S.~Chen, J.~Zhang, R.~Mei, Y.~Cai, H.~Yin, T.~Chen, W.~Sui, and C.~Yang,
  ``Camav2: A vision-centric approach for static map element annotation,''
  2024. [Online]. Available: \url{https://arxiv.org/abs/2407.21331}
\BIBentrySTDinterwordspacing

\bibitem{dosovitskiy2017carla}
A.~Dosovitskiy, G.~Ros, F.~Codevilla, A.~Lopez, and V.~Koltun, ``Carla: An open
  urban driving simulator,'' in \emph{Conference on robot learning}.\hskip 1em
  plus 0.5em minus 0.4em\relax PMLR, 2017, pp. 1--16.

\bibitem{wang2022learning}
T.-H. Wang, A.~Amini, W.~Schwarting, I.~Gilitschenski, S.~Karaman, and D.~Rus,
  ``Learning interactive driving policies via data-driven simulation,'' in
  \emph{2022 International Conference on Robotics and Automation (ICRA)}.\hskip
  1em plus 0.5em minus 0.4em\relax IEEE, 2022.

\bibitem{viworldsim}
VI-grade, ``Vi-worldsim,''
  \url{https://www.vi-grade.com/en/products/vi-worldsim/}, accessed Oct, 2025.

\bibitem{elmquist2024methodology}
A.~Elmquist, R.~Serban, and D.~Negrut, ``A methodology to quantify
  simulation-vs-reality differences in images for autonomous robots,''
  \emph{IEEE Sensors Journal}, 2024.

\bibitem{pahk2023effects}
J.~Pahk, J.~Shim, M.~Baek, Y.~Lim, and G.~Choi, ``Effects of sim2real image
  translation via dclgan on lane keeping assist system in carla simulator,''
  \emph{IEEE Access}, vol.~11, pp. 33\,915--33\,927, 2023.

\bibitem{pasios2025carla2real}
S.~Pasios and N.~Nikolaidis, ``Carla2real: A tool for reducing the sim2real
  appearance gap in carla simulator,'' \emph{IEEE Transactions on Intelligent
  Transportation Systems}, 2025.

\bibitem{yan2025drivingsphere}
T.~Yan, D.~Wu, W.~Han, J.~Jiang, X.~Zhou, K.~Zhan, C.-z. Xu, and J.~Shen,
  ``Drivingsphere: Building a high-fidelity 4d world for closed-loop
  simulation,'' in \emph{Proceedings of the Computer Vision and Pattern
  Recognition Conference}, 2025, pp. 27\,531--27\,541.

\bibitem{wang2025vggt}
J.~Wang, M.~Chen, N.~Karaev, A.~Vedaldi, C.~Rupprecht, and D.~Novotny, ``Vggt:
  Visual geometry grounded transformer,'' in \emph{Proceedings of the Computer
  Vision and Pattern Recognition Conference (CVPR)}, 2025, pp. 5294--5306.

\bibitem{zhou2024behaviorgpt}
Z.~Zhou, H.~Haibo, X.~Chen, J.~Wang, N.~Guan, K.~Wu, Y.-H. Li, Y.-K. Huang, and
  C.~J. Xue, ``Behaviorgpt: Smart agent simulation for autonomous driving with
  next-patch prediction,'' \emph{Advances in Neural Information Processing
  Systems}, vol.~37, pp. 79\,597--79\,617, 2024.

\bibitem{wu2024smart}
W.~Wu, X.~Feng, Z.~Gao, and Y.~Kan, ``Smart: Scalable multi-agent real-time
  motion generation via next-token prediction,'' \emph{Advances in Neural
  Information Processing Systems}, vol.~37, pp. 114\,048--114\,071, 2024.

\bibitem{bruce2024genie}
J.~Bruce, M.~D. Dennis, A.~Edwards, J.~Parker-Holder, Y.~Shi, E.~Hughes,
  M.~Lai, A.~Mavalankar, R.~Steigerwald, C.~Apps \emph{et~al.}, ``Genie:
  Generative interactive environments,'' in \emph{Forty-first International
  Conference on Machine Learning}, 2024.

\bibitem{rtfm}
W.~Labs, ``Rtfm: A real-time frame model,'' [Online],
  \url{https://www.worldlabs.ai/blog/rtfm}.

\bibitem{zhou2025hermesunifiedselfdrivingworld}
\BIBentryALTinterwordspacing
X.~Zhou, D.~Liang, S.~Tu, X.~Chen, Y.~Ding, D.~Zhang, F.~Tan, H.~Zhao, and
  X.~Bai, ``Hermes: A unified self-driving world model for simultaneous 3d
  scene understanding and generation,'' 2025. [Online]. Available:
  \url{https://arxiv.org/abs/2501.14729}
\BIBentrySTDinterwordspacing

\bibitem{sang2025weather}
C.~Sang, Y.~Qian, J.~Zhang, C.~Wang, and M.~Yang, ``Weather-magician:
  Reconstruction and rendering framework for 4d weather synthesis in real
  time,'' \emph{arXiv preprint arXiv:2505.19919}, 2025.

\bibitem{qian2025weatheredit}
C.~Qian, W.~Li, Y.~Guo, and G.~Markkula, ``Weatheredit: Controllable weather
  editing with 4d gaussian field,'' \emph{arXiv preprint arXiv:2505.20471},
  2025.

\bibitem{bi2020neural}
S.~Bi, Z.~Xu, P.~Srinivasan, B.~Mildenhall, K.~Sunkavalli, M.~Ha{\v{s}}an,
  Y.~Hold-Geoffroy, D.~Kriegman, and R.~Ramamoorthi, ``Neural reflectance
  fields for appearance acquisition,'' \emph{arXiv preprint arXiv:2008.03824},
  2020.

\bibitem{gardner2024sky}
J.~A. Gardner, E.~Kashin, B.~Egger, and W.~A. Smith, ``The sky’s the limit:
  Relightable outdoor scenes via a sky-pixel constrained illumination prior and
  outside-in visibility,'' in \emph{Proceedings of the European conference on
  computer vision (ECCV)}, 2024, pp. 126--143.

\bibitem{gs-ir}
Z.~Liang, Q.~Zhang, Y.~Feng, Y.~Shan, and K.~Jia, ``{ GS-IR: 3D Gaussian
  Splatting for Inverse Rendering },'' in \emph{2024 IEEE/CVF Conference on
  Computer Vision and Pattern Recognition (CVPR)}, 2024.

\bibitem{kaleta2024lumigauss}
J.~Kaleta, K.~Kania, T.~Trzcinski, and M.~Kowalski, ``Lumigauss: Relightable
  gaussian splatting in the wild,'' 2024.

\bibitem{chen2025gigs}
H.~Chen, Z.~Lin, and J.~Zhang, ``Gi-gs: Global illumination decomposition on
  gaussian splatting for inverse rendering,'' in \emph{ICLR}, 2025.

\bibitem{zhang2025scaling}
L.~Zhang, A.~Rao, and M.~Agrawala, ``Scaling in-the-wild training for
  diffusion-based illumination harmonization and editing by imposing consistent
  light transport,'' in \emph{The Thirteenth International Conference on
  Learning Representations}, 2025.

\bibitem{chen2025multi}
M.~Chen, M.~Yang, Y.~Zhang, T.~Han, X.~Li, H.~Zhao, and P.~Liu, ``Multi-modal
  bev enhancement fusion for 3d object detection in autonomous driving,''
  \emph{IEEE Transactions on Intelligent Transportation Systems}, 2025.

\bibitem{wang2024towards}
J.~Wang, F.~Li, Y.~An, X.~Zhang, and H.~Sun, ``Towards robust lidar-camera
  fusion in bev space via mutual deformable attention and temporal
  aggregation,'' \emph{IEEE Transactions on Circuits and Systems for Video
  Technology}, 2024.

\bibitem{lang2024bev}
B.~Lang, X.~Li, and M.~C. Chuah, ``Bev-tp: end-to-end visual perception and
  trajectory prediction for autonomous driving,'' \emph{IEEE Transactions on
  Intelligent Transportation Systems}, 2024.

\bibitem{hu2023planning}
Y.~Hu, J.~Yang, L.~Chen, K.~Li, C.~Sima, X.~Zhu, S.~Chai, S.~Du, T.~Lin,
  W.~Wang \emph{et~al.}, ``Planning-oriented autonomous driving,'' in
  \emph{Proceedings of the IEEE/CVF conference on computer vision and pattern
  recognition}, 2023, pp. 17\,853--17\,862.

\bibitem{jiang2023vad}
B.~Jiang, S.~Chen, Q.~Xu, B.~Liao, J.~Chen, H.~Zhou, Q.~Zhang, W.~Liu,
  C.~Huang, and X.~Wang, ``Vad: Vectorized scene representation for efficient
  autonomous driving,'' in \emph{Proceedings of the IEEE/CVF International
  Conference on Computer Vision}, 2023, pp. 8340--8350.

\bibitem{zhou2025opendrivevla}
X.~Zhou, X.~Han, F.~Yang, Y.~Ma, and A.~C. Knoll, ``Opendrivevla: Towards
  end-to-end autonomous driving with large vision language action model,''
  \emph{arXiv preprint arXiv:2503.23463}, 2025.

\bibitem{zhou2025autovla}
Z.~Zhou, T.~Cai, S.~Z. Zhao, Y.~Zhang, Z.~Huang, B.~Zhou, and J.~Ma, ``Autovla:
  A vision-language-action model for end-to-end autonomous driving with
  adaptive reasoning and reinforcement fine-tuning,'' \emph{arXiv preprint
  arXiv:2506.13757}, 2025.

\bibitem{huang2025seele}
X.~Huang, H.~Zhu, Z.~Liu, W.~Lin, X.~Liu, Z.~He, J.~Leng, M.~Guo, and Y.~Feng,
  ``Seele: A unified acceleration framework for real-time gaussian splatting,''
  \emph{arXiv preprint arXiv:2503.05168}, 2025.

\bibitem{liu2025mobilegaussian}
X.~Liu, D.~Shi, and S.~Yang, ``Mobilegaussian: Efficient 3d gaussian-based open
  vocabulary scene understanding on mobile devices,'' in \emph{Proceedings of
  the 2025 2nd International Conference on Computer Network and Cloud
  Computing}, 2025, pp. 76--81.

\bibitem{hu2024lowlatency}
Y.~Hu, R.~Gong, Q.~Sun, and Y.~Wang, ``Low latency point cloud rendering with
  learned splatting,'' in \emph{Proceedings of the IEEE/CVF Conference on
  Computer Vision and Pattern Recognition (CVPR) Workshops}, June 2024, pp.
  5752--5761.

\bibitem{liu2025voyager}
Z.~Liu, H.~Zhu, X.~Li, Y.~Wang, Y.~Shi, W.~Li, J.~Leng, M.~Guo, and Y.~Feng,
  ``Voyager: Real-time splatting city-scale 3d gaussians on your phone,'' 2025.

\bibitem{ISO26262}
{International Organization for Standardization}, ``Iso 26262-1:2018 road
  vehicles — functional safety,''
  \url{https://www.iso.org/standard/68383.html}, 2018, [Online; accessed
  8-October-2025].

\bibitem{ISO21448}
------, ``Iso 21448:2022 road vehicles — safety of the intended
  functionality,'' \url{https://www.iso.org/standard/77490.html}, 2022,
  [Online; accessed 8-October-2025].

\bibitem{koopman2023ul}
P.~Koopman, ``Ul 4600: What to include in an autonomous vehicle safety case,''
  \emph{Computer}, vol.~56, no.~05, pp. 101--104, 2023.

\bibitem{eu}
{the European Commission}, ``Commission implementing regulation (eu)
  2022/1426,'' \url{https://eur-lex.europa.eu/eli/reg_impl/2022/1426/oj/eng},
  2022, [Online; accessed 8-October-2025].

\bibitem{japan}
{National Police Agency of Japan}, ``Guidelines for public road testing of
  automated driving systems,''
  \url{https://www.npa.go.jp/english/bureau/traffic/guideline.pdf}, 2016,
  [Online; accessed 8-October-2025].

\end{thebibliography}

\section{Biography}
        
\begin{IEEEbiography}[{\includegraphics[width=1.0in,height=1.25in,clip,keepaspectratio]{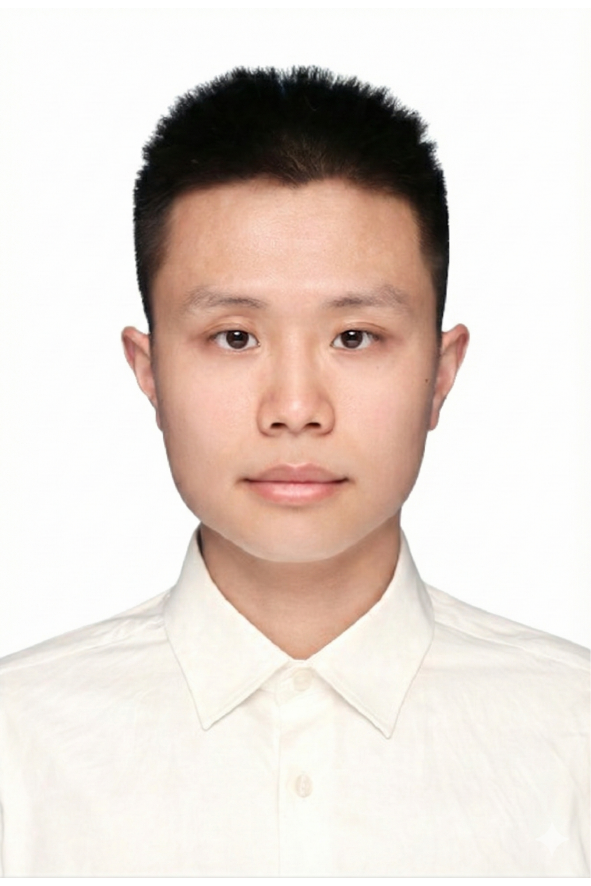}}] {\textbf{Liewen Liao}}
received a Master's degree in Computer Science from Shanghai Jiao Tong University, Shanghai, China, in 2022. He is working towards a Ph.D. degree in Control Science and Engineering from Shanghai Jiao Tong University. His main fields of interest are 3D reconstruction, 3D generation and closed-loop simulation in autonomous driving.
\end{IEEEbiography}

\begin{IEEEbiography}[{\includegraphics[width=1.0in,height=1.25in,clip,keepaspectratio]{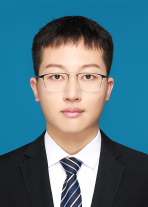}}]{\textbf{Weihao Yan}}
received a Bachelor's degree in Automation from Shanghai Jiao Tong University, Shanghai, China, in 2020. He is working towards a Ph.D. degree in Control Science and Engineering from Shanghai Jiao Tong University. His main fields of interest are autonomous driving systems, computer vision, and domain adaptation. His current research activities include virtual-to-real transfer learning, scene segmentation, and foundation models.
\end{IEEEbiography}

\begin{IEEEbiography}[{\includegraphics[width=1.0in,height=1.25in,clip,keepaspectratio]{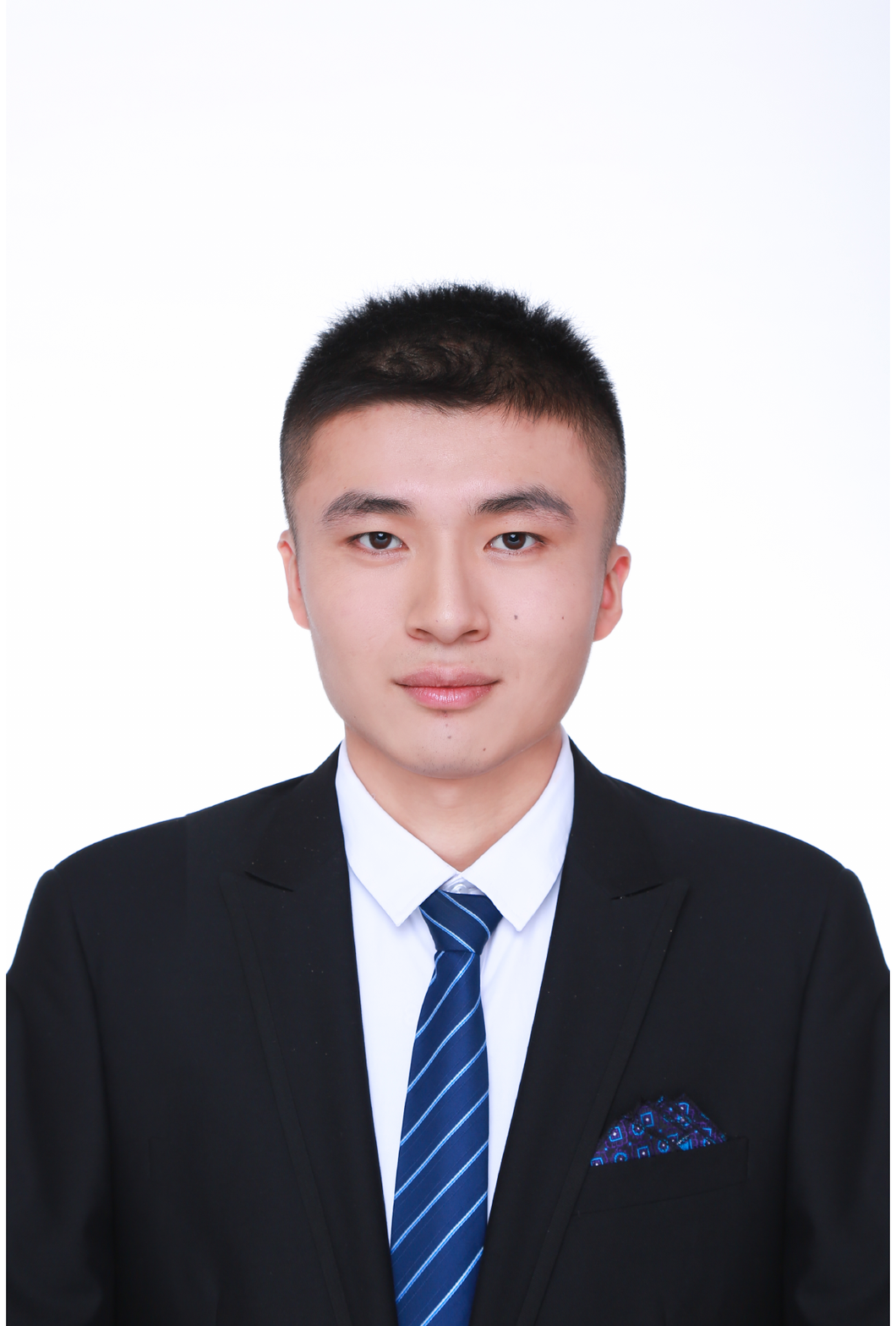}}]{\textbf{Wang Xu}}
received the B.S. in school of instrument science and technology from the Southeast University, Nanjing, China, in 2022. He is currently working toward the Ph.D degree in Control Science and Engineering at Shanghai Jiao Tong University. His research interests main include 3D scene reconstruction, computer vision and autonomous driving simulation.
\end{IEEEbiography}

\begin{IEEEbiography}[{\includegraphics[width=1.0in,height=1.25in,clip,keepaspectratio]{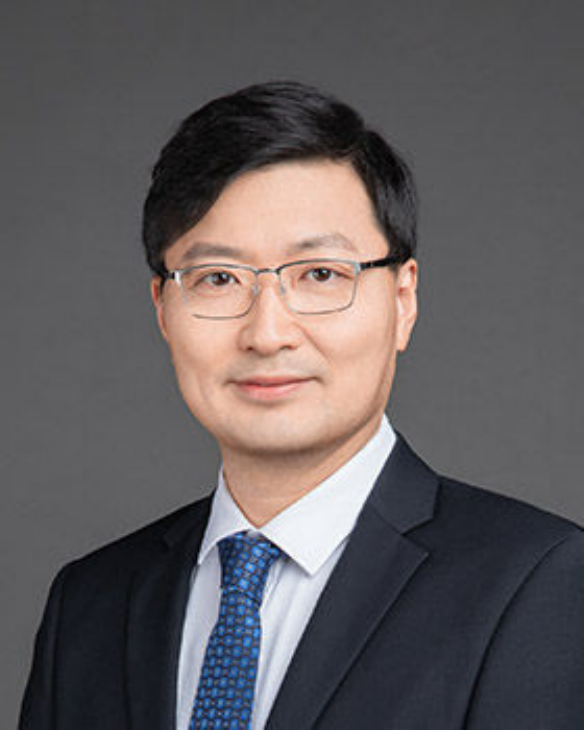}}]{\textbf{Ming Yang}}
received the Master and Ph.D. degrees from Tsinghua University, Beijing, China, in 1999 and 2003, respectively. He is currently the Full Tenure Professor at Shanghai Jiao Tong University, the deputy director of the Innovation Center of Intelligent Connected Vehicles. He has been working in the field of intelligent vehicles for more than 20 years. He participated in several related research projects, such as the THMR-V project (first intelligent vehicle in China), European CyberCars and CyberMove projects, CyberC3 project, CyberCars-2 project, ITER transfer cask project, AGV, etc.
\end{IEEEbiography}

\begin{IEEEbiography}[{\includegraphics[width=1.25in,height=1.75in,clip,keepaspectratio]{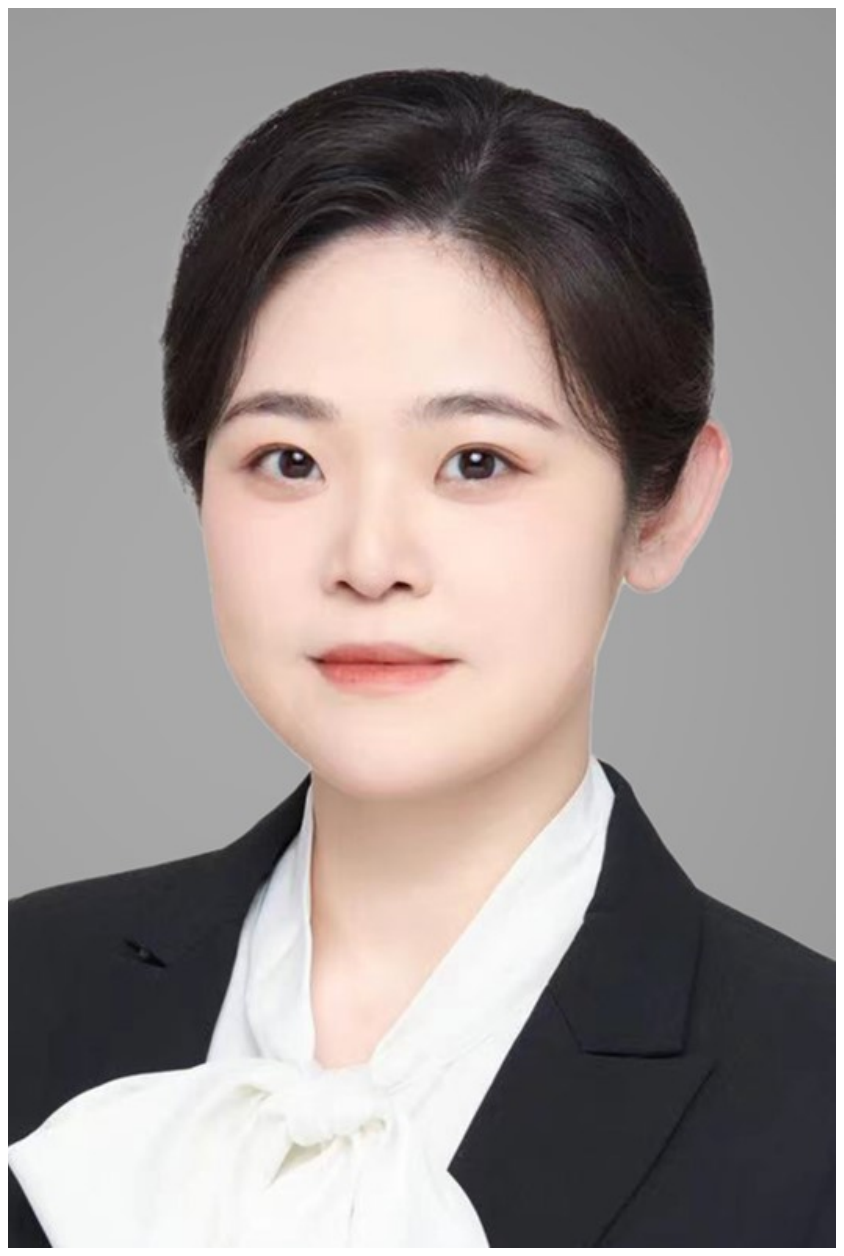}}]{\textbf{Songan Zhang}} received the B.S. and M.S. degrees in automotive engineering from Tsinghua University, Beijing, China, in 2013 and 2016, respectively. In 2021, she earned the Ph.D. degree in mechanical engineering from the University of Michigan, Ann Arbor, MI, USA. Upon graduation, she joined Ford Motor Company as a Research Scientist where she made contributions to pioneering innovations in smart manufacturing and advanced driver assist systems. She is currently an Assistant Professor with the Global Institute of Future Technology, Shanghai Jiao Tong University, Shanghai, China. Her research interests include accelerated and safety evaluation of autonomous vehicles; verification methods for autonomous driving systems; model-based, trustworthy, and data-efficient reinforcement learning and meta-learning; and the application of foundation models for decision-making in autonomous vehicles and intelligent transportation systems.
\end{IEEEbiography}

\vspace{-35.0em}
\begin{IEEEbiography}[{\includegraphics[width=1.0in,height=1.25in,clip,keepaspectratio]{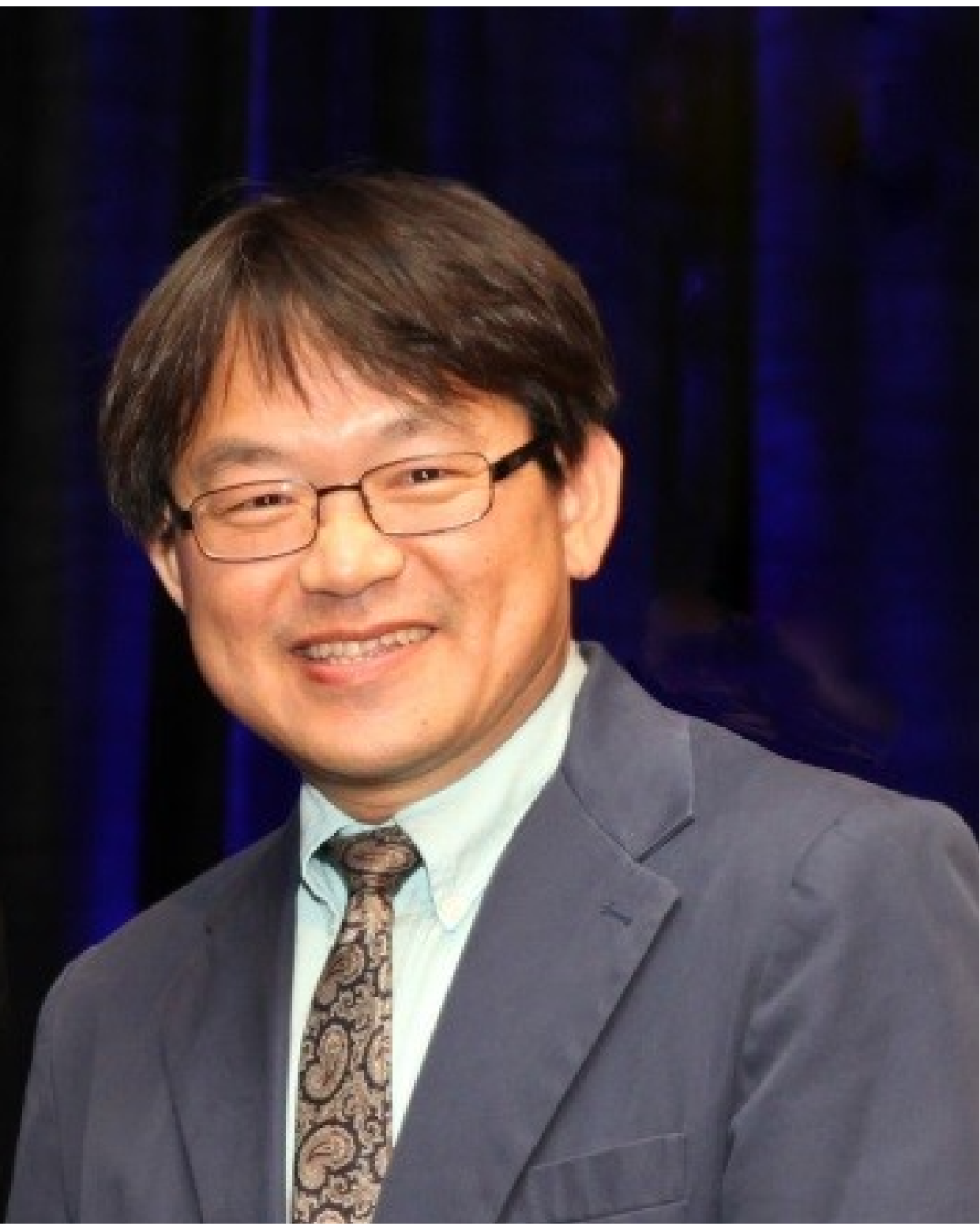}}]{\textbf{H. Eric Tseng}}
H. Eric Tseng received the B.S. degree from the National Taiwan University in 1986, and the M.S. and Ph.D. degrees in mechanical engineering from the University of California, Berkeley in 1991 and 1994, respectively. In May 2024, he joined the University of Texas at Arlington as a distinguished university professor in the Department of Electrical Engineering.
In 1994, he joined Ford Motor Company. At Ford (1994-2022), he had a productive career and retired as a Senior Technical Leader of Controls and Automated Systems in Research and Advanced Engineering. Many of his contributed technologies led to production vehicles implementation. His technical achievements have been honored with Ford's annual technology award, the Henry Ford Technology Award, on seven occasions. Additionally, he was the recipient of the Control Engineering Practice Award from the American Automatic Control Council in 2013, and the recipient of the Soichiro Honda Medal from the American Society of Mechanical Engineering in 2024. He has over 100 U.S. patents and over 160 publications. He is a member of the National Academy of Engineering as of 2021.
\end{IEEEbiography}

\end{document}